\documentclass[nohyperref]{article}

\usepackage{microtype}
\usepackage{graphicx}
\usepackage{subcaption}
\usepackage{booktabs} % for professional tables
\usepackage{makecell}
\usepackage{multirow}
\usepackage{bm}
\usepackage{tikz}
\usetikzlibrary{arrows.meta}
\usepackage{xcolor,mdframed}
\usepackage{comment}

\usepackage{hyperref}

\usepackage[accepted]{arxiv}

\usepackage{amsmath}
\usepackage{amssymb}
\usepackage{mathtools}
\usepackage{amsthm}
\usepackage{euscript}

\renewcommand{\hat}{\widehat}

\renewcommand{\bar}{\overline}

\newcommand{\define}{\textbf}

\renewcommand{\hat}{\widehat}

\renewcommand{\epsilon}{\varepsilon}

\newcommand{\DON}{\mathrm{DON}}

\newcommand{\don}{\DON}

\newcommand{\FNO}{\mathrm{FNO}}
\newcommand{\fno}{\FNO}

\newcommand{\du}{{d_u}}
\newcommand{\R}{\mathbb{R}}
\newcommand{\C}{\mathbb{C}}

\newcommand{\N}{\mathbb{N}}

\newcommand{\Z}{\mathbb{Z}}

\newcommand{\cB}{\mathcal{B}}

\newcommand{\cE}{\mathcal{E}}
\newcommand{\cF}{\mathcal{F}}

\newcommand{\cL}{\mathcal{L}}

\newcommand{\cN}{\mathcal{N}}

\newcommand{\cX}{\mathcal{X}}
\newcommand{\cY}{\mathcal{Y}}

\newcommand{\eA}{\EuScript{A}}
\newcommand{\eU}{\EuScript{U}}
\newcommand{\eD}{\EuScript{D}}
\newcommand{\eG}{\EuScript{G}}
\newcommand{\eH}{\EuScript{H}}
\newcommand{\eB}{\EuScript{B}}
\newcommand{\eF}{\EuScript{F}}
\newcommand{\eL}{\EuScript{L}}
\newcommand{\eS}{\EuScript{S}}
\newcommand{\eQ}{\EuScript{Q}}
\newcommand{\eR}{\EuScript{R}}
\newcommand{\eE}{\EuScript{E}}
\newcommand{\eM}{\EuScript{M}}
\newcommand{\eI}{\EuScript{I}}

\usepackage[capitalize,noabbrev]{cleveref}

\theoremstyle{plain}

\theoremstyle{definition}

\theoremstyle{remark}

\usepackage[textsize=tiny]{todonotes}

\icmltitlerunning{Neural Inverse Operators}

\begin{document}

\twocolumn[
\icmltitle{Neural Inverse Operators for Solving PDE Inverse Problems}




\begin{icmlauthorlist}
\icmlauthor{Roberto Molinaro}{sam}
\icmlauthor{Yunan Yang}{its}
\icmlauthor{Bj\"orn Engquist}{tex}
\icmlauthor{Siddhartha Mishra}{sam,ai}
\icmlcorrespondingauthor{Roberto Molinaro}{roberto.molinaro@sam.math.ethz.ch}
\end{icmlauthorlist}

\icmlaffiliation{sam}{Seminar for Applied Mathematics (SAM), ETH, Z\"urich.}

\icmlaffiliation{its}{Institute for Theoretical Studies (ITS), ETH Z\"urich.}

\icmlaffiliation{tex}{Department of Mathematics and the Oden Institute, The University of Texas at Austin, USA.}

\icmlaffiliation{ai}{ETH AI Center, ETH, Z\"urich}

\icmlkeywords{PDE, Inverse Problems, Machine Learning, ICML}

\vskip 0.3in
]



\printAffiliationsAndNotice{}  % leave blank if no need to mention equal contribution

\begin{abstract}
A large class of inverse problems for PDEs are only well-defined as mappings from operators to functions. Existing operator learning frameworks map functions to functions and need to be modified to learn inverse maps from data. We propose a novel architecture termed Neural Inverse Operators (NIOs) to solve these PDE inverse problems. Motivated by the underlying mathematical structure, NIO is based on a suitable composition of DeepONets and FNOs to approximate mappings from operators to functions. A variety of experiments are presented to demonstrate that NIOs significantly outperform baselines and solve PDE inverse problems robustly, accurately and are several orders of magnitude faster than existing direct and PDE-constrained optimization methods.
\end{abstract}
\section{Introduction.}
Partial differential equations (PDEs) are ubiquitous as mathematical models in the sciences and engineering \cite{Evansbook}. Often, solving PDEs entails solving the so-called \emph{forward problem}. That is, given inputs such as initial and boundary conditions, coefficients, and sources, compute (observables of) the solution of the PDE. However, in many important contexts in applications, one is instead interested in solving the so-called \emph{inverse problem}~\cite{INVbook}. That is, given measurements of (observables of) the solution of a PDE, infer the underlying inputs. 

A large class of such inverse problems takes the following abstract form: given observables as operators (mappings between function spaces), infer the underlying input coefficient (functions) of the associated PDE. A prototypical example is the well-studied \emph{Calder\'{o}n Problem}~\cite{eit} that arises in electrical impedance tomography (EIT) in medical imaging. Here, the observable is the Dirichlet-to-Neumann (DtN) operator that maps the voltage on the boundary to the current, and one is interested in inferring the underlying conductivity field. A related example is inverse wave scattering for geophysical applications. Other examples include optical tomography \cite{lai2019inverse} where the observable is the so-called \emph{Albedo operator}, and one needs to infer the scattering and absorption coefficients of the underlying medium. Another prominent example arises in \emph{seismic imaging} in geophysics \cite{SIbook} where the observable is the source-to-receiver (StR) operator, and the task at hand is to infer the underlying sub-surface properties such as wave velocity or material density. In many of these examples, the solution to the resulting inverse problem is unique and stable if and only if the inverse problem is posed as a \emph{mapping from operators to functions}. 

Given the nonlinear nature of most of these inverse problems, analytical solution formulas are not available except in some simple cases. Instead, iterative numerical algorithms based on 
the PDE-constrained optimization are commonly used to approximate the solution~\cite{chavent2010nonlinear}. These algorithms repeatedly apply the forward and adjoint PDE solvers to converge to the unknown coefficient. However, a large number of iterations might be necessary, which leads to prohibitively high computational costs as numerous calls to PDE solvers are very expensive, particularly in two and three space dimensions. Moreover, these iterative algorithms can be sensitive to the choice of initial coefficient. Given these factors, the design of alternative approaches to solving inverse problems is imperative. 

Data-driven approximation of PDEs is rapidly emerging as a powerful paradigm. Most of the available results pertain to \emph{forward problems} for PDEs. A particularly popular framework is \emph{operator learning}, where one seeks to learn the underlying forward solution operator of the PDE from data. Existing approaches to operator learning include Deep operator networks (DeepONet)~\cite{,lu2021learning,donet1,donet2} and their variants, as well as the so-called neural operators~\cite{NO}, which include the widely used Fourier Neural Operators (FNO) and its variants~\cite{FNO1,FNO2}. Graph neural network-based algorithms~\cite{bengioneural,well} are also emerging as an alternative. 

Given the widespread success of operator learning and other deep learning-based algorithms in the context of forward problems for PDEs, it is natural to investigate their utility in learning the solutions of the corresponding inverse problems from data. However, this task is very challenging as existing operator learning algorithms map functions to functions. On the other hand, the aforementioned inverse problems are only well-defined as maps from operators to functions. Hence, one needs non-trivial modifications of existing operator learning architectures to handle inverse problems. This is precisely the rationale for the current paper, where our main contributions are the following.
\begin{itemize}
\item Motivated by the underlying mathematical structure of the considered class of inverse problems, we propose a novel architecture, termed as \emph{Neural Inverse Operators} (NIOs), for learning solutions of these inverse problems from data. NIOs compose (stack) two existing architectures, DeepONet and FNO, in order to map operators to functions.
\item We test NIOs extensively on a suite of problems, including the Calder\'{o}n problem in EIT, inverse wave scattering for object detection, reconstructing the absorption and scattering coefficients in optical tomography, and seismic wave migration to infer sub-surface properties. We show that NIO significantly outperforms baselines on all these benchmarks.
\item Given the inherent sensitivity of PDE inverse problems to noise and perturbations, we also demonstrate empirically that NIO is robust with respect to these factors as well as to the number and location of sensors and underlying grid resolutions. Moreover, we also show that, unlike baselines, NIO can handle fluctuations in the number of input measurements as well as generalize \emph{out-of-distribution} without any significant loss of accuracy. Furthermore, NIO is found to be several orders of magnitude \emph{faster} while still being more accurate than traditional PDE-constrained optimization methods. 
\end{itemize}
Thus, we propose a new learning framework for the fast, robust and accurate solution of PDE inverse problems. 
\section{A Class of Inverse Problems.}\label{sec:ips}
%We start by providing an abstract formalism for the class of PDE inverse problems that we consider herein.
\subsection{Mathematical Framework.} Let $D \subset \R^d$ be a bounded open set, with (smooth) boundary $\partial D$. Let $T > 0$ and $\Omega = D$ or $\Omega = D \times (0,T)$, depending on whether the PDE is time-(in)dependent. Correspondingly, $\partial \Omega = \partial D$ or $\partial \Omega = \partial D \times (0,T)$, respectively. 
Let $a \in \eA(D)$, with $\eA$ denoting a suitable function space over $D$, be a coefficient. Then, an abstract PDE can be written as 
\begin{equation}
\label{eq:pde}
\eD_a(u) = s, \quad \eB(u) = g, 
\end{equation}
where $u \in \eU(\Omega)$ is the solution, $s \in \eS(\Omega)$ is the source term and $g \in \eG(\partial \Omega)$ is the boundary condition, for the PDE \eqref{eq:pde}. Here, $\eD_a: \eU \mapsto \eF$ and $\eB: \eU \mapsto \eG$ are the differential and boundary operators, respectively and $\eU,\eS,\eG$ are suitable function spaces, defined over their respective domains.

The \emph{forward problem} for the abstract PDE \eqref{eq:pde} amounts to the following: given the coefficient $a \in \eA$, source term $s \in \eS$ and boundary condition $g \in \eG$, find the solution $u \in \eU$ of the PDE \eqref{eq:pde}. Often, one is interested in not only the solution itself but also \emph{observables} of the solution, which can be measured in practice. Since measurements are usually easier to perform at boundaries, a particularly relevant class of such observables are given by the following \emph{boundary observation operator},
\begin{equation}
\label{eq:obsop}
\Lambda_a: \eG(\partial \Omega) \mapsto \eH(\partial \Omega),
\end{equation}
which maps the boundary data $g \in \eG(\partial \Omega)$ to a measurement $\Lambda_a(g) = h(u) \in \eH(\partial \Omega)$, a function space on $\partial \Omega$. 
Thus, for a fixed coefficient $a$ (and source $s$), solving the forward problem amounts to solving the PDE \eqref{eq:pde}, with a given boundary data $g$ to obtain the solution $u$ and then post-processing $u$ to compute the boundary observation operator $h(u) = \Lambda_a(g)$. Hence, one can rewrite the forward problem associated with the PDE~\eqref{eq:pde} to obtain the map, 
\begin{equation}
\label{eq:fmap}
\eF: \eA(D) \mapsto \eL\left(\eG(\partial \Omega), \eH (\partial \Omega)\right), ~ a \mapsto \eF(a) = \Lambda_a,
\end{equation}
where $\Lambda_a$ is the boundary observation operator~\eqref{eq:obsop} and $\eL(X,Y)$ denotes continuous operators between function spaces $X$ and $Y$. 

In practice, one is often interested in the \emph{inverse problem} associated with the PDE \eqref{eq:pde}. For instance, in tomography (imaging), one needs to infer the unknown coefficient $a$ from some measurements of the solution $u$. In general, this problem is \emph{ill-posed}, and a single instance (or small number) of boundary conditions $g$ and measurements $h(u)$ of the corresponding solutions $u$, do not suffice in inferring the underlying coefficient $a$. Instead, many deep mathematical results have provided suitable frameworks where such inverse problems can be well-posed. The inverse map for the forward problem~\eqref{eq:fmap} takes the form
\begin{equation}
\label{eq:imap}
\eF^{-1}: \eL\left(\eG(\partial \Omega), \eH (\partial \Omega)\right) \mapsto \eA(D), ~ \Lambda_a \mapsto a = \eF^{-1}(\Lambda_a),
\end{equation}
The rigorous guarantee of the existence and, more importantly, the uniqueness of this inverse map $\eF^{-1}$, for a large class of PDEs, is a crowning achievement of the mathematical theory of inverse problems~\cite{INVbook}. Moreover, one can also show Lipschitz or H\"older-\emph{stability} of the inverse problem by proving estimates of the form, 
\begin{equation}
\label{eq:stab1}
\|\eF^{-1}(a) - \eF^{-1}(\bar{a})\|_{\eL} \sim \|a - \bar{a}\|_\eA^{\alpha}, 0 < \alpha \leq 1.
\end{equation}
In some cases, the right-hand side of the above stability estimate is replaced by a logarithm of $\|a-\bar{a}\|_{\eA}$, which only guarantees (weak) logarithmic stability.

After presenting this abstract framework, we provide four concrete examples of PDE inverse problems (see {\bf SM} Figures \ref{fig:sin_sample}-\ref{fig:curve_sample} for illustrations) to which this abstract framework applies. 
\subsection{Calder\'{o}n Problem (EIT).}
Let the coefficient $0 < a \in C^{2}(D)$ represent the conductivity of the underlying medium (domain $D \subset \R^d$) and the associated PDE~\eqref{eq:pde} is the following elliptic equation,
\begin{equation}
        \label{eq:calderon}
        \begin{aligned}
            -\nabla\cdot\Big(a(z)\nabla u\Big) = 0, \quad &z\in D, \\
            u(z) = g(z), \quad &z\in \partial D,
        \end{aligned}
        \end{equation}
with Dirichlet boundary value $g \in H^{\frac{1}{2}}(\partial D)$ representing the voltage and the current source term is $s=0$.  The associated boundary observation operator $\Lambda_a$ is the well-known \emph{Dirichlet-to-Neumann} (DtN) map,
\begin{equation}
\label{eq:dtn}
\begin{aligned}
 &\Lambda_a: H^{1/2}(\partial D) \mapsto H^{-1/2}(\partial D), \\
 & \Lambda_a[g]  = a \frac{\partial u}{\partial \nu}\Big|_{\partial D}, ~ \forall g\in H^{1/2}(\partial D),
\end{aligned}
\end{equation}
which maps the input voltage $g$ into the current $a (z) \frac{\partial u}{\partial \nu} = a\,\nabla u \cdot \nu$ (with $\nu$ being the unit outward normal vector) at the boundary and $u$ is the solution of~\eqref{eq:calderon}.

The inverse problem, often referred to as the Calder\'{o}n problem, constitutes the basis of EIT~\cite{eit}. It aims to find the conductivity $a$ of the medium, given different measurements of the DtN (voltage-to-current) pairs. Thus, this inverse problem falls into the considered abstract formalism and the inverse map \eqref{eq:imap} is given by, 
\begin{equation}
\label{eq:cpimap}
\begin{aligned}
&\eF^{-1}: \eL\left(H^{1/2}(\partial D), H^{-1/2}(\partial D) \right) \mapsto C^{2}(D), \\
&\eF^{-1}:\Lambda_a \mapsto a = \eF^{-1}(\Lambda_a),
\end{aligned}
\end{equation}
with $\eL(\cdot,\cdot)$ denoting the corresponding bounded linear operators. This inverse problem is shown to be well-defined and (logarithmic-) stable~\cite{clop2010stability}. 
\subsection{Inverse Wave Scattering.}
\label{sec:helm}
In many applications of interest, wave propagation in the frequency domain is used to infer material properties of the medium, modelled by the squared slowness $0 < a \in L^\infty(D)$. The associated PDE is the Helmholtz equation,
\begin{equation}
        \label{eq:helmholtz}
        \begin{aligned}
           - \Delta u - \omega^2a(z)u = 0, \quad &z\in D,\\
            u(z) = g(z), \quad &z\in \partial D,
        \end{aligned}
        \end{equation}
for some frequency $\omega$ and Dirichlet boundary condition $g \in H^{\frac{1}{2}}(\partial D)$. The resulting boundary observation operator is again the Dirichlet-to-Neumann (DtN) map
\begin{equation}
\label{eq:dtn2}
\begin{aligned}
 &\Lambda_a: H^{1/2}(\partial D) \mapsto H^{-1/2}(\partial D), \\
 & \Lambda_a[g]  = \frac{\partial u}{\partial \nu}\Big|_{\partial D}, ~ \forall g\in H^{1/2}(\partial D),
\end{aligned}
\end{equation}
where $u$ is the solution to~\eqref{eq:helmholtz} with the coefficient $a$. The corresponding inverse problem amounts to inferring the wave coefficient $a$ from the DtN map \eqref{eq:dtn2}. Thus, it can be formulated similar to the inverse map~\eqref{eq:cpimap}. Its well-posedness and stability have been demonstrated for the Helmholtz equation in \cite{Nach} and references therein. 
\subsection{Radiative Transport and Optical Imaging.}
In optical imaging or tomography, the material properties of the medium $D \subset \R^d$ are expressed in terms of the scattering and absorption coefficients, $0\leq a,\sigma_a \in C(D)$. The associated PDE is the well-known radiative transport equation (RTE) for the particle density $u(z,v)$ at location $z \in D$ and velocity $v \in V \subset \R^d$, given by
\begin{equation}
        \begin{aligned} \label{eq:RTE}
            v\cdot \nabla_z u(z,v)  + \sigma_a(z)u(z,v)= \frac{1}{\epsilon} a(z) \eQ[u], \quad & z\in D,\\
            u(z,v) = \phi(z,v), \quad & z \in \Gamma_{-},
            \end{aligned}
        \end{equation}
        where $\eQ[u] = \int k(v,v') u(z,v') dv' - u(z,v) $ is the collision term, $\epsilon$ is the Knudsen number, $\Gamma_{\pm} = \{(z,v) \in \partial D \times V : \pm \,n_z \cdot v \geq 0 \}$ are the inflow (outflow) boundaries and $n_z$ is the unit outer normal vector at  $z\in \partial D$. Thus, the input to this problem is provided by the particle density, $u_{\Gamma_{-}}\in L^1(\partial D)$, prescribed on the inflow boundary. The associated boundary observation operator $\Lambda_a$ defined in~\eqref{eq:obsop} is the so-called \emph{Albedo} operator,
        \begin{equation}
        \label{eq:albop}
\Lambda_a: L^1(\Gamma_-) \mapsto L^1(\Gamma_+), ~ \Lambda_a: u\big|_{\Gamma_{-}} = \phi \mapsto u\big|_{\Gamma_+},
        \end{equation}
that maps the incident boundary values on $\Gamma_-$ to the observed boundary values on the outflow boundary $\Gamma_+$. 

The corresponding inverse problem aims to  infer the medium properties characterized by the scattering and absorption coefficients $a,\sigma_a$ from the measurements of the Albedo operator. It leads to the following inverse map,
\begin{equation}
\label{eq:otimap}
\begin{aligned}
&\eF^{-1}: \eL\left(L^1(\Gamma_-), L^1(\Gamma_+) \right) \mapsto C(D), \\
&\eF^{-1}:\Lambda_a \mapsto a = \eF^{-1}(\Lambda_a).
\end{aligned}
\end{equation}
The well-posedness and Lipschitz-stability of this inverse map were shown in \cite{bal2008stability}.
\subsection{Seismic Imaging.} Seismic imaging is widely used in geophysics to infer and reconstruct sub-surface material properties for various applications such as $\text{CO}_2$ storage monitoring and seismic hazard assessment. Given a domain $D \subset \R^d$, we are interested in reconstructing the velocity coefficient $0 < a \in L^\infty(D)$ by sending in acoustic waves from the top boundary into the medium and measuring the response in the time domain. The associated PDE is the acoustic wave equation, 
\begin{equation}
    \label{eq:wave}
    \begin{aligned}
        u_{tt}(t, z) + a^2(z)\Delta u  = s, ~ (z,t) \in D \times [0,T],
    \end{aligned}
    \end{equation}
with a time-dependent source term $s$. Here, $u$ is the pressure variation. The wave equation is supplemented with zero initial conditions, i.e., $u(\cdot,0) = u_t(\cdot,0) = 0$ and suitable boundary conditions. In particular, \emph{sources} are placed on a subset of the boundary and it is common to consider point sources $s(t,z) = g(t)\delta_{S}(z)$ with $g(t) \in L^2([0,T])$ and $\delta_{S}(z)$ being the Dirac measure concentrated on a set $S\subset \partial D$. These waves are transmitted, reflected, and refracted through the medium. Under certain assumptions (see Sec.~2.3 in~\citet{symes2009seismic}), the \textit{effective} source $s$ can be treated as in  $L^2([0,T]\times D)$, which ensures the well-posedness of the PDE. The resulting signal is recorded at a set of receivers $\eR  \subset \partial D$ on the boundary that take continuous measurement in time $[0,T]$. The boundary observation operator \eqref{eq:obsop} for this wave inverse problem is the \emph{Source-to-Receiver} (StR) operator,
\begin{equation}\label{eq:str}
\begin{aligned}
& \Lambda_a: L^2([0,T]\times D)\mapsto L^2([0,T]; X_\eR),\\
& \Lambda_a: s \mapsto u\big|_{[0,T]\times \eR},
\end{aligned}
\end{equation}
where $X_\eR$ is the metric space for the (discrete) set $\eR$. The inverse problem that underpins seismic imaging is
\begin{eqnarray}
&\eF^{-1}: &\eL\left(L^2([0,T]\times D), L^2([0,T]; X_\eR) \right) \mapsto L^\infty(D), \nonumber \\
&\eF^{-1}: &\Lambda_a \mapsto a = \eF^{-1}(\Lambda_a),\label{eq:fwimap} 
\end{eqnarray}
with $\Lambda_a$ being the StR operator \eqref{eq:str}. Thus, seismic imaging aims to infer the subsurface spatial medium properties from spatial-temporal StR signals. This process is also termed as \emph{migration}, or \emph{Full waveform Inversion} (FWI) in the literature \cite{deng2021openfwi}. There have been studies on the well-posedness of the inverse problem for the wave equation~\eqref{eq:wave}~\cite{liu2016lipschitz,stefanov2016stable,caday2019scattering} although they do not directly apply to the setting considered here.

% In particular, \emph{sources} are placed on a subset of the boundary, $\Sigma \subset \partial D \times [0,T]$. The surface and incident acoustic waves are sent into the domain through a (time-dependent) Dirichlet boundary condition, $u\big|_{\Sigma} = g$. These waves are transmitted, reflected and refracted through the medium and the resulting signal is recorded at a set of receivers on the boundary given by $\eR \subset (D \times [0,T])\setminus \Sigma$. Consequently, the boundary observation operator \eqref{eq:obsop} in this case is the \emph{Source-to-Receiver} (StR) operator given by, 
%  \begin{equation}
%         \label{eq:str}
% \Lambda_a: L^p(\Sigma) \mapsto L^p(\eR), ~ \Lambda_a: u\big|_{\Sigma} \mapsto u\big|_{\eR}.
%         \end{equation}

% The resulting inverse problem, that underpins seismic imaging is given by, 
% \begin{equation}
% \label{eq:fwimap}
% \begin{aligned}
% &\eF^{-1}: \eL\left(L^p(\Sigma), L^p(\eR) \right) \mapsto C^k(D), \\
% &\eF^{-1}:\Lambda_a \mapsto a = \eF^{-1}(\Lambda_a),
% \end{aligned}
% \end{equation}
% with $\Lambda_a$ is the StR operator \eqref{eq:str}. Thus in seismic imaging, one aims to infer the sub-surface (spatial) velocity profile from temporal StR signals. This process is also termed as \emph{migration} or \emph{Full waveform Inversion} (FWI) in the literature \cite{FWI}.
\section{Neural Inverse Operators.}
In this section, we present the neural network architecture for the proposed  Neural Inverse Operators (NIOs).
\subsection{Learning Task and Challenges.} 
\label{sec:challenges}
All four examples described in the previous section were particular instances of the abstract framework summarized in \eqref{eq:imap}. Thus, the solution of the inverse problem \eqref{eq:imap} boils down to inferring (learning) the inverse map $\eF^{-1}$ from relevant data. Given sufficient training data in the form of pairs $\left(\Lambda_a,\eF^{-1}(\Lambda_a)\right)$ (or given the injectivity of the forward map, data in the form of pairs $\left(a,\Lambda_a\right)$), we aim to learn the inverse map $\eF^{-1}$ and evaluate it on \emph{test} (unseen) data. This task is very challenging on account of the following factors:
\begin{enumerate}
\item The inputs to the inverse map $\eF^{-1}$ \eqref{eq:imap} are specified on the boundaries $\partial \Omega$ whereas the output is the coefficient $a$, defined in the interior of the underlying domain $D$. Thus, there is a mismatch in the domains of the inputs and outputs for the inverse map $\eF^{-1}$.
\item The learning task requires us to learn \emph{mappings from operators to functions} for $\eF^{-1}$ defined in~\eqref{eq:imap}. 
\item In general, the inverse map $\eF^{-1}$ \eqref{eq:imap} may only be weakly stable, for instance, either in terms of small values of the H\"older exponent $\alpha$ in \eqref{eq:stab1} or even only logarithmic-stable. In these cases, the learning task can be very sensitive to noises from the input, and additional regularization terms might be necessary.
\end{enumerate}
\subsection{Existing Operator Learning Architectures.}
Before proposing a suitable architectures for learning the inverse map $\eF^{-1}$ \eqref{eq:imap}, we briefly summarize existing operator learning architectures to examine whether they can be useful in this context. To this end, let $D \subset \R^{d_x}, U \subset \R^{d_u}$ and $\cX = \cX(D)$ and $\cY = \cY(U)$ be suitable function spaces. Then, a \define{DeepONet} \cite{lu2021learning} is the operator, $\cN^\don: \cX \to \cY$, given by
\begin{equation}
    \label{eq:donet1}
\cN^\don(\bar{u})(y) = \sum_{k=1}^p \beta_k(\bar{u}) \tau_k(y), 
\quad \bar{u} \in \cX, \; y \in U,
\end{equation}
where the \define{branch-net} $\bm{\beta}$ is a neural network that maps $\cE(\bar{u}) = (\bar{u}(x_1),\dots,\bar{u}(x_m)) \in \R^m$, evaluations of the input $\bar{u}$ at sensor points $\bm{x} := (x_1,\dots, x_m)\in D$, to $\R^p$:
\begin{align}
   \label{eq:branch}
\bm{\beta}: \R^m \to \R^p, ~\cE(\bar{u}) \mapsto (\beta_1(\cE(\bar{u})),\dots, \beta_p(\cE(\bar{u})),
\end{align}
and the \define{trunk-net} $\bm{\tau}(y) = (\tau_1(y),\dots, \tau_p(y))$ is another neural network mapping,
\begin{align}
  \label{eq:trunk}
\bm{\tau}: U \to \R^p, \quad y\mapsto (\tau_1(y),\dots, \tau_p(y)).
\end{align}
Thus, a DeepONet combines the branch net (as coefficient functions) and trunk net (as basis functions) to create a mapping between functions. 

On the other hand, a Fourier neural operator ({\bf FNO}) $\cN^\fno$ \citep{FNO} is a composition
\begin{equation}
\label{eq:fno}
\cN^\fno: \cX \mapsto \cY:\quad 
\cN^\fno = Q \circ \cL_T \circ \dots \circ \cL_1 \circ R.
\end{equation}
For simplicity let us define $\eM = Q \circ \cL_T \circ \dots \circ \cL_1$, and $\cN^\fno = \eM \circ R$, where $R: (x, \bar{u}) \mapsto R(\bar{u}(x),x)$ is a ``lifting operator'' represented by a linear transformation $R: \R^{\du\times d} \to \R^{d_v}$ where $\du$ is the number of components of the input function, $d$ is the dimension of the domain and $d_v$ is the ``lifting dimension'' (a hyperparameter). The operator $Q$ is a nonlinear projection, instantiated by a shallow neural network,  such that $v^{T+1}(x) \mapsto \cN^\fno(\bar{u})(x) = Q \left( v^{T+1}(x)\right)$. Each \emph{hidden layer} $\cL_\ell: v^\ell(x) \mapsto v^{\ell+1}(x)$ is of the form 
\[
v^{\ell+1}(x)  = \sigma\left(
W_\ell \cdot v^{\ell}(x) + b_\ell(x) + \left( K_\ell v^{\ell} \right) (x) 
\right),
\]
with $W_\ell \in \R^{d_v\times d_v}$ a weight matrix (residual connection), $b_\ell(x) \in \R^{d_v}$ a bias function, $\sigma$ an activation function, and the \emph{non-local Fourier layer},
\[
K_\ell v^\ell = \cF_N^{-1} \left(P_\ell(k) \cdot \cF_N v^\ell(k) \right),
\]
where $\cF_N v^\ell (k)$ denotes the (truncated)-Fourier coefficients of the discrete Fourier transform (DFT) of $v^\ell(x)$, computed based on the given $N$ grid values in each direction. Here, $P_\ell(k) \in \C^{d_v \times d_v}$ is a complex Fourier multiplication matrix indexed by $k\in \Z^d$, and $\cF_N^{-1}$ denotes the inverse DFT. 

Both operator learning frameworks (DeepONet and FNO) and their variants map \emph{functions to functions}. Hence, they cannot directly be used to learn the inverse map $\eF^{-1}$ \eqref{eq:imap}, which maps \emph{operators to functions}. Therefore, we need to modify and adapt these architectures to learn the inverse map. The following sections present our proposed approach. 
\subsection{A Motivating (Formal) Calculation.}
\label{sec:33}

We start by providing a heuristic motivation for our proposed architecture to learn the inverse map \eqref{eq:imap}. To this end and for definiteness, we consider the inverse  wave scattering problem for the Helmholtz equation \eqref{eq:helmholtz}, presented in section \ref{sec:helm}. Given the domain $D \subset \R^d$, we consider the following eigenvalue problem with Neumann boundary conditions,
\begin{equation}
\label{eq:nep}
\begin{aligned}
-\Delta \varphi_k &= \lambda_k\varphi_k, \quad \forall z \in D. \\
\frac{\partial \varphi_k}{\partial \nu}\big|_{\partial D} &=0, \quad \int\limits_D \varphi_k dz = 0. 
\end{aligned}
\end{equation}
By standard PDE theory \cite{Evansbook}, there exist eigenvalues $0 \leq \lambda_k \in \R$ for $k \in \N$, and the corresponding eigenfunctions $\{\varphi_k\}$ form an orthonormal basis for $L^2(D)$. We fix $K \in \N$ sufficiently large and without loss of generality, we assume $\omega =1$ in the Helmholtz equation~\eqref{eq:helmholtz} to consider the following Dirichlet boundary value problems,
\begin{equation}
        \label{eq:helm1}
        \begin{aligned}
           - \Delta u_k - a(z)u_k = 0, \quad &z\in D,~1 \leq k \leq K,\\
            u(z) = g_k(z), \quad &z\in \partial D,
        \end{aligned}
        \end{equation}
where $g_k = \varphi_k\big|_{\partial D}$. Using \eqref{eq:nep} and \eqref{eq:helm1}, we prove in {\bf SM} \ref{app:pf}, the following formal \emph{representation formula} for all $1 \leq k \leq K$,
\begin{equation}
\label{eq:arep}
\int\limits_{D} a u_k \varphi_k dz = \int\limits_D \lambda_k u_k \varphi_k dz - \int\limits_{\partial D} g_k \frac{\partial u_k}{\partial \nu} d\sigma(z).
\end{equation}
The formula \eqref{eq:arep} can be used to construct an approximation to the coefficient $a \in L^2(D)$ in the following manner. Writing $a \approx \sum\limits_{\ell =1}^K a_\ell \varphi_\ell$ (using the orthonormality of $\varphi$'s) for $K$ sufficiently large, we can evaluate the coefficients $a_{\ell}$ by solving the following Matrix equation for $A = \{a_\ell\}_{\ell=1}^K$, 
\begin{equation}
\label{eq:mat}
\begin{aligned}
{\bf C} A &= B, ~ {\bf C}_{k \ell} = \int\limits_{D}u_k \varphi_k \varphi_l dx, \quad \forall k,l, \\
B_k &= \int\limits_D \lambda_k u_k \varphi_k dz + \int\limits_{\partial D} g_k \frac{\partial u_k}{\partial \nu} d\sigma(z), \forall k.
\end{aligned}
\end{equation}
Further setting $\Psi_k = \frac{\partial u_k}{\partial \nu}$, we observe that the formal approximation of the coefficient $a$ relies on the following building blocks,
\begin{itemize}
\item {\bf Basis Construction:} The operations $\cB_k:z \mapsto \left(\varphi_k(z),\lambda_k\right)$, $1\leq k \leq K$, that form a basis. Note that they are independent of the coefficient $a$. 
\item {\bf PDE Solve:} The operation $\eE_k:(g_k,\Psi_k) \mapsto \big(\{u^k_j\}_{j=1}^K,\int\limits_{\partial D} g_k \Psi_k d\sigma(z)\big)$ that amounts to (approximately) inferring the coefficients of the solution $u_k$ of the Helmholtz equation \eqref{eq:helm1}, given the Dirichlet $g_k$ and Neumann $\Psi_k$ boundary values. A part of the right-hand side term $B_k$ is also appended to this operation. Once the coefficients $u^k_j$ are computed, the approximation $u_k$ to the solution of \eqref{eq:helm1} is readily computed in terms of the basis $\{\varphi_k\}$ by setting $u_k \approx \sum\limits_{j=1}^K u^k_j \varphi_j$.
\item {\bf Mode Mixing:} The previous two operations were restricted to individual modes, i.e., to each $k$, for $1 \leq k \leq K$. However, to construct the coefficients ${\bf C}_{kl}$ in \eqref{eq:mat}, we need to mix different modes. One way to do so is through multiplication. We denote this operation by
$\eM:\left(\{\varphi_k\}_{k=1}^K, \{u_k\}_{k=1}^K\right) \mapsto \left(\{u_k\varphi_k \varphi_\ell\}_{k,\ell=1}^K,\{\lambda_k u_k \varphi_k \}_{k=1}^K\right)$.
\item {\bf Matrix Inversion:} In the final step, we need to build the Matrix ${\bf C}$ in \eqref{eq:mat} and (approximately) invert it. This operation can be summarized by $\eI:  \left(\{u_k\varphi_k \varphi_\ell\}_{k,\ell=1}^K,\{\lambda_k u_k \varphi_k \}_{k=1}^K\right) \mapsto \sum_{j=1}^K a_j \varphi_j$, with $A = \{a_j\}$ being the solution of \eqref{eq:mat}.
\end{itemize}
% These building blocks, together with how they are combined to provide a representation formula for $a$, motivate us to propose the following ML architectures for learning the inverse map \eqref{eq:imap} from data,
\begin{figure*}[ht]
\begin{center}
{\includegraphics[width=0.75\textwidth]{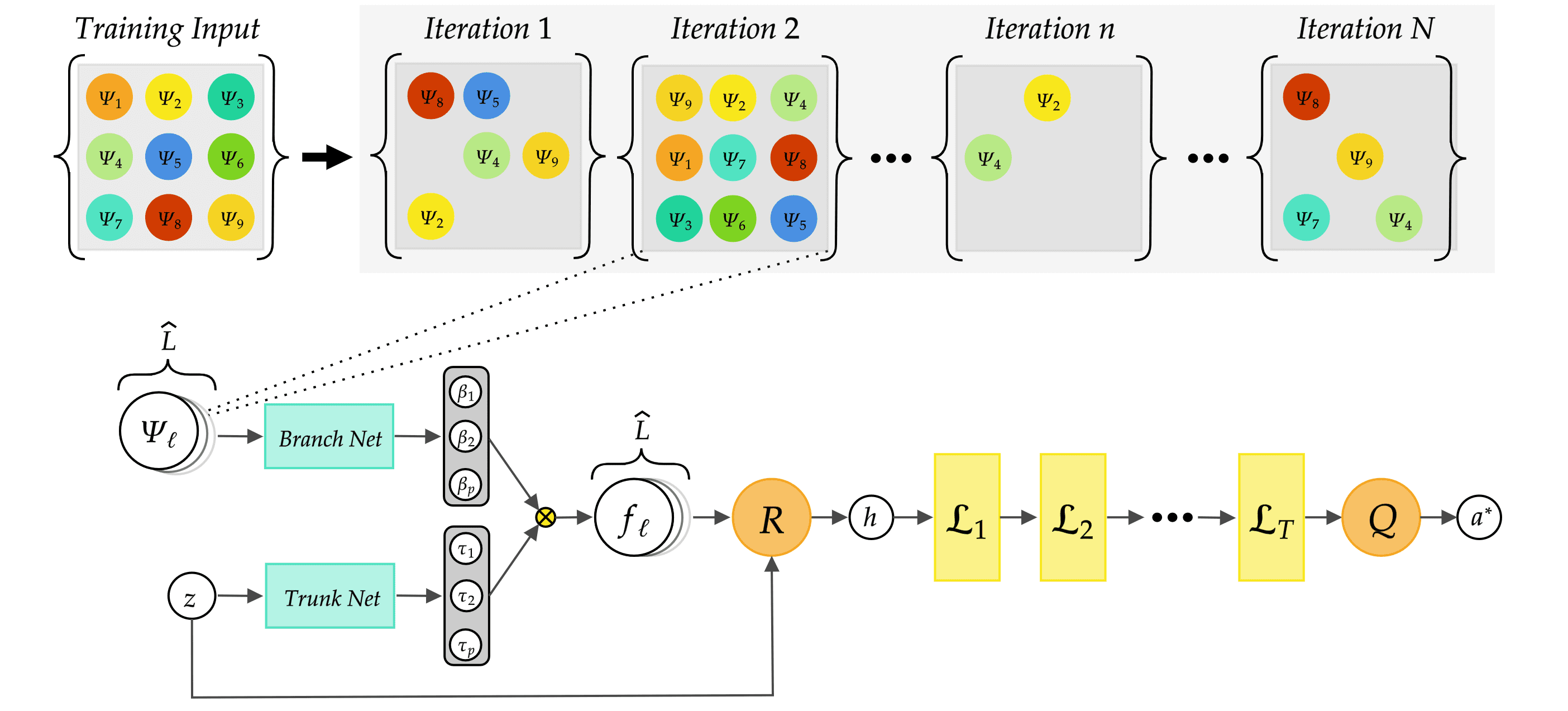}}
\caption{Schematic representation of Neural Inverse Operator  (NIO) architecture, with $R$ given by equation \eqref{eq:r2}, and Randomized Batching.}
\label{fig:ffDeepFNO}
\end{center}
\end{figure*}
\subsection{The Architecture.}
\label{sec:nio}
The formal approximation of the inverse map $\eF^{-1}$ \eqref{eq:imap} for the Helmholtz equation by formulas \eqref{eq:arep}-\eqref{eq:mat} cannot be directly used in practice as one cannot solve the PDE \eqref{eq:helm1} without knowing the coefficient $a$. However, the building blocks enumerated above motivate either an iterative fixed-point procedure or, in our case, a learning algorithm approximating $\eF^{-1}$ from data. To this end, we observe that the basis construction $z \mapsto \varphi_k(z)$ amounts to a particular instantiation of a trunk-net \eqref{eq:trunk} of a DeepONet. Similarly, the PDE solve map $\eE_k:(g_k,\Psi_k) \mapsto \{u^k_j\}_{j=1}^K$ is a particular instance of the application of a branch-net \eqref{eq:branch} of a DeepONet. Moreover, they can be combined in a DeepONet \eqref{eq:donet1} to approximate the solutions $u_k$ of the PDE \eqref{eq:helm1}. However, a DeepONet \eqref{eq:donet1} is linear in its trunk-net basis functions and thus  cannot represent the nonlinear mode mixing operator $\eM$. Instead, one can do so by passing the outputs of the DeepONet to the nonlinear layers of an FNO \eqref{eq:fno}, which also performs the final inversion operator $\eI$. 

These heuristic considerations are generalized to the abstract formalism of the inverse problem \eqref{eq:imap} and motivate us to propose the composition (stacking) of DeepONets and FNO to result in the following map, 
\begin{equation}
\label{eq:FCNIO}
\begin{aligned}
&\cN_{NIO}: \begin{pmatrix}
z \\
\{\Psi_\ell\}_{\ell=1}^L
\end{pmatrix}
\overset{\tau,\beta}\longmapsto 
\begin{pmatrix} 
\{\tau_k(z)\}_{k=1}^p \\
\{\beta_k\}_{k=1}^p
\end{pmatrix} \ldots \\
&\ldots \overset{\cN^{DON}}\longmapsto \{f_\ell(z)\}_{\ell=1}^L 
\overset{R}\longmapsto h(z) \overset{\eM}\longmapsto a^\ast(z),
\end{aligned}
\end{equation}
for approximating the abstract inverse map $\eF^{-1}$ \eqref{eq:imap}.
Here, the linear map $R$ can be explicitly defined as
\begin{equation}
    \label{eq:r1}
    \begin{aligned}
        h(z) = R(f_1,\dots,f_L, z) = &\frac{1}{L}\sum_{\ell=1}^L D_\ell f_\ell + Ez,
    \end{aligned}
\end{equation}
with $E, D_\ell \in \R^{d_v}$. In other words, the inputs $z \in D$ and $\Psi_\ell = \Lambda_a(g_\ell)$ \eqref{eq:obsop}, for $1 \leq \ell \leq L$, are fed into the trunk- and branch-nets of a DeepONet $\cN^{DON}$~\eqref{eq:donet1}, respectively, to create $L$ representations $\{f_{\ell}\}_{\ell = 1}^L$, defined in the interior of the underlying domain. These representations are first linearly transformed through $R$ yielding $h(z)$, and finally \emph{mixed} by the nonlinear component  $\eM$ of FNO resulting in an approximation of the underlying coefficient $a^{\ast}$. 
We observe that the DeepONet $\cN^{DON}$ in $\cN^{NIO}$ \eqref{eq:FCNIO} is flexible enough to handle inputs defined on the boundary, i.e., $\Psi_\ell$, and produce outputs $f_\ell$, defined on the interior of the underlying domain. 

It is important to note that the model takes only $z$ and $\{\Psi_\ell\}_{\ell=1}^L$ as input, rather than $z$ and the input-output pair $\{(g_\ell,\Psi_\ell)\}_{\ell=1}^L$. This choice is motivated by the following consideration: let us define $\mu_g$ as the underlying measure (distribution) on the boundary data $g$, where $g_\ell \sim \mu_g$, and $\mu_{\Psi} = \Lambda_a\#\mu_g$ represents the pushforward measure given the boundary operator $\Lambda_a$. Hence, $\Psi_\ell \sim \mu_{\Psi}$ for all $\ell=1,...,L$, with $\{\Psi_\ell\}_{\ell=1}^L$ representing the empirical distribution approximating $\mu_{\Psi}$. This, together with the injectivity of the boundary observation operator~\eqref{eq:obsop}, implies that $\{\Psi_\ell\}_{\ell=1}^L$ suffices to provide statistical information about the operator $\Lambda_a$ given $\mu_g$ satisfying certain properties.  
We denote $L$ as the number of samples discretizing the measure $\mu_\Psi$ or, with abuse of notation, the operator $\Lambda_a$.
%provide to the model the output $\Psi_\ell = \Lambda_a(g_\ell)$ rather than the input-output pair $(g_\ell,\Psi_\ell)$ as this provides the empirical measure approximation to the underlying push-forward measure. 

The construction above addresses point (1) in Section~\ref{sec:challenges}.
However, it does not necessarily satisfy point (2) in its current form. For the model to effectively process $\mu_\Psi$ as the input rather than a particular discrete realization of it, the following desirable properties should be met:
\begin{itemize}
\item The architecture should exhibit invariance under permutations of the input measurements $\{\Psi_\ell\}_{\ell=1}^L$ as they are i.i.d.~samples of $\mu_\Psi$. 
\item The learning framework must be able to handle an empirical measure of the distribution $\mu_\Psi$ with an arbitrary sample size $\Tilde{L}$. In particular, the input sample size at the training and testing stages could be different. 
\item The performance of the model should be independent of the sample size $\Tilde{L}$ used to discretize $\mu_\Psi$.  
\end{itemize}

To address the first two points, we can modify the architecture  by redefining the linear transformation $R$ as follows:
\begin{equation}
\label{eq:r2}
h(z) = R(f_1,\dots,f_L, z) = \frac{D}{L}\sum_{\ell=1}^L  f_\ell + Ez, \quad E, D\in \R^{d_v}
\end{equation}
To ensure the performance is independent of $\Tilde{L}$, a naive and inefficient approach would involve constructing all possible permutations of $\{\Psi_\ell\}_{\ell=1}^L$ with size $\hat{L}$ for all $\hat{L}=2,\ldots, L$ and providing them as input to the model. However, inspired by the well-known \textit{bagging} algorithm widely used in machine learning, we propose an efficient and novel method to tackle this point which we term \textit{randomized batching}. In this approach, given the sequence $\{\Psi_\ell\}_{\ell=1}^L$, during each training iteration, an integer number $\hat{L}$ is randomly drawn from the set $\{2,\ldots,L\}$. Then, $\hat{L}$ samples are randomly picked from $\{\Psi_\ell\}_{\ell=1}^L$ and the new sequence $\{\Psi_k\}_{k=1}^{\hat{L}}$ are fed into the model at each iteration during training. 

We refer to the architecture \eqref{eq:FCNIO} with the linear transformation $R$ given in \eqref{eq:r2}, and incorporating the \textit{randomized batching} algorithm, as \emph{Neural Inverse Operator} (see Figure \ref{fig:ffDeepFNO} for a schematic representation of the architecture).

\section{Empirical Results.}  We empirically test NIO on benchmark PDE inverse problems below. The exact details of the training, as well as the architecture and hyperparameter choices, are presented in {\bf SM} \ref{app:tadet}. 

As baselines in the following experiments, we choose two models. First, we consider a DeepONet with a CNN as the branch net in~\eqref{eq:donet1}, which performs mixing of the input function within the branch itself
\begin{equation}
\label{eq:DON}
\begin{aligned}
&\cN_{DONet}: \begin{pmatrix}
z \\
\{\Psi_\ell\}_{\ell=1}^L
\end{pmatrix}
\overset{\tau,\beta_{con}}\longmapsto 
\begin{pmatrix} 
\{\tau_k(z)\}_{k=1}^p \\
\{\beta_k\}_{k=1}^p
\end{pmatrix} \overset{\cN^{DON}} \longmapsto a^\ast(z).
\end{aligned}
\end{equation}
Second, we consider a fully convolutional image-to-image neural network architecture (details in {\bf SM} \ref{app:numex}). A variant of this architecture was already used in seismic imaging (full waveform inversion) in \cite{deng2021openfwi}. We have extended this architecture significantly to apply it to the abstract inverse problem~\eqref{eq:imap}. 
\subsection{Calder\'{o}n Problem for EIT.} We start with the Calder\'{o}n problem for the elliptic equation \eqref{eq:calderon} on the computational domain $D=[0,1]^2$, with source $s = 0$. The training (and test) data are generated by sampling from a probability distribution on the conductivity coefficient $a$. Once a sample conductivity is drawn, a set of Dirichlet boundary conditions $\{g_{\ell}\}_{\ell=1}^{L}$ are drawn from a probability distribution on the boundary values. For each $g_{\ell}$, the underlying elliptic equation is solved numerically with a standard five-point finite difference scheme and the current, $\Psi_\ell = a\frac{\partial u}{\partial \nu}$, is evaluated on the boundary. We choose the boundary data $g_\ell$, for $1 \leq \ell \leq L=20$ as the boundary values of $\cos(\omega(x\cos(\theta_\ell) + y\sin(\theta_\ell)))$, with $\theta_\ell = \frac{2\pi\ell}{20}$. For the coefficient $a$, we sample from {\bf trigonometric functions} by setting $a(x, y) = \exp\big(\sum_{k=1}^{m}c_{k}\sin(k\pi x)\sin(k\pi y) \big),$ with $m=\mod(\bar{m})$ where $\bar{m}\sim \mathcal{U}([1,5])$ and $\{c_k\}\sim \mathcal{U}([-1,1]^{m})$.
%or from {\bf piecewise-constant functions} by setting $a(x_1, x_2) =1 + \sum_{k=1}^{m}  d_k \tanh\left(200(x_2 - s_k) \right)$, with $m = \mod(\bar{m})$, $\bar{m} \sim \mathcal{U}([1,4])$, $\{d_k\}\sim \mathcal{U}([0,0.5]^{m})$ and $
%\{s_k\}\sim \mathcal{U}([0,1]^{m})$. 
All the models are trained with $4096$ training coefficient samples, and the relative (percentage) test errors (with respect to $2048$ test samples) in both $L^1$ and $L^2$ norms for NIO (and the baselines) are presented in Table~\ref{tab:res}. As the table shows, NIO is the best-performing model, outperforming the next-best FCNN model by almost halving the errors. Moreover, the total errors are very small ($<1\%$) with NIO. 
%Moreover, as expected, when tested on different discretization of the boundary input operator based on $\Tilde{L}$ measurements instead of $L$ (as for training), NIO results are very robust, as shown in \ref{fig:diffL_main}, whereas the baselines' generalization error increases by almost 30 times (see \textbf{SM} for additional details). 

\begin{figure*}[htbp]
    \begin{subfigure}{0.25\textwidth}
        \centering
        \includegraphics[width=1\linewidth]{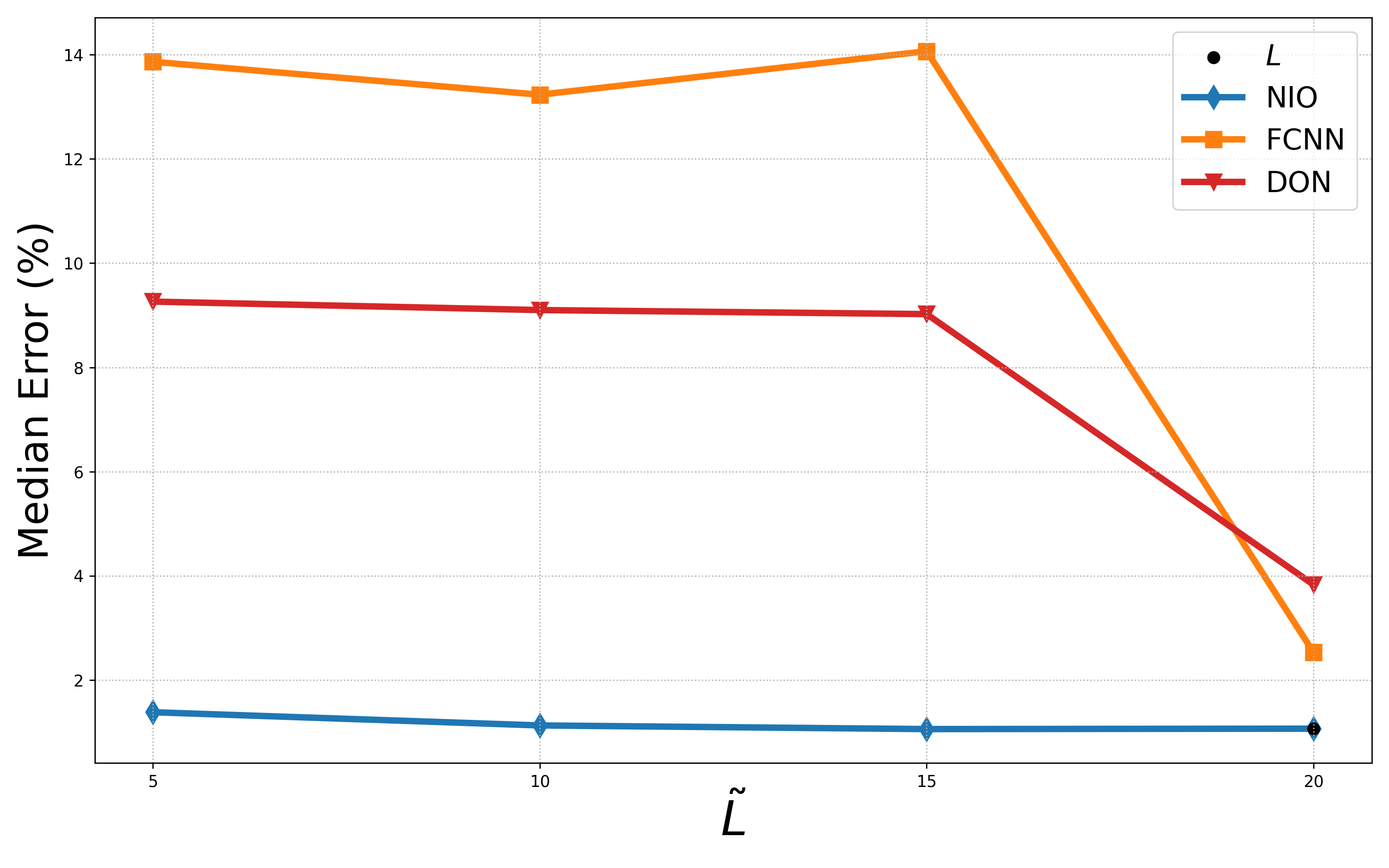}
         \label{fig:diffL_main:helm}
    \end{subfigure}
    \begin{subfigure}{0.25\textwidth}
        \centering
        \includegraphics[width=1\linewidth]{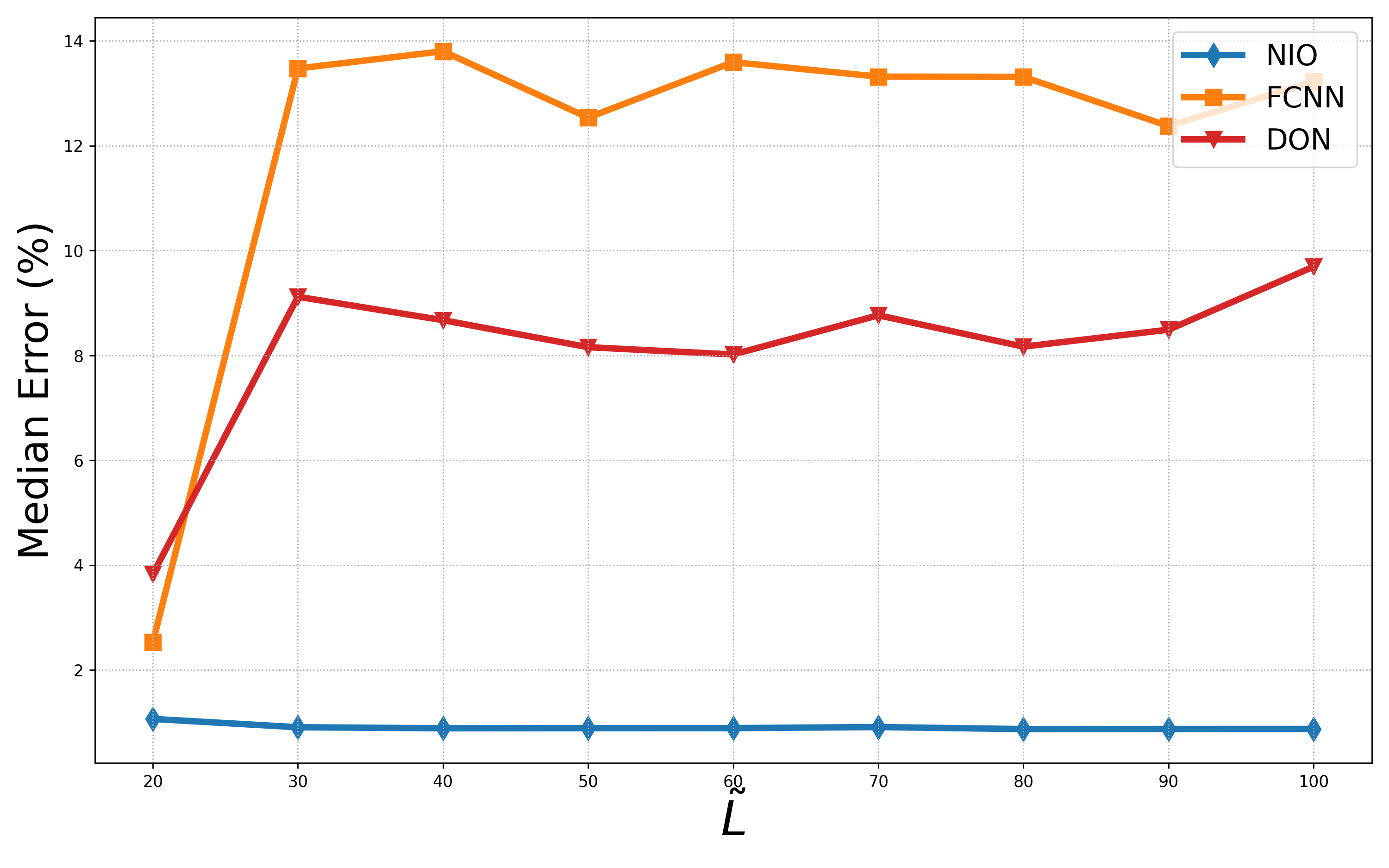}
        \label{fig:diffL_main:helm2}
    \end{subfigure}
    \begin{subfigure}{0.5\textwidth}
        \centering
        \includegraphics[width=1\linewidth]{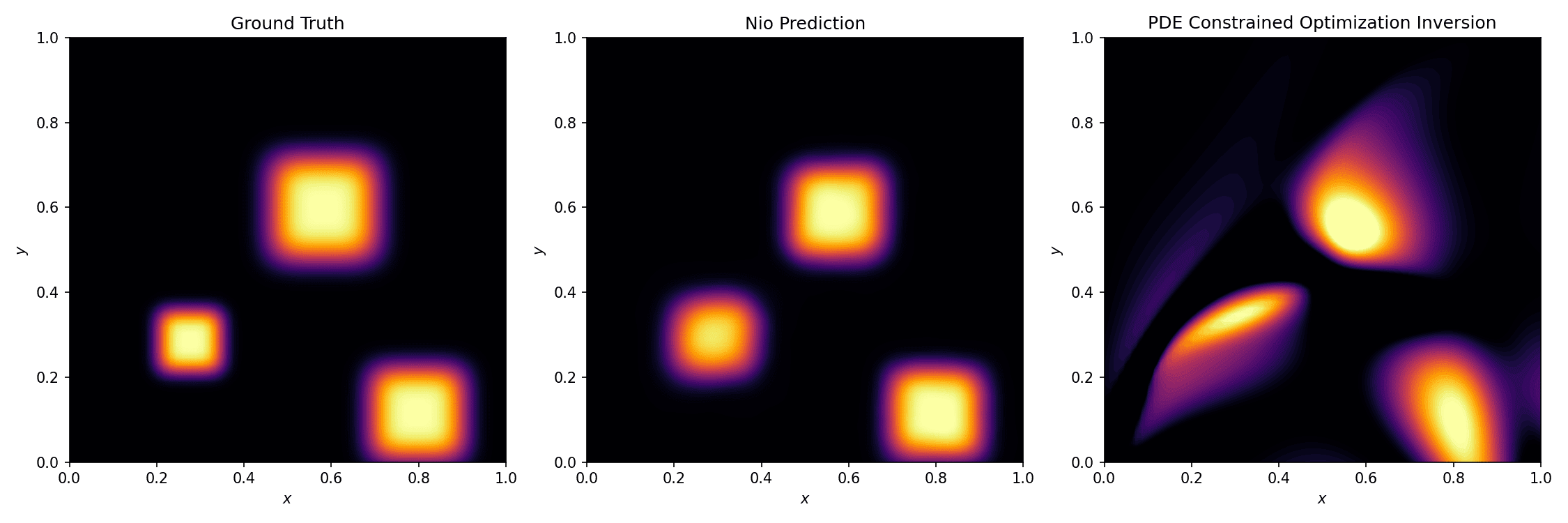}
        \label{fig:comphelm}
    \end{subfigure}
\vspace{-1.2cm}\caption{Inverse Wave Scttering. Left Panel: testing median $L^1$-error VS number of input samples $\Tilde{L}$ (left: $\Tilde{L}<L$, right: $\Tilde{L}>L$). The black dot represents the number of training samples $L$. Right Panel: ground truth (left), NIO reconstruction (middle), and reconstruction with the PDE constrained optimization method (right)  for an out-of-distribution test sample }
\label{fig:diffL_main}
\end{figure*}

As a second experiment for EIT, we consider a more practical example suggested in \cite{MSbook}, where the authors model the EIT imaging of the heart and lungs of a patient using electrodes on the body. This \emph{discontinuous heart and lungs phantom} is depicted in {\bf SM} Figure \ref{fig:heart_sample}. The underlying domain is the unit circle, and the elliptic equation \eqref{eq:calderon} is solved with a standard finite element scheme. The boundary conditions are given by $g_\ell(\theta) = \frac{1}{2\pi}\exp(i2\pi\theta f_\ell)$, with $\ell=1,\ldots,32$ and $f=[-16,\ldots, -1, 1, 14, 15,16]$. The coefficient $a$ is modelled by adding $8\%$ white noise to the location, shape, and conductivity of the configuration of heart and lungs shown in {\bf SM} Figure \ref{fig:heart_sample}. The input of the learning operators is obtained by computing the Fourier transform at frequencies $f$ of the difference between the Neumann trace of the PDE solution with the coefficient $a$ and the one with the unit coefficient $a=1$. Again, the results presented in Table \ref{tab:res} and \ref{fig:diffL_main} show that NIO is the best-performing model and yields very low reconstruction errors, solving this practical problem with high accuracy. In contrast, a traditional direct method such as the D-bar method~\cite{MSbook} has a larger error of $8.75\%$  for this numerical inversion test (see {\bf SM} Section \ref{sec:comparison_num} for details). %concerning comparison of the proposed approach with D-bar and PDE-constrained optimization method).

\subsection{Inverse Wave Scattering.}
\label{sec:helm_exp}
In this problem, the Helmholtz equation \eqref{eq:helmholtz} is considered on the domain $D =[0,1]^2$, and the task is to learn coefficients sampled from a distribution,
    $a(x, y) = \sum_{k=1}^{m} \exp\big(-c(x-c_{1,k} )^4 - c(y-c_{2,k})^4\big),$
with $c = 2\times 10^4/3$. It
represents a homogeneous medium with square-shaped inclusions, randomly spread in the domain (see {\bf SM} Figure \ref{fig:step_sample}). Here,  $m = \mod(\bar{m})$, $\bar{m}\sim \mathcal{U}([1,4])$ and $\{(c_{1,k}, c_{2,k})\} \sim \mathcal{U}([0,1]^{m\times 2})$. For each draw of the coefficient, $20$ Dirichlet boundary values are prescribed, exactly as in the EIT experiment with trigonometric coefficients. The corresponding (approximate) solutions of the Helmholtz equation \eqref{eq:helmholtz} are computed with a central finite difference scheme, and the Neumann trace is evaluated to represent the DtN map. We train the models with $4096$ training samples and present the relative (median) test errors, on a test set of $1600$ samples, in Table \ref{tab:res}. Again, NIO is the best-performing model, beating the next-best FCNN by a significant margin.
\subsection{Radiative Transport Equation and Optical Imaging.} 
Next, we consider the radiative transport equation~\eqref{eq:RTE}  in the domain $X\times V$, where $X=[0,1]$ and $V = [-1,1]$, with $\varepsilon=1$. Consequently, $\Gamma_{-}  = \{(0,v), v\in[0,1] \}\cup \{(1,v), v\in[-1,0] \}$. The task is to infer the absorption and scattering coefficients from the Albedo operator \eqref{eq:albop}. To this end, we fix $k(v,v')=1$, $\sigma_a = 1-a$ in \eqref{eq:RTE} and draw the absorption coefficient $a$ from the distribution, $a(x) = c \chi_{[-1/2,1/2]}(rx - x_0) + 1$, with $\chi$ denoting the characteristic function
and with $c\sim \mathcal{U}([0.5,1])$, $x_0\sim \mathcal{U}([0,1])$ and $r\sim \mathcal{U}([0,0.8])$.  Once the coefficient is drawn, boundary conditions on the inflow boundary $\Gamma_-$ are imposed by setting $\phi_\ell(0,v) = \exp{\big(-200\left(v - v_\ell\right)^2 \big)}$,  and $\phi_\ell(1,v) = 0$, if $v_\ell>0$, and $\phi_\ell(0,v) = 0$,  and $\phi_\ell(1,v) = \exp{\big(-200\left(v - v_\ell\right)^2 \big)}$, if $v_\ell<0$, with $v_\ell$ being the $\ell$-th quadrature point used to approximate the integral term in \eqref{eq:RTE}, $1 \leq \ell \leq 32$. Then, the radiative transport equation is approximated with a finite-element method, and the resulting solution $u_\ell$ is evaluated at the outflow boundary $\Gamma_+$ as the output of the Albedo operator~\eqref{eq:albop}. All the models are trained on $4096$ training samples, and the relative median test errors on a test set of $2048$ samples are presented in Table \ref{tab:res}, demonstrating that NIO significantly outperforms both FCNN and DeepONet, also in this case, resulting in low test errors even for the underlying discontinuous absorption coefficient.
\subsection{Seismic Imaging.}
In the final test, we model seismic imaging by considering the acoustic wave equation \eqref{eq:wave} in the space-time domain $[0,1]^2 \times [0,T]$ and the task at hand is to learn the underlying squared-slowness coefficients $a$ from the \emph{source-to-receiver} map \eqref{eq:str}. To this end, we choose two types of coefficients from \cite{deng2021openfwi}, the so-called \emph{Style-A} and \emph{CurveVel-A} datasets. For each medium, waves are generated at source locations  $(x_{s_\ell,0})$ on the vertical boundary, for $\ell=1,\ldots,5$. The corresponding acoustic wave equation is solved with a finite difference scheme, and the temporal data is recorded at receivers on the vertical boundary. We follow~\cite{deng2021openfwi} and train all the models with $55000$ and $22000$  training samples for the \emph{Style-A} and \emph{CurveVel-A} datasets, respectively, and present the resulting (median) relative test errors, on a test set of $7000$ and $6000$ samples, in Table~\ref{tab:res}. We observe from the table that even for this problem, NIO is either outperforming or on par with FCNN. This is particularly noteworthy as the FCNN architecture was demonstrated to be one of the states of the art on this problem in \cite{deng2021openfwi} among several machine learning models. 

%In {\bf SM} \ref{app:addnum}, we present additional empirical results illustrating the performance of NIO as well as demonstrating that its solution of the PDE inverse solutions is robust to noise at the time of inference.  
\subsection{Robustness and Computational Efficiency of NIO.}
In the {\bf SM} \ref{app:addnum}, we provide further details on the performance of NIO and its comparison to the considered baselines in terms of several factors. Here, we present a succinct summary of these results. First, inverse problems are characterized by sensitivity to perturbations of different types. Hence, it is imperative to test the robustness of the proposed NIO framework to these factors. To this end, we start with adding noise to the inputs of each of the benchmarks at test time and present the resulting test errors in {\bf SM} Table \ref{tab:res_noise} to observe that NIO (as well as the baselines) are not sensitive to this noise. 
Next, in {\bf SM} Table \ref{tab:res_res}, we plot the test errors for the models when the underlying grid resolution is varied at test time to observe that NIO is not sensitive to the input grid resolution. Finally, in {\bf SM} Table \ref{tab:res_randomloc}, we present errors when the locations of input sensors are chosen randomly at test time, in contrast to sensors located on a uniform grid at training, to see that NIO is also robust to this variation. However, these factors are less salient in this context than the discretization of the input boundary observation operator. The key element of this discretization is the number of input samples. As argued in the previous section, NIO has been constructed to be \emph{independent} of the number of samples that discretize the boundary observation operator. We test this behavior for the inverse wave scattering problem (see {\bf SM} \ref{sec:rob_disc} for other benchmarks) and present the errors in Figure \ref{fig:diffL_main}, where the number of input samples at the testing stage are decreased (left) and increased (right) at test time. We clearly see that NIO is very robust to this fluctuation, completely contrasting to the baselines whose performance deteriorates, and the test errors increase by almost an order of magnitude. This demonstrates that NIO genuinely learns the underlying inverse map rather than just a discrete representation of it. Finally, in {\bf SM} \ref{app:ood}, it is shown that NIO \emph{generalizes} very well \emph{out-of-distribution} for a variety of problems, outperforming baselines considerably. 

We conclude this section by comparing NIO with the well-established PDE-constrained optimization algorithms. To this end, we consider an out-of-distribution example for the inverse wave scattering benchmark and compare the results of NIO and PDE-constrained optimization (see details in {\bf SM} \ref{sec:comparison_num}) in Figure \ref{fig:diffL_main} (Right) to observe that NIO is much more accurate in reconstructing the scatterers than PDE-constrained optimization. The $L^1$-test errors are $2.3\%$ for NIO and $11.1\%$ for PDE-constrained optimization. Moreover, it took less than $1$ sec of inference time for NIO (on a CPU) compared to a single GPU training time of $8.5$ hours for the PDE-constrained optimization problem. Thus, this experiment demonstrates the real efficiency of NIO as it is $5$ times more accurate while being $4$ orders of magnitude faster than the PDE-constrained optimization algorithm.  

\section{Related Work.}
Existing methods for solving the class of PDE inverse problems considered here include the so-called \emph{Direct Methods}, such as the D-bar method \cite{isaacson2004reconstructions} for EIT and the so-called Imaging condition \cite{claerbout1985imaging} for seismic inversion. Iterative methods approximating fixed points \cite{bakushinsky2005iterative} are also used, including the very popular gradient-based PDE-constrained optimization methods \cite{chavent2010nonlinear}. Another approach falls in the category of Bayesian formulation of inverse problems \cite{tarantola2005inverse,stuart2010inverse}, which also quantifies the uncertainty of the solution. Finally, directly learning some PDE inverse operators from data has been considered in \cite{maarten2022deep} and references therein.

\begin{table}[ht]
\caption{Relative median $L^1$-error and $L^2$-error computed over testing samples for different benchmarks and models with the best performing model highlighted in bold. }
\label{tab:res}
\begin{center}
\begin{small}
%\begin{sc}
\resizebox{\columnwidth}{!}{
  \begin{tabular}{ l l l l l l l l l l l}
    \toprule
    \multirow{3}{*}{} &
    \multicolumn{2}{c}{\bfseries DONet} &
      \multicolumn{2}{c}{\bfseries FCNN} &
      \multicolumn{2}{c}{\bfseries NIO}  \\
      \midrule
    & $L^1\downarrow$ & $L^2\downarrow$ & 
    $L^1\downarrow$ & $L^2\downarrow$ &  
    $L^1\downarrow$ & $L^2\downarrow$   \\
    \midrule\midrule
\bfseries\makecell{Calder\'{o}n Problem \\ Trigonometric}& \makecell{1.97\%}& \makecell{2.36\%}& \makecell{1.49\%}& \makecell{1.82\%}& \bf\makecell{0.85\%}& \bf\makecell{1.05\%}\\ \midrule

\bfseries\makecell{Calder\'{o}n Problem \\ 
Heart\&Lungs}& \makecell{0.95\%}& \makecell{3.69\%}& \makecell{0.27\%}& \makecell{1.62\%}& \bf\makecell{0.18\%}& \bf\makecell{1.16\%}\\ \midrule

\bfseries\makecell{Inverse Wave Scattering }& \makecell{3.83\%}& \makecell{7.41\%}& \makecell{2.53\%}& \makecell{7.55\%}& \bf\makecell{1.07\%}& \bf\makecell{2.94\%}\\ \midrule

\bfseries\makecell {Radiative transport} & \makecell{2.35\%}& \makecell{4.35\%}& \makecell{1.46\%}& \makecell{3.71\%}& \bf\makecell{1.1\%}& \bf\makecell{2.9\%}\\ \midrule

\bfseries\makecell {Seismic Imaging \\ CurveVel - A}& \makecell{3.98\%}& \makecell{5.86\%}& \bf\makecell{2.65\%}& \makecell{5.05\%}& \makecell{2.71\%}& \bf\makecell{4.71\%}\\ \midrule

\bfseries\makecell{Seismic Imaging \\ Style - A}& \makecell{3.82\%}& \makecell{5.17\%}& \makecell{3.12\%}& \makecell{4.63\%}& \bf\makecell{3.04\%}& \bf\makecell{4.36\%}\\ 

    \bottomrule
  \end{tabular}}
  \end{small}
\end{center}
\end{table}

\section{Discussion.}
For PDEs, written in the abstract form \eqref{eq:pde}, we consider a large class of inverse problems that are only well-defined when the underlying inverse operator \eqref{eq:imap}, maps an operator (the boundary observation operator \eqref{eq:obsop}) to the underlying coefficient (a function). The resulting inverse problem amounts to inferring the unknown coefficient $a$ from data pairs $(\Lambda_a, \eF^{-1}(\Lambda_a))$ representing the observation operator. Existing operator learning frameworks such as DeepONets \eqref{eq:donet1} and FNOs \eqref{eq:fno} only map functions to functions. Hence, one needs to adapt them to be able to learn \emph{mappings between operators and functions} in order to solve the inverse problem \eqref{eq:imap}. To this end, we have proposed a novel architecture, termed \emph{Neural Inverse Operators} (NIO), based on a composition of DeepONets and FNOs, augmented with suitable architectural priors (definition of $R$ in \eqref{eq:r2}), and trained with \textit{randomized batching}, to guarantee invariance of the generalization error to the different discretization of the input operator. Our architecture is motivated by the underlying structure of the inverse map. We tested the NIO on a variety of benchmark inverse problems. These include the Calder\'on Problem in electrical impedance tomography, inverse wave scattering modelled with the Helmholtz equation, optical imaging using the radiative transport equation, and seismic imaging with the acoustic wave equation. For all these problems, NIO outperformed baselines significantly and provided accurate and, more importantly, robust approximations to the unknown coefficients with small errors (see {\bf SM} \ref{app:addnum}). Finally, a series of experiments were also presented to demonstrate that NIO is robust with respect to various factors such as varying sensor locations, grid resolutions, noise, and discretizations of the input operator while being able to generalize \emph{out-of-distribution} and being more accurate and much faster than direct and PDE constrained optimization algorithms. As this is the first paper where an end-to-end machine learning framework is proposed for learning maps between operators and functions, various extensions are possible. For instance, other architectures, such as recently proposed  LOCA~\cite{loca}, VIDON~\cite{vidon}, or graph-based approaches \cite{bengioneural,well}, can be adapted in this context. Problems in higher-dimensional (particularly with seismic) imaging need to be considered to explore how NIOs scale with increasing problem size. Finally, approximation bounds and universality results, in the spirit of \cite{LMK1,kovachki2021universal} need to be derived in order to place NIOs on a solid theoretical footing. 

\section*{Acknowledgements}
We would like to express our deepest gratitude to Nicholas Nelsen and Tobias Rohner for their constructive feedbacks.

% One major challenge in inverse problems is (poor) stability. However, our tests showed that the learned NIOs are very robust to both large noise in the observed data and varying grid sizes in PDE solvers, which is a source of modelling error (see {\bf SM} Tables \ref{tab:res_noise} and \ref{tab:res_res}). Therefore, we have provided a unifying machine learning framework that can accurately solve a large class of PDE inverse problems arising in applications in science and engineering. The most attractive aspect of NIO is the extremely low computational cost at inference compared to traditional PDE-constrained optimization methods while still retaining satisfactory accuracy levels. As an example, the inference time for the inverse wave scattering with NIO is $0.5$ secs. On the other hand, obtaining similar accuracy with a PDE-constrained optimization approach will require parallelization and approximately $20000$ Helmholtz PDE solves of $2$ secs each, \emph{making NIO at least three orders of magnitude faster} ({\bf SM} Section \ref{sec:comparison_num}). 

\newpage
\bibliography{bibliography}

\begin{thebibliography}{35}
\providecommand{\natexlab}[1]{#1}
\providecommand{\url}[1]{\texttt{#1}}
\expandafter\ifx\csname urlstyle\endcsname\relax
  \providecommand{\doi}[1]{doi: #1}\else
  \providecommand{\doi}{doi: \begingroup \urlstyle{rm}\Url}\fi

\bibitem[Bakushinsky \& Kokurin(2005)Bakushinsky and
  Kokurin]{bakushinsky2005iterative}
Bakushinsky, A.~B. and Kokurin, M.~Y.
\newblock \emph{Iterative methods for approximate solution of inverse
  problems}, volume 577.
\newblock Springer Science \& Business Media, 2005.

\bibitem[Bal \& Jollivet(2008)Bal and Jollivet]{bal2008stability}
Bal, G. and Jollivet, A.
\newblock Stability estimates in stationary inverse transport.
\newblock \emph{Inverse Problems \& Imaging}, 2\penalty0 (4):\penalty0 427,
  2008.

\bibitem[Boussif et~al.(2022)Boussif, Assouline, Benabbou, and
  Bengio]{bengioneural}
Boussif, O., Assouline, D., Benabbou, L., and Bengio, Y.
\newblock {MAgNet: Mesh Agnostic Neural PDE solver}.
\newblock \emph{arXiv preprint arXiv:2210.05495}, 2022.

\bibitem[Brandstetter et~al.(2022)Brandstetter, Worrall, and Welling]{well}
Brandstetter, J., Worrall, D.~E., and Welling, M.
\newblock {Message Passsing Neural PDE solvers}.
\newblock \emph{arXiv preprint arXiv:2202.03376}, 2022.

\bibitem[Caday et~al.(2019)Caday, de~Hoop, Katsnelson, and
  Uhlmann]{caday2019scattering}
Caday, P., de~Hoop, M.~V., Katsnelson, V., and Uhlmann, G.
\newblock Scattering control for the wave equation with unknown wave speed.
\newblock \emph{Archive for Rational Mechanics and Analysis}, 231\penalty0
  (1):\penalty0 409--464, 2019.

\bibitem[Cai et~al.(2021)Cai, Wang, Lu, Zaki, and Karniadakis]{donet2}
Cai, S., Wang, Z., Lu, L., Zaki, T.~A., and Karniadakis, G.~E.
\newblock {DeepM\&Mnet: Inferring the electroconvection multiphysics fields
  based on operator approximation by neural networks}.
\newblock \emph{Journal of Computational Physics}, 436:\penalty0 110296, 2021.

\bibitem[Chavent(2010)]{chavent2010nonlinear}
Chavent, G.
\newblock \emph{Nonlinear least squares for inverse problems: theoretical
  foundations and step-by-step guide for applications}.
\newblock Springer Science \& Business Media, 2010.

\bibitem[Claerbout(1985)]{claerbout1985imaging}
Claerbout, J.~F.
\newblock \emph{Imaging the earth's interior}, volume~1.
\newblock Blackwell scientific publications Oxford, 1985.

\bibitem[Clop et~al.(2010)Clop, Faraco, and Ruiz]{clop2010stability}
Clop, A., Faraco, D., and Ruiz, A.
\newblock {Stability of Calder{\'o}n's inverse conductivity problem in the
  plane for discontinuous conductivities}.
\newblock \emph{Inverse Problems \& Imaging}, 4\penalty0 (1):\penalty0 49,
  2010.

\bibitem[Deng et~al.(2021)Deng, Feng, Wang, Zhang, Jin, Feng, Zeng, Chen, and
  Lin]{deng2021openfwi}
Deng, C., Feng, S., Wang, H., Zhang, X., Jin, P., Feng, Y., Zeng, Q., Chen, Y.,
  and Lin, Y.
\newblock {OpenFWI: Large-Scale Multi-Structural Benchmark Datasets for Seismic
  Full Waveform Inversion}.
\newblock \emph{arXiv preprint arXiv:2111.02926}, 2021.

\bibitem[Evans(2010)]{Evansbook}
Evans, L.~C.
\newblock \emph{Partial differential equations}, volume~19.
\newblock American Mathematical Soc., 2010.

\bibitem[Goodfellow et~al.(2016)Goodfellow, Bengio, and Courville]{DLbook}
Goodfellow, I., Bengio, Y., and Courville, A.
\newblock \emph{Deep learning}.
\newblock MIT press, 2016.

\bibitem[Isaacson et~al.(2004)Isaacson, Mueller, Newell, and
  Siltanen]{isaacson2004reconstructions}
Isaacson, D., Mueller, J.~L., Newell, J.~C., and Siltanen, S.
\newblock {Reconstructions of chest phantoms by the D-bar method for electrical
  impedance tomography}.
\newblock \emph{IEEE Transactions on medical imaging}, 23\penalty0
  (7):\penalty0 821--828, 2004.

\bibitem[Isakov(2017)]{INVbook}
Isakov, V.
\newblock \emph{Inverse Problems for Partial Differential Equations}.
\newblock Springer, 2017.

\bibitem[Kissas et~al.(2022)Kissas, Seidman, Guilhoto, Preciado, Pappas, and
  Perdikaris]{loca}
Kissas, G., Seidman, J.~H., Guilhoto, L.~F., Preciado, V.~M., Pappas, G.~J.,
  and Perdikaris, P.
\newblock Learning operators with coupled attention.
\newblock \emph{Journal of Machine Learning Research}, 23\penalty0
  (215):\penalty0 1--63, 2022.

\bibitem[Kovachki et~al.(2021{\natexlab{a}})Kovachki, Lanthaler, and
  Mishra]{kovachki2021universal}
Kovachki, N., Lanthaler, S., and Mishra, S.
\newblock {On universal approximation and error bounds for Fourier Neural
  Operators}.
\newblock \emph{Journal of Machine Learning Research}, 22:\penalty0 Art--No,
  2021{\natexlab{a}}.

\bibitem[Kovachki et~al.(2021{\natexlab{b}})Kovachki, Li, Liu, Azizzadensheli,
  Bhattacharya, Stuart, and Anandkumar]{NO}
Kovachki, N., Li, Z., Liu, B., Azizzadensheli, K., Bhattacharya, K., Stuart,
  A., and Anandkumar, A.
\newblock Neural operator: Learning maps between function spaces.
\newblock \emph{arXiv preprint arXiv:2108.08481v3}, 2021{\natexlab{b}}.

\bibitem[Lai et~al.(2019)Lai, Li, and Uhlmann]{lai2019inverse}
Lai, R.-Y., Li, Q., and Uhlmann, G.
\newblock Inverse problems for the stationary transport equation in the
  diffusion scaling.
\newblock \emph{SIAM Journal on Applied Mathematics}, 79\penalty0 (6):\penalty0
  2340--2358, 2019.

\bibitem[Lanthaler et~al.(2022)Lanthaler, Mishra, and Karniadakis]{LMK1}
Lanthaler, S., Mishra, S., and Karniadakis, G.~E.
\newblock Error estimates for {D}eep{ON}ets: {A} deep learning framework in
  infinite dimensions.
\newblock \emph{Transactions of Mathematics and Its Applications}, 6\penalty0
  (1):\penalty0 tnac001, 2022.

\bibitem[Li et~al.(2021{\natexlab{a}})Li, Kovachki, Azizzadenesheli, Liu,
  Bhattacharya, Stuart, and Anandkumar]{FNO}
Li, Z., Kovachki, N.~B., Azizzadenesheli, K., Liu, B., Bhattacharya, K.,
  Stuart, A., and Anandkumar, A.
\newblock Fourier neural operator for parametric partial differential
  equations.
\newblock In \emph{International Conference on Learning Representations},
  2021{\natexlab{a}}.

\bibitem[Li et~al.(2021{\natexlab{b}})Li, Zheng, Kovachki, Jin, Chen, Liu,
  Azizzadenesheli, and Anandkumar]{FNO1}
Li, Z., Zheng, H., Kovachki, N., Jin, D., Chen, H., Liu, B., Azizzadenesheli,
  K., and Anandkumar, A.
\newblock Physics-informed neural operator for learning partial differential
  equations.
\newblock \emph{arXiv preprint arXiv:2111.03794}, 2021{\natexlab{b}}.

\bibitem[Liu \& Oksanen(2016)Liu and Oksanen]{liu2016lipschitz}
Liu, S. and Oksanen, L.
\newblock {A Lipschitz stable reconstruction formula for the inverse problem
  for the wave equation}.
\newblock \emph{Transactions of the American Mathematical Society},
  368\penalty0 (1):\penalty0 319--335, 2016.

\bibitem[Lu et~al.(2021)Lu, Jin, Pang, Zhang, and Karniadakis]{lu2021learning}
Lu, L., Jin, P., Pang, G., Zhang, Z., and Karniadakis, G.~E.
\newblock {Learning nonlinear operators via DeepONet based on the universal
  approximation theorem of operators}.
\newblock \emph{Nature Machine Intelligence}, 3\penalty0 (3):\penalty0
  218--229, 2021.

\bibitem[Maarten et~al.(2022)Maarten, Lassas, and Wong]{maarten2022deep}
Maarten, V., Lassas, M., and Wong, C.~A.
\newblock Deep learning architectures for nonlinear operator functions and
  nonlinear inverse problems.
\newblock \emph{Mathematical Statistics and Learning}, 4\penalty0 (1):\penalty0
  1--86, 2022.

\bibitem[Mao et~al.(2020)Mao, Lu, Marxen, Zaki, and Karniadakis]{donet1}
Mao, Z., Lu, L., Marxen, O., Zaki, T., and Karniadakis, G.~E.
\newblock Deep{M}and{M}net for hypersonics: {P}redicting the coupled flow and
  finite-rate chemistry behind a normal shock using neural-network
  approximation of operators.
\newblock Preprint, available from arXiv:2011.03349v1, 2020.

\bibitem[Muller \& Siltanen(2012)Muller and Siltanen]{MSbook}
Muller, J. and Siltanen, S.
\newblock \emph{Linear and nonlinear inverse problems with practical
  applications}.
\newblock SIAM, 2012.

\bibitem[Nachman(1988)]{Nach}
Nachman, A.
\newblock Reconstructions from boundary measurements.
\newblock \emph{Ann. Math.}, 128\penalty0 (3):\penalty0 531--576, 1988.

\bibitem[Pathak et~al.(2022)Pathak, Subramanian, Harrington, Raja,
  Chattopadhyay, Mardani, Kurth, Hall, Li, Azizzadenesheli, Hassanzadeh,
  Kashinath, and Anandkumar]{FNO2}
Pathak, J., Subramanian, S., Harrington, P., Raja, S., Chattopadhyay, A.,
  Mardani, M., Kurth, T., Hall, D., Li, Z., Azizzadenesheli, K., Hassanzadeh,
  p., Kashinath, K., and Anandkumar, A.
\newblock {Fourcastnet: A global data-driven high-resolution weather model
  using adaptive Fourier neural operators}.
\newblock \emph{arXiv preprint arXiv:2202.11214}, 2022.

\bibitem[Prasthofer et~al.(2022)Prasthofer, De~Ryck, and Mishra]{vidon}
Prasthofer, M., De~Ryck, T., and Mishra, S.
\newblock Variable input deep operator networks.
\newblock \emph{arXiv preprint arXiv:2205.11404}, 2022.

\bibitem[Stefanov et~al.(2016)Stefanov, Uhlmann, and Vasy]{stefanov2016stable}
Stefanov, P., Uhlmann, G., and Vasy, A.
\newblock {On the stable recovery of a metric from the hyperbolic DN map with
  incomplete data}.
\newblock \emph{Inverse Problems \& Imaging}, 10\penalty0 (4):\penalty0 1141,
  2016.

\bibitem[Stuart(2010)]{stuart2010inverse}
Stuart, A.~M.
\newblock {Inverse problems: a Bayesian perspective}.
\newblock \emph{Acta numerica}, 19:\penalty0 451--559, 2010.

\bibitem[Symes(2009)]{symes2009seismic}
Symes, W.~W.
\newblock The seismic reflection inverse problem.
\newblock \emph{Inverse problems}, 25\penalty0 (12):\penalty0 123008, 2009.

\bibitem[Tarantola(2005)]{tarantola2005inverse}
Tarantola, A.
\newblock \emph{Inverse problem theory and methods for model parameter
  estimation}.
\newblock SIAM, 2005.

\bibitem[Uhlmann(2009)]{eit}
Uhlmann, G.
\newblock {Electrical impedance tomography and Calder{\'o}n's problem}.
\newblock \emph{Inverse problems}, 25\penalty0 (12):\penalty0 123011, 2009.

\bibitem[Yilmaz(2011)]{SIbook}
Yilmaz, O.
\newblock \emph{Seismic Data Analysis}.
\newblock Society for exploration geophysicists, 2011.

\end{thebibliography}
\bibliographystyle{icml2023}

\newpage
\appendix
\onecolumn
\icmltitlerunning{Supplementary Material for "Neural Inverse Operators for solving PDE Inverse problems"}
\begin{center}
{\bf Supplementary Material for:}\\
Neural Inverse Operators for solving PDE Inverse problems.
\end{center}

\section{Depiction of PDE Inverse Problems.}
In the following figures, we illustrate the different PDE inverse problems considered in the main text.

\begin{figure*}[htbp]
\captionsetup[subfloat]{font=large,labelformat=empty,skip=0pt}
\begin{subfigure}[hb]{0.5\linewidth}
\begin{mdframed}[backgroundcolor=gray!10,linecolor=gray!10]
\centering
\subfloat[]{\tikz[remember picture]{\node(1ALa){\includegraphics[width=0.4\textwidth]{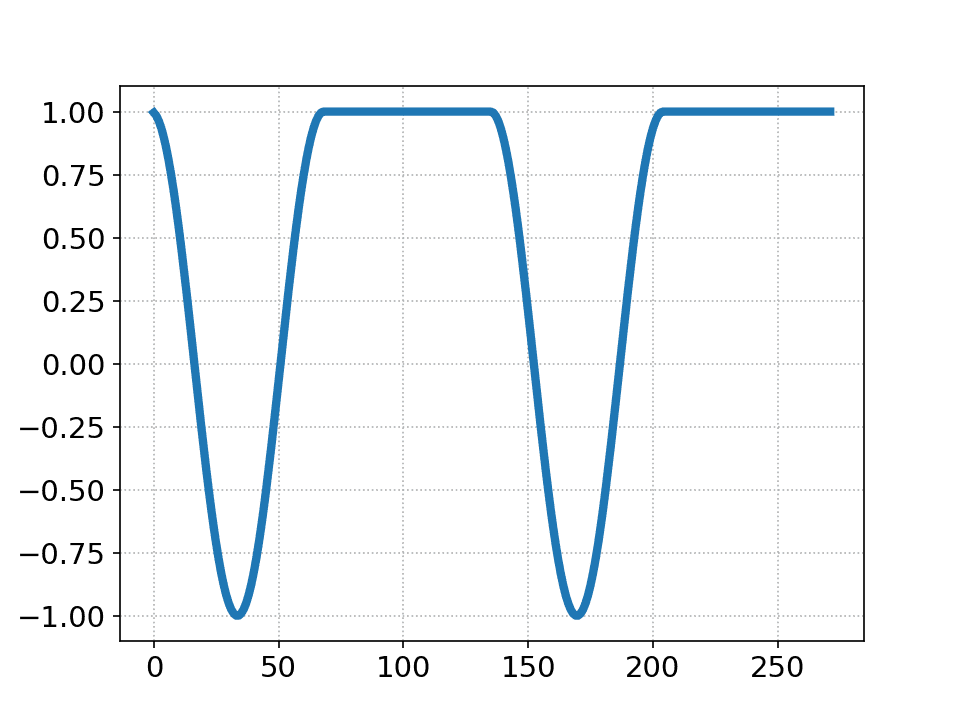}};}
}%
\hspace*{0.5cm}
\subfloat[]{\tikz[remember picture]{\node(1ALb){\includegraphics[width=0.4\textwidth]{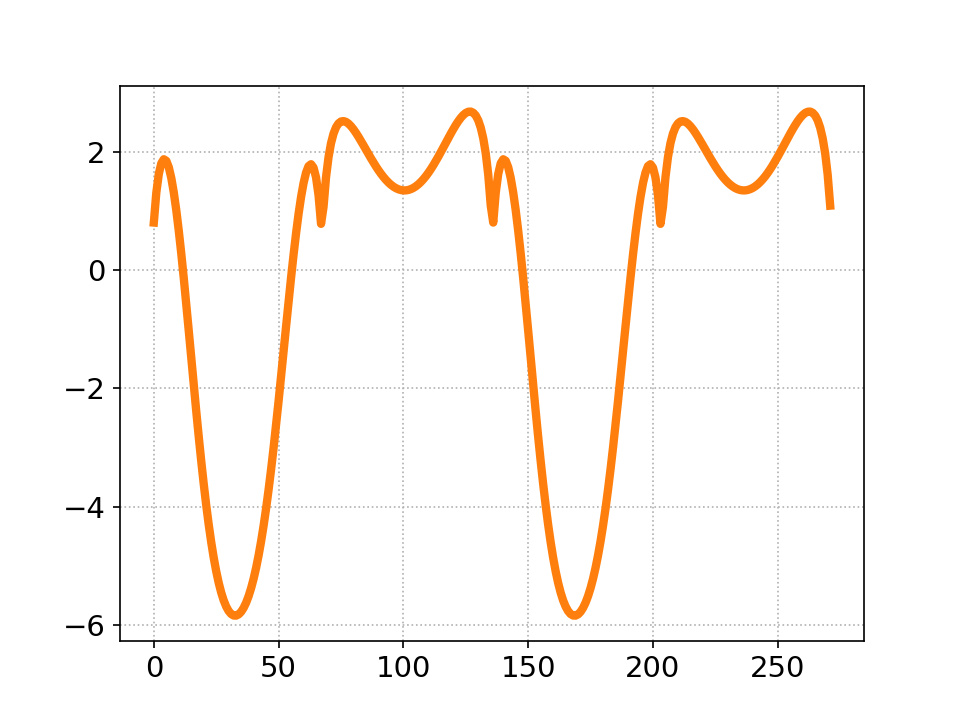}};}
}%
\vspace*{-0.8cm}
\subfloat[]{\tikz[remember picture]{\node(2ALa){\includegraphics[width=0.4\textwidth]{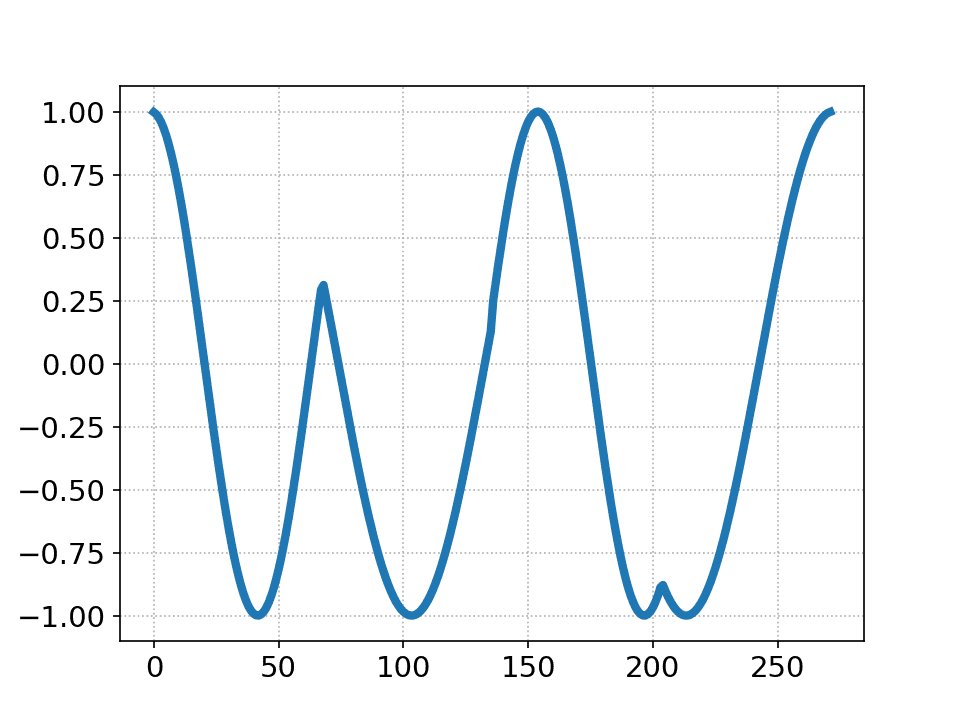}};}
}%
\hspace*{0.5cm}
\subfloat[]{\tikz[remember picture]{\node(2ALb){\includegraphics[width=0.4\textwidth]{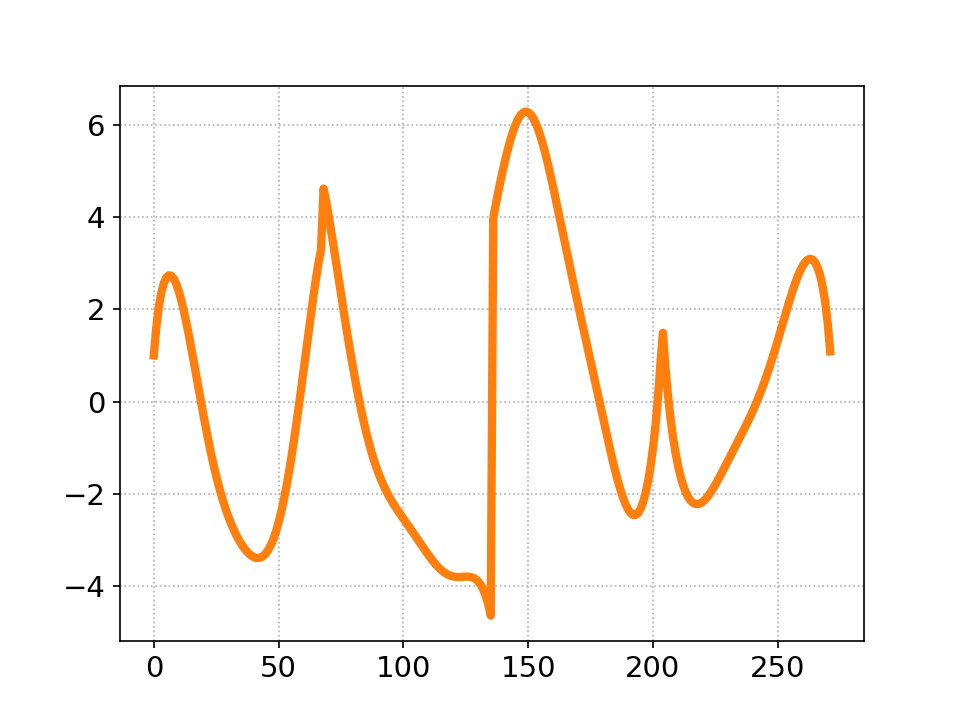}};}
}%
\vspace*{-0.8cm}
\subfloat[{$g$}]{\tikz[remember picture]{\node(3ALa){\includegraphics[width=0.4\textwidth]{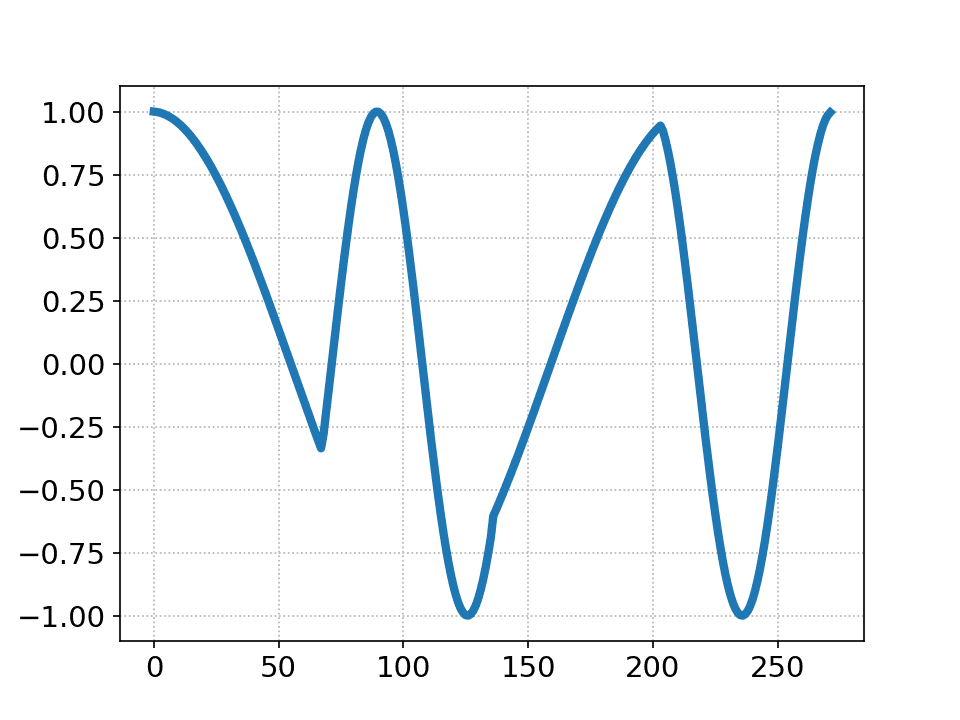}};}
}%
\hspace*{0.5cm}
\subfloat[{$\frac{\partial u}{\partial \nu}\Big|_{\partial D}$}]{\tikz[remember picture]{\node(3ALb){\includegraphics[width=0.4\textwidth]{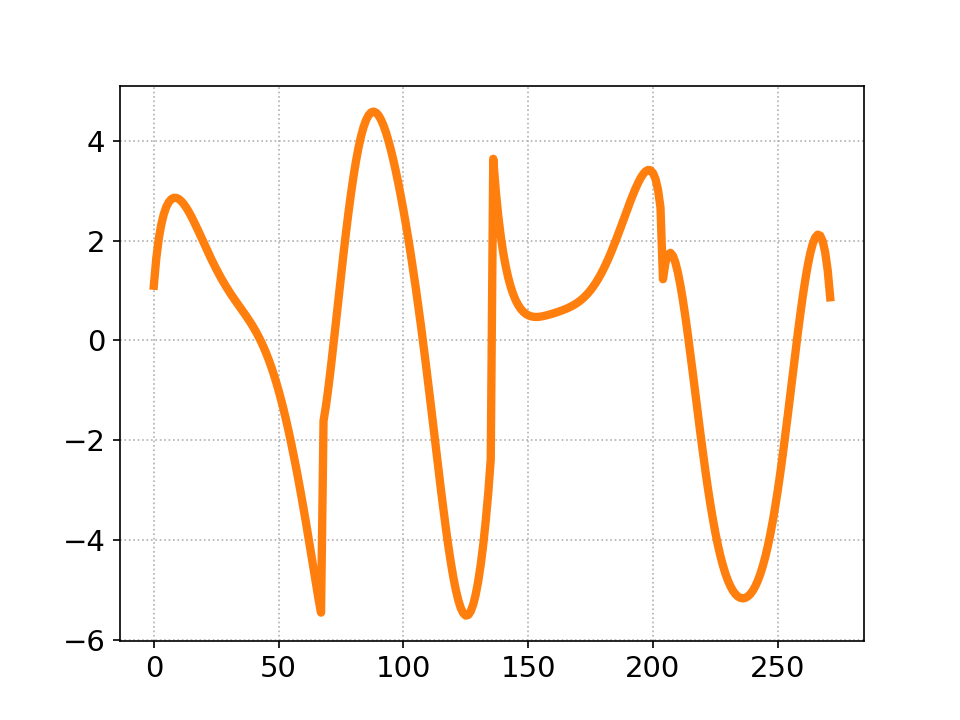}};}
}%
\tikz[overlay,remember picture]{\draw[-{Triangle[width=4pt,length=2pt]}, line width=2pt, ,draw=gray](1ALa) -- (1ALa-|1ALb.west) node[midway,above]{ {$\Lambda_a$}};}
\tikz[overlay,remember picture]{\draw[-{Triangle[width=4pt,length=2pt]}, line width=2pt, ,draw=gray](2ALa) -- (2ALa-|2ALb.west) node[midway,above,]{ {$\Lambda_a$}};}
\tikz[overlay,remember picture]{\draw[-{Triangle[width=4pt,length=2pt]}, line width=2pt, ,draw=gray](3ALa) -- (3ALa-|3ALb.west) node[midway,above,]{ {$\Lambda_a$}};}
\end{mdframed}
\end{subfigure}
%\hspace*{2cm}
\hspace{-0.65cm}
\qquad\tikz[baseline=-\baselineskip]\draw[-{Triangle[width=18pt,length=8pt]}, line width=8pt, ,draw=gray] (0,0) -- ++ (1,0) node[midway,above]{\pmb {$\eF^{-1}$}};
\captionsetup[subfigure]{font=Large,labelformat=empty,skip=10pt}
\hspace{-0.15cm}
\begin{subfigure}[ht]{0.38\linewidth}
\centering
{\includegraphics[width=\textwidth]{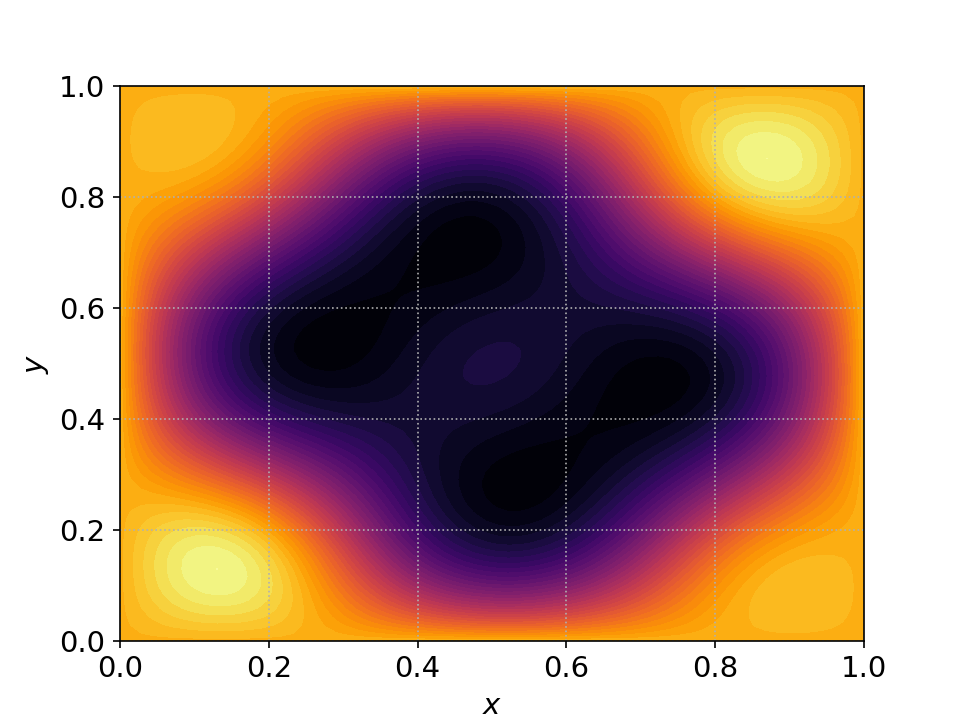}}
\caption{$a$}
\end{subfigure}
\caption{Illustration of a typical input (left) and output (right) sample for the Calder\'{o}n Problem for EIT with trigonometric coefficients. The input is the Dirichlet-to-Neumann (DtN) map \eqref{eq:dtn}, represented here by three Dirichlet Boundary conditions (Voltage) to Current pairs, and the output is the conductivity coefficient $a$.}
\label{fig:sin_sample}
\end{figure*}
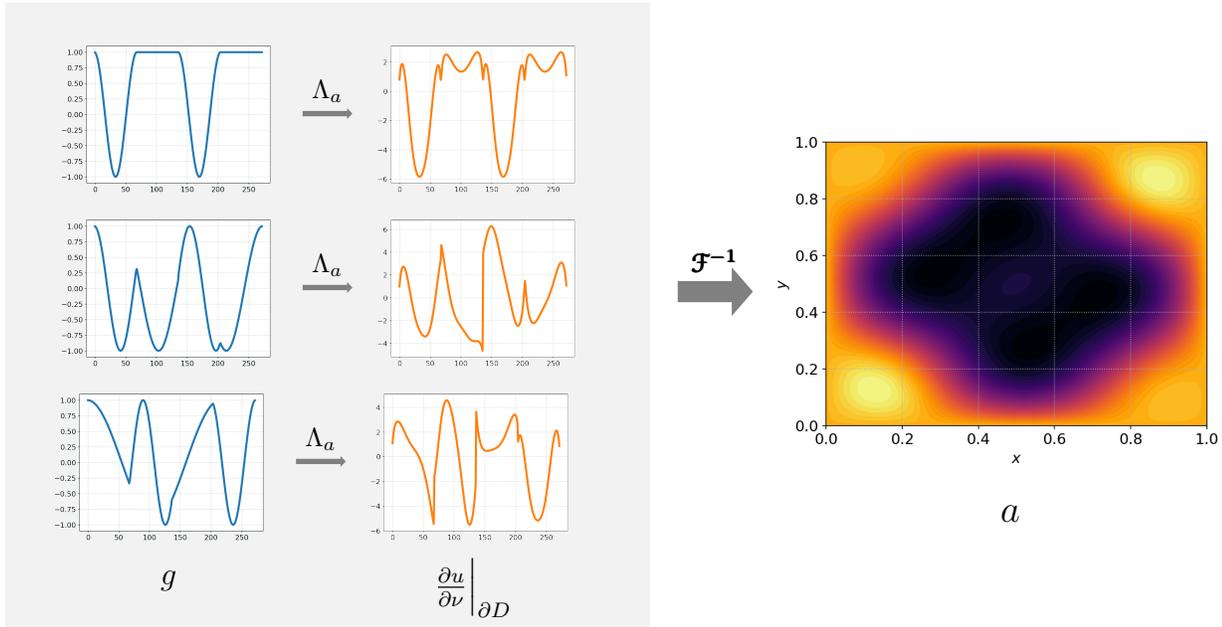

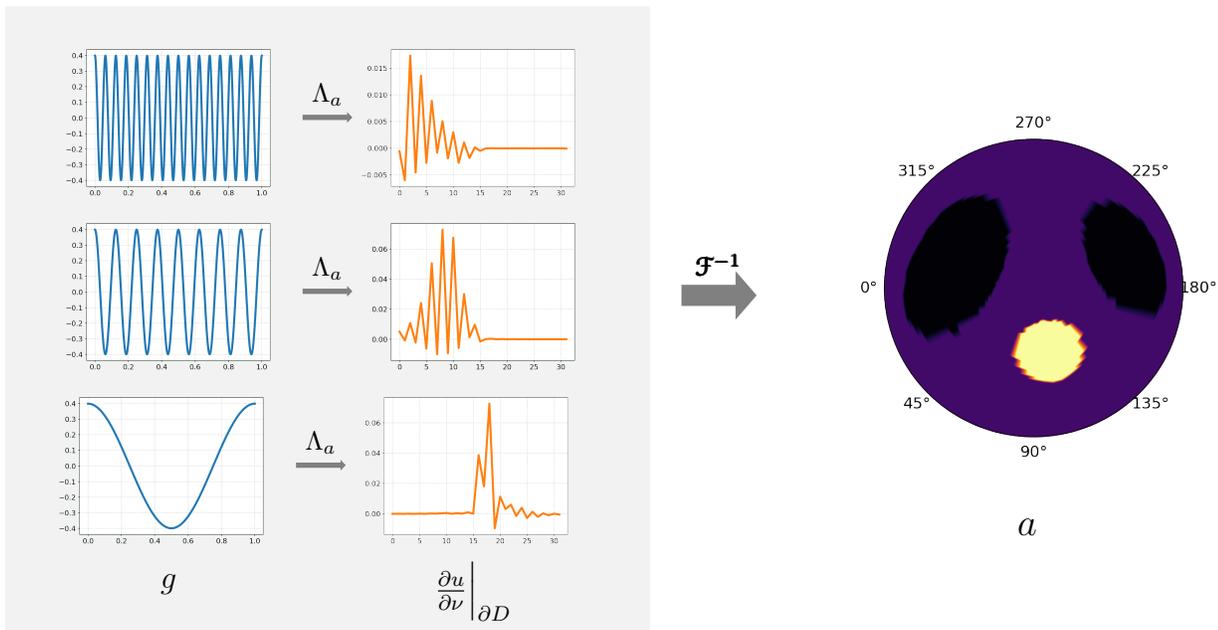
\begin{figure*}[ht!]
\captionsetup[subfloat]{font=large,labelformat=empty,skip=0pt}
\begin{subfigure}[ht]{0.5\linewidth}
\begin{mdframed}[backgroundcolor=gray!10,linecolor=gray!10]
\centering
\subfloat[]{\tikz[remember picture]{\node(1ALa){\includegraphics[width=0.4\textwidth]{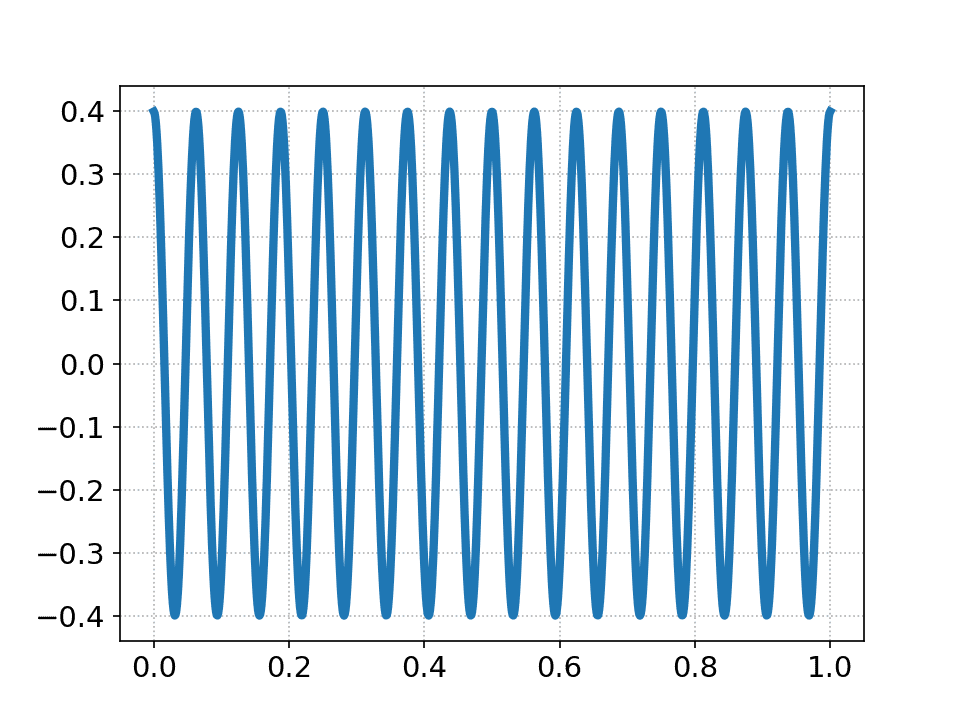}};}
}%
\hspace*{0.5cm}
\subfloat[]{\tikz[remember picture]{\node(1ALb){\includegraphics[width=0.4\textwidth]{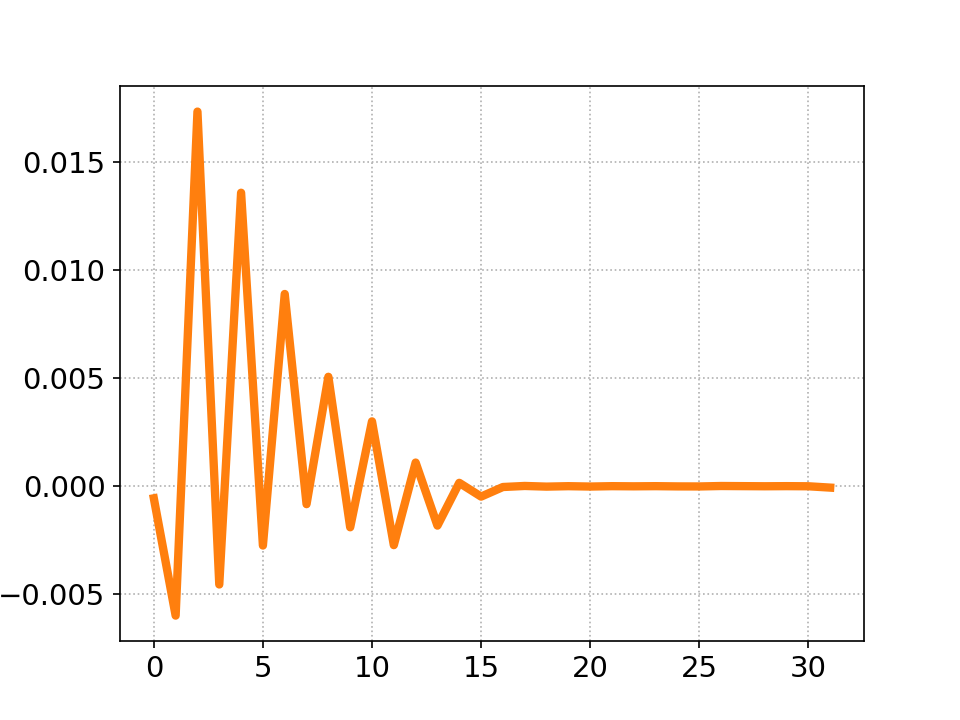}};}
}%
\vspace*{-0.8cm}
\subfloat[]{\tikz[remember picture]{\node(2ALa){\includegraphics[width=0.4\textwidth]{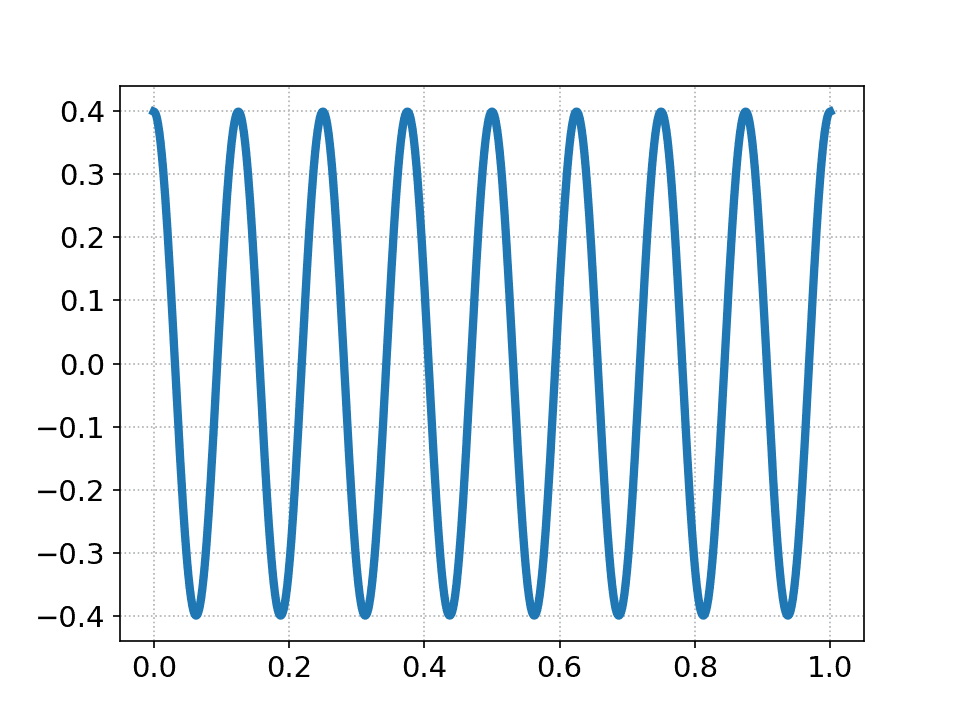}};}
}%
\hspace*{0.5cm}
\subfloat[]{\tikz[remember picture]{\node(2ALb){\includegraphics[width=0.4\textwidth]{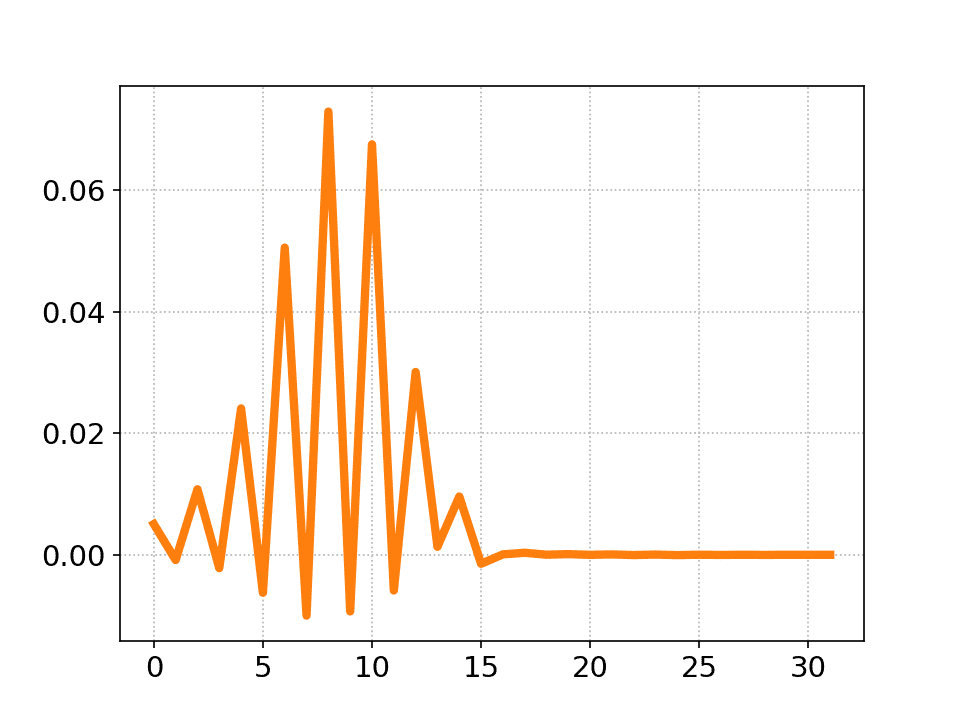}};}
}%
\vspace*{-0.8cm}
\subfloat[{$g$} ]{\tikz[remember picture]{\node(3ALa){\includegraphics[width=0.4\textwidth]{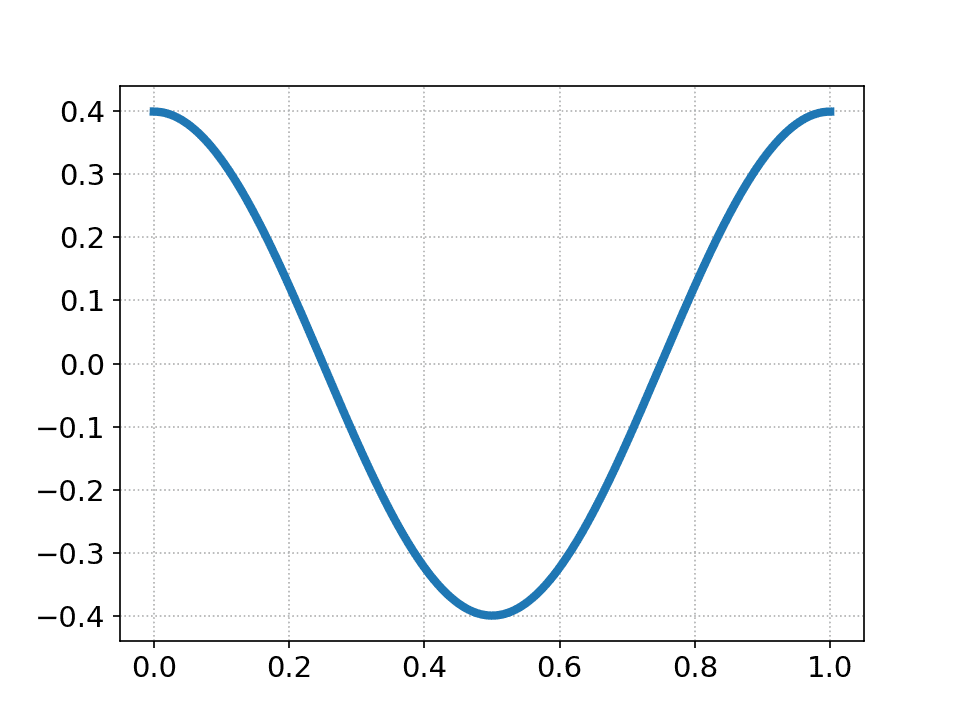}};}
}%
\hspace*{0.5cm}
\subfloat[{$\frac{\partial u}{\partial \nu}\Big|_{\partial D}$} ]{\tikz[remember picture]{\node(3ALb){\includegraphics[width=0.4\textwidth]{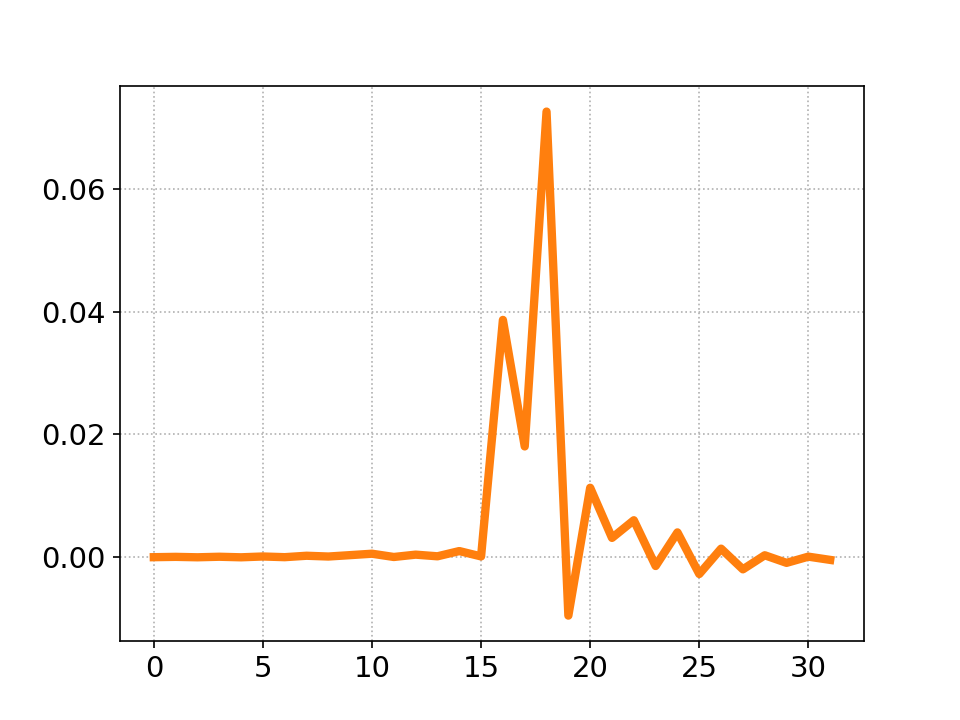}};}
}%
\tikz[overlay,remember picture]{\draw[-{Triangle[width=4pt,length=2pt]}, line width=2pt, ,draw=gray](1ALa) -- (1ALa-|1ALb.west) node[midway,above]{ {$\Lambda_a$}};}
\tikz[overlay,remember picture]{\draw[-{Triangle[width=4pt,length=2pt]}, line width=2pt, ,draw=gray](2ALa) -- (2ALa-|2ALb.west) node[midway,above,]{ {$\Lambda_a$}};}
\tikz[overlay,remember picture]{\draw[-{Triangle[width=4pt,length=2pt]}, line width=2pt, ,draw=gray](3ALa) -- (3ALa-|3ALb.west) node[midway,above,]{ {$\Lambda_a$}};}
\end{mdframed}
\end{subfigure}
%\hspace*{2cm}
\hspace{-0.6cm}
\qquad\tikz[baseline=-\baselineskip]\draw[-{Triangle[width=18pt,length=8pt]}, line width=8pt, ,draw=gray] (0,0) -- ++ (1,0) node[midway,above]{\pmb {$\eF^{-1}$}};
\captionsetup[subfigure]{font=Large,labelformat=empty,skip=10pt}
\hspace{-0.15cm}
\begin{subfigure}[ht]{0.4\linewidth}
\centering
{\includegraphics[width=\textwidth]{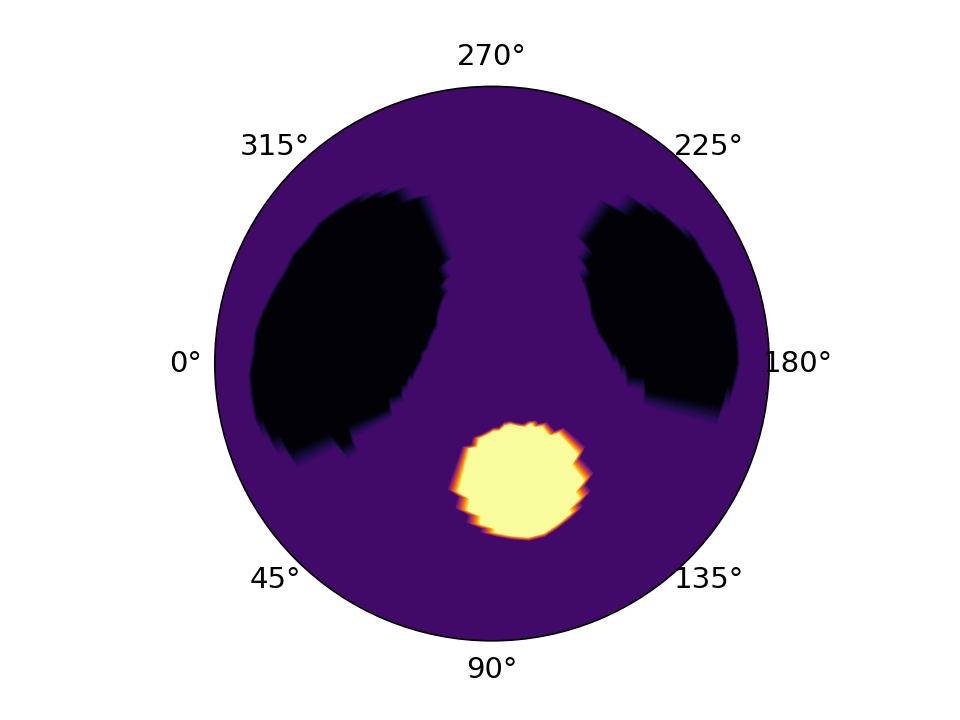}}
\caption{$a$}
\end{subfigure}
\caption{Illustration of EIT for the discontinuous heart-lung Phantom of \cite{MSbook}. Left: Input through the DtN (voltage-to-current) map. Right: Conductivity field showing the Phantom of the heart and lungs.}
\label{fig:heart_sample}
\end{figure*}

\noindent
\begin{figure*}[ht!]
\captionsetup[subfloat]{font=large,labelformat=empty,skip=0pt}
\begin{subfigure}[ht]{0.5\linewidth}
\begin{mdframed}[backgroundcolor=gray!10,linecolor=gray!10]
\centering
\subfloat[]{\tikz[remember picture]{\node(1ALa){\includegraphics[width=0.4\textwidth]{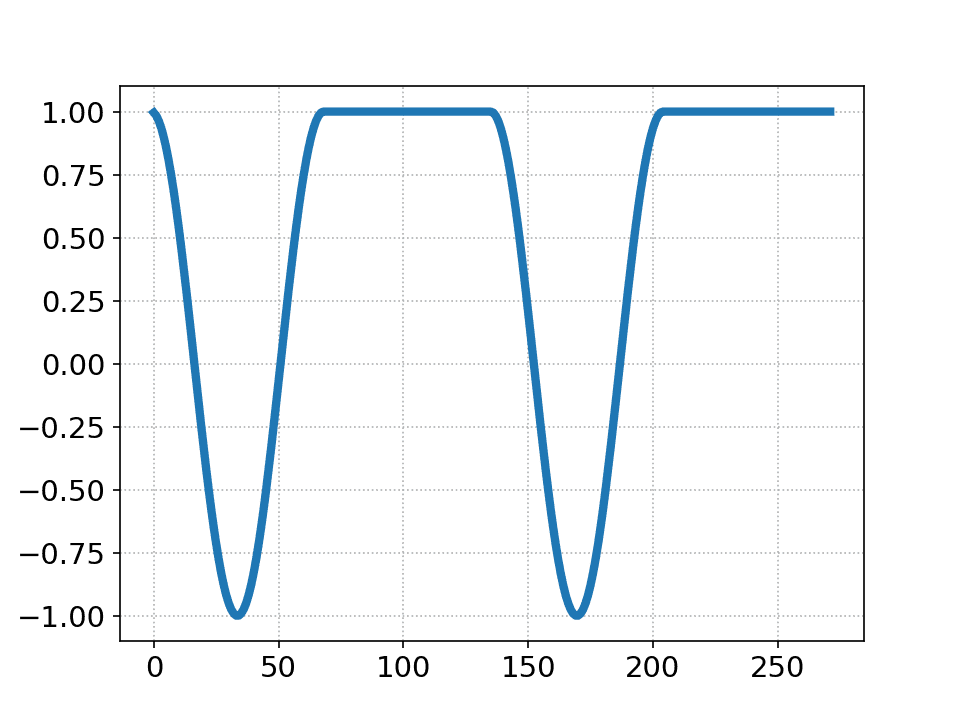}};}
}%
\hspace*{0.5cm}
\subfloat[]{\tikz[remember picture]{\node(1ALb){\includegraphics[width=0.4\textwidth]{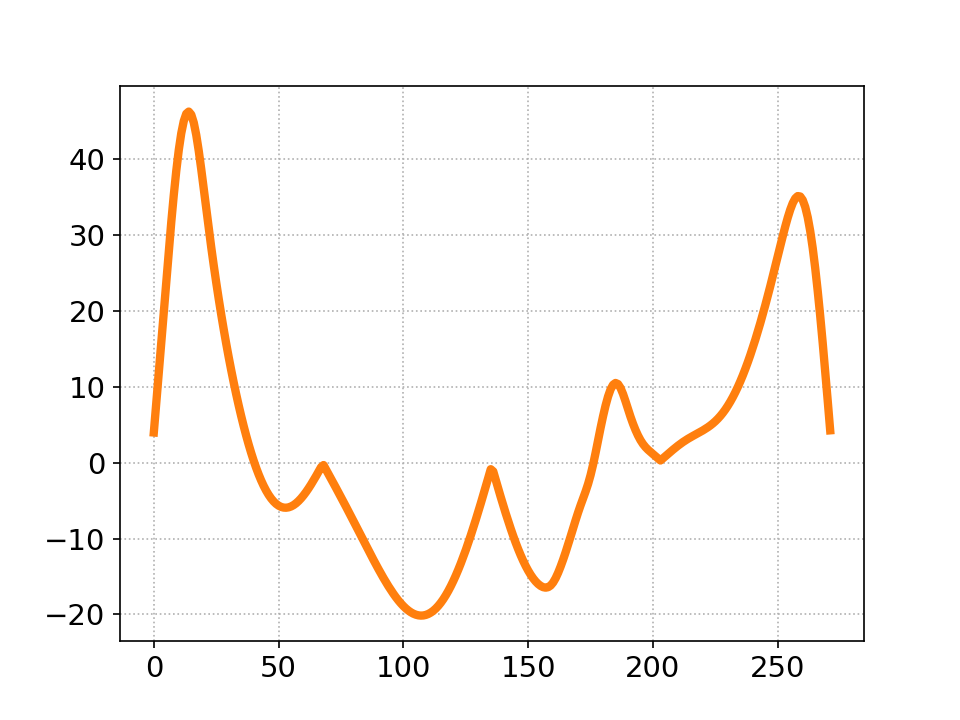}};}
}%
\vspace*{-0.8cm}
\subfloat[]{\tikz[remember picture]{\node(2ALa){\includegraphics[width=0.4\textwidth]{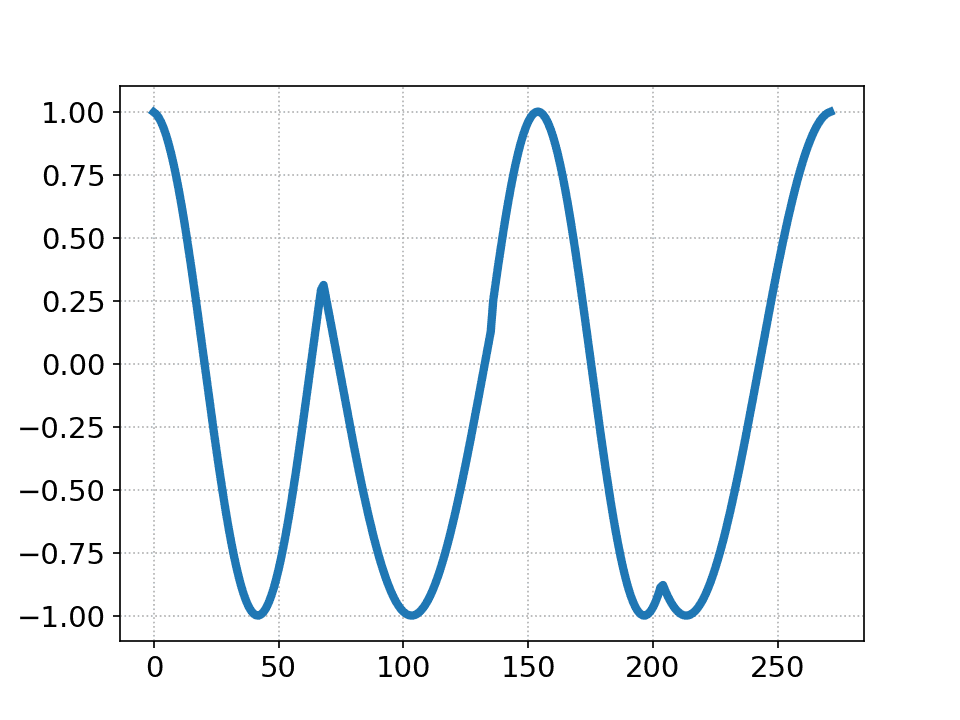}};}
}%
\hspace*{0.5cm}
\subfloat[]{\tikz[remember picture]{\node(2ALb){\includegraphics[width=0.4\textwidth]{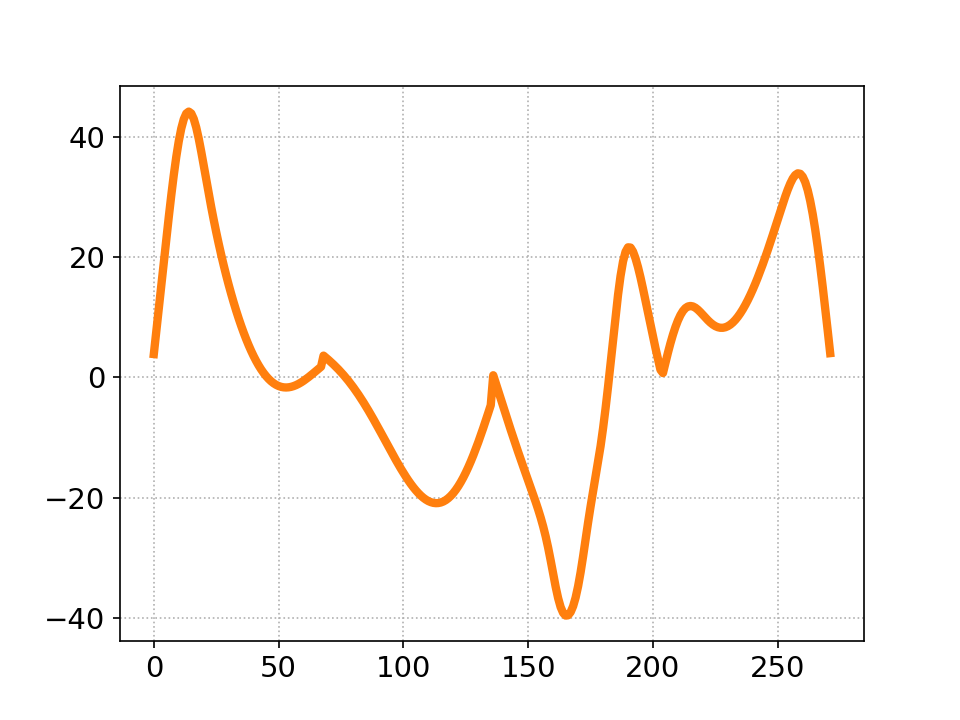}};}
}%
\vspace*{-0.8cm}
\subfloat[{$g$}]{\tikz[remember picture]{\node(3ALa){\includegraphics[width=0.4\textwidth]{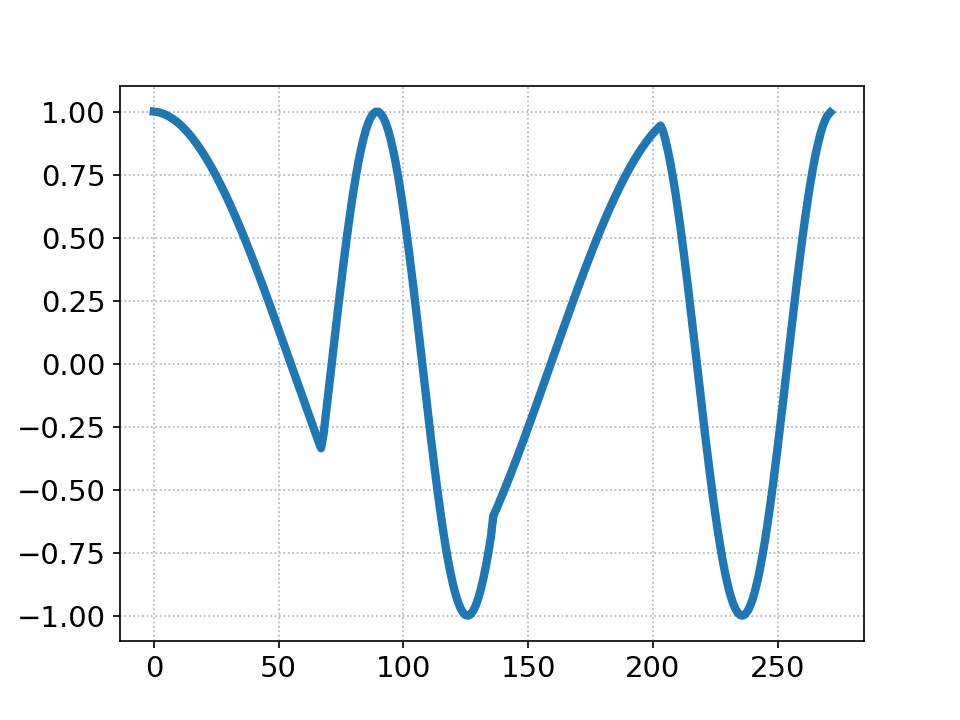}};}
}%
\hspace*{0.5cm}
\subfloat[{$\frac{\partial u}{\partial \nu}\Big|_{\partial D}$}]{\tikz[remember picture]{\node(3ALb){\includegraphics[width=0.4\textwidth]{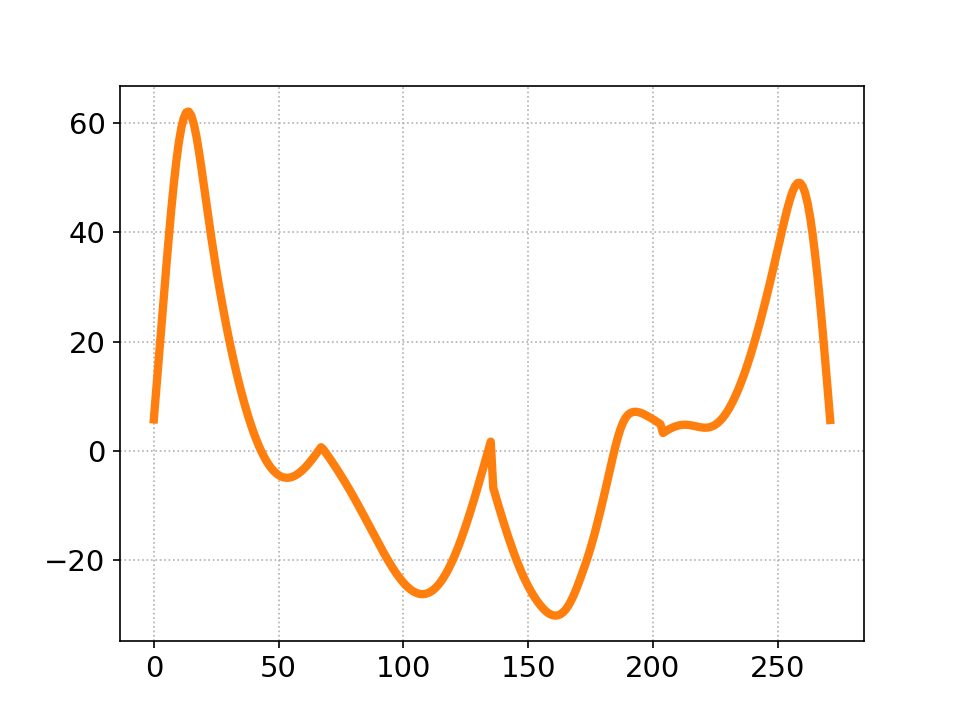}};}
}%
\tikz[overlay,remember picture]{\draw[-{Triangle[width=4pt,length=2pt]}, line width=2pt, ,draw=gray](1ALa) -- (1ALa-|1ALb.west) node[midway,above]{ {$\Lambda_a$}};}
\tikz[overlay,remember picture]{\draw[-{Triangle[width=4pt,length=2pt]}, line width=2pt, ,draw=gray](2ALa) -- (2ALa-|2ALb.west) node[midway,above,]{ {$\Lambda_a$}};}
\tikz[overlay,remember picture]{\draw[-{Triangle[width=4pt,length=2pt]}, line width=2pt, ,draw=gray](3ALa) -- (3ALa-|3ALb.west) node[midway,above,]{ {$\Lambda_a$}};}
\end{mdframed}
\end{subfigure}
%\hspace*{2cm}
\hspace{-0.6cm}
\qquad\tikz[baseline=-\baselineskip]\draw[-{Triangle[width=18pt,length=8pt]}, line width=8pt, ,draw=gray] (0,0) -- ++ (1,0) node[midway,above]{\pmb {$\eF^{-1}$}};
\captionsetup[subfigure]{font=Large,labelformat=empty,skip=10pt}
\hspace{-0.15cm}
\begin{subfigure}[ht]{0.38\linewidth}
\centering
{\includegraphics[width=\textwidth]{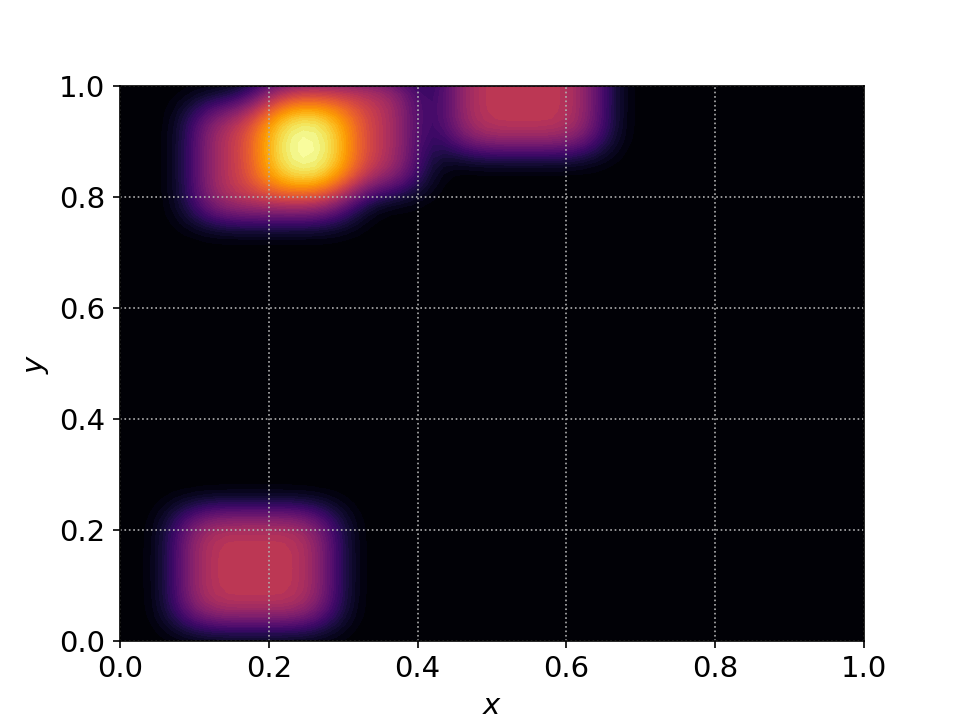}}
\caption{$a$}
\end{subfigure}
\caption{Illustration of detection of inclusions through the Inverse Wave Scattering with the Helmholtz equation. Left: Input represented through 3 samples for the DtN map. Right: Coefficient $a$.}
\label{fig:step_sample}
\end{figure*}
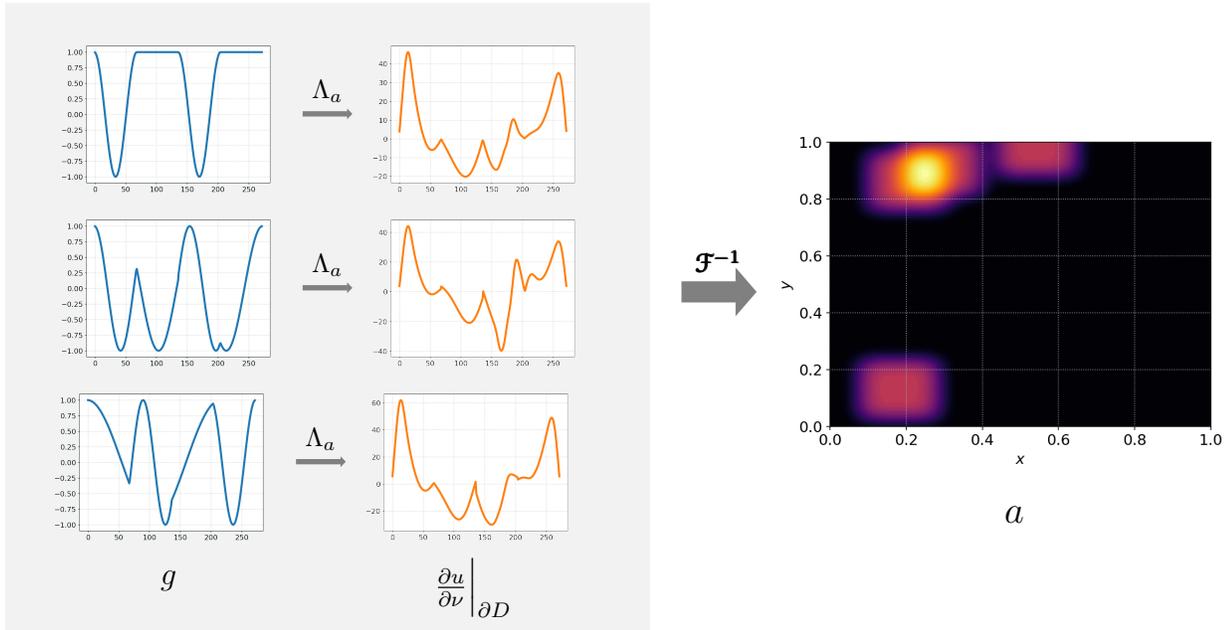

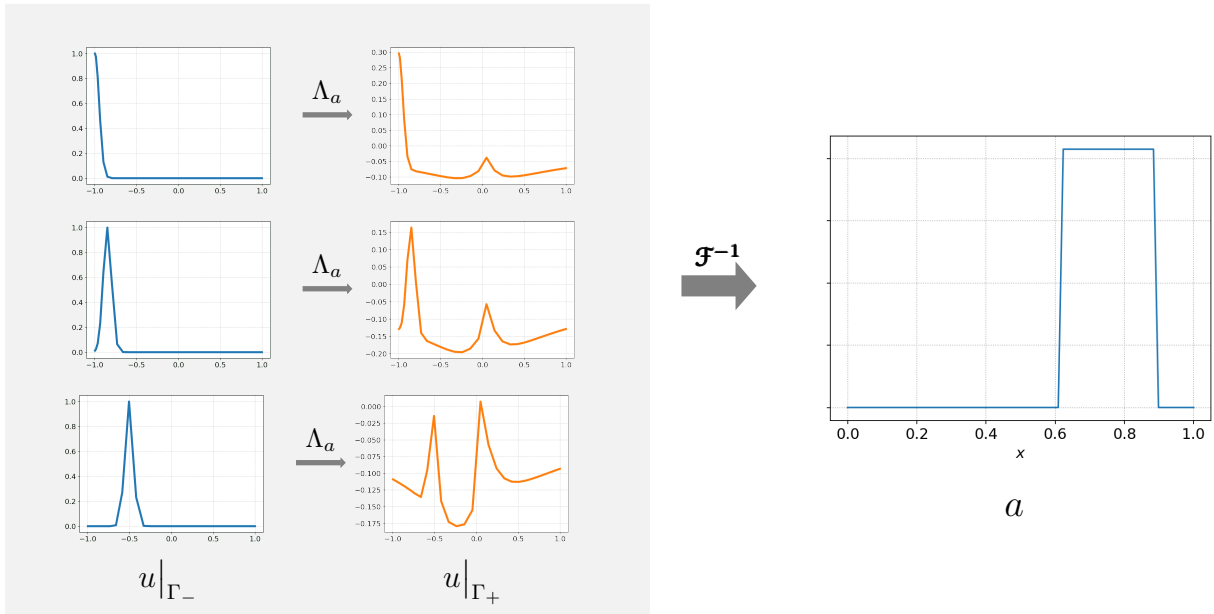
\begin{figure*}[ht!]
\captionsetup[subfloat]{font=large,labelformat=empty,skip=0pt}
\begin{subfigure}[ht]{0.5\linewidth}
\begin{mdframed}[backgroundcolor=gray!10,linecolor=gray!10]
\centering
\subfloat[]{\tikz[remember picture]{\node(1ALa){\includegraphics[width=0.4\textwidth]{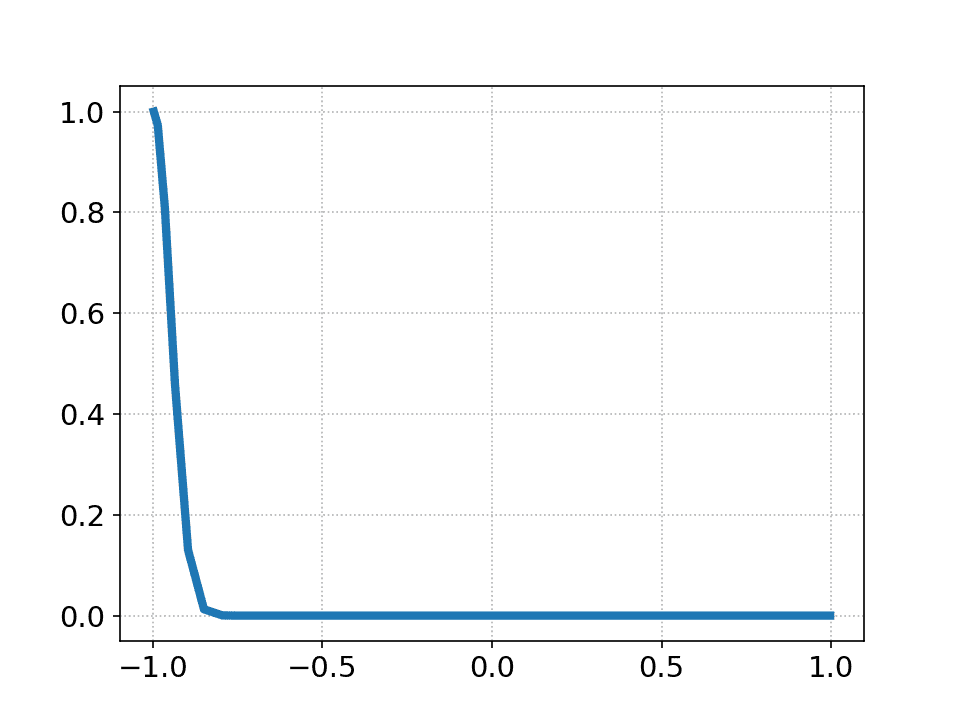}};}
}%
\hspace*{0.5cm}
\subfloat[]{\tikz[remember picture]{\node(1ALb){\includegraphics[width=0.4\textwidth]{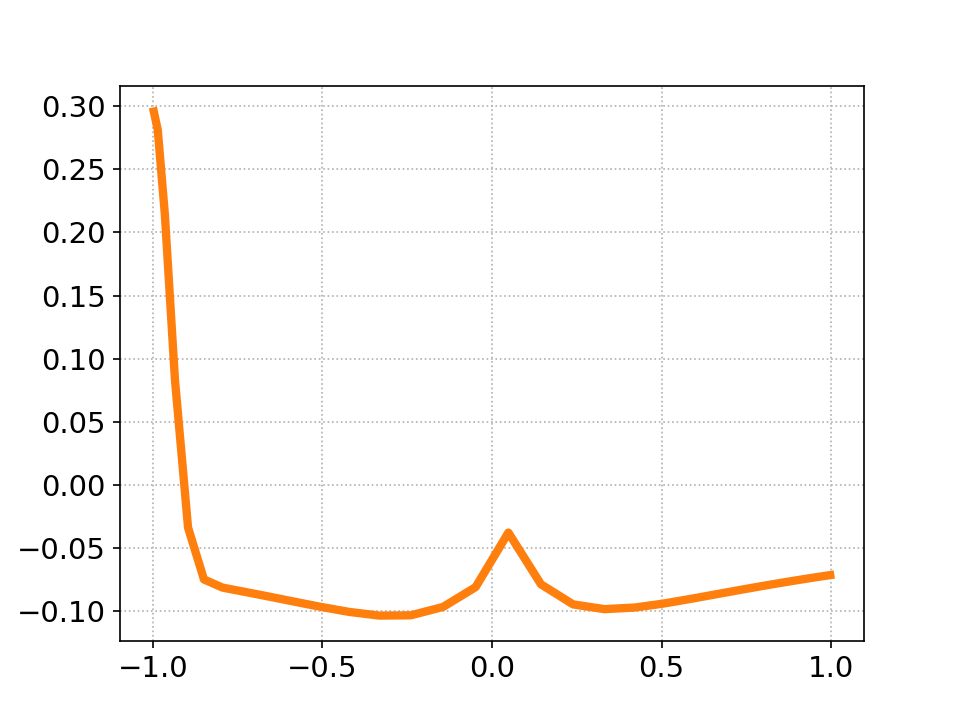}};}
}%
\vspace*{-0.8cm}
\subfloat[]{\tikz[remember picture]{\node(2ALa){\includegraphics[width=0.4\textwidth]{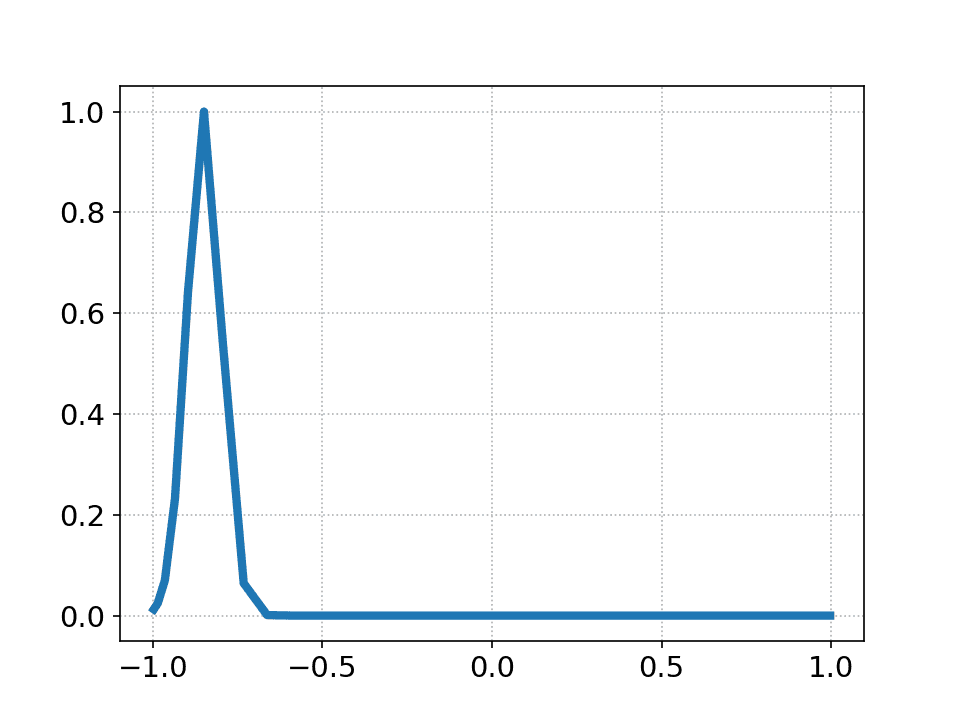}};}
}%
\hspace*{0.5cm}
\subfloat[]{\tikz[remember picture]{\node(2ALb){\includegraphics[width=0.4\textwidth]{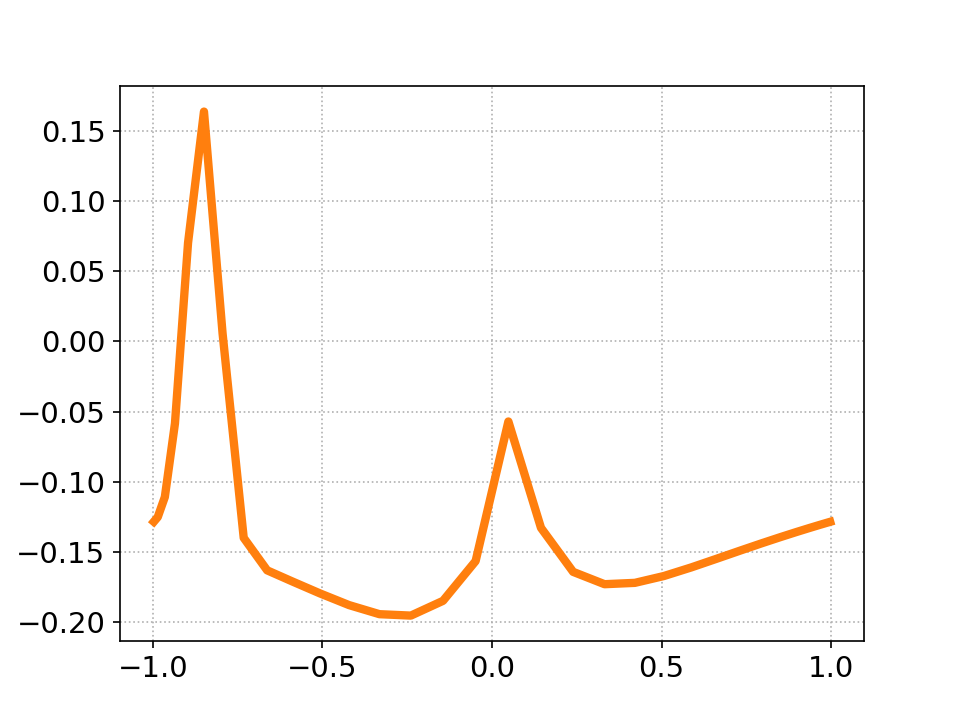}};}
}%
\vspace*{-0.8cm}
\subfloat[{$u\big|_{\Gamma_{-}}$}]{\tikz[remember picture]{\node(3ALa){\includegraphics[width=0.4\textwidth]{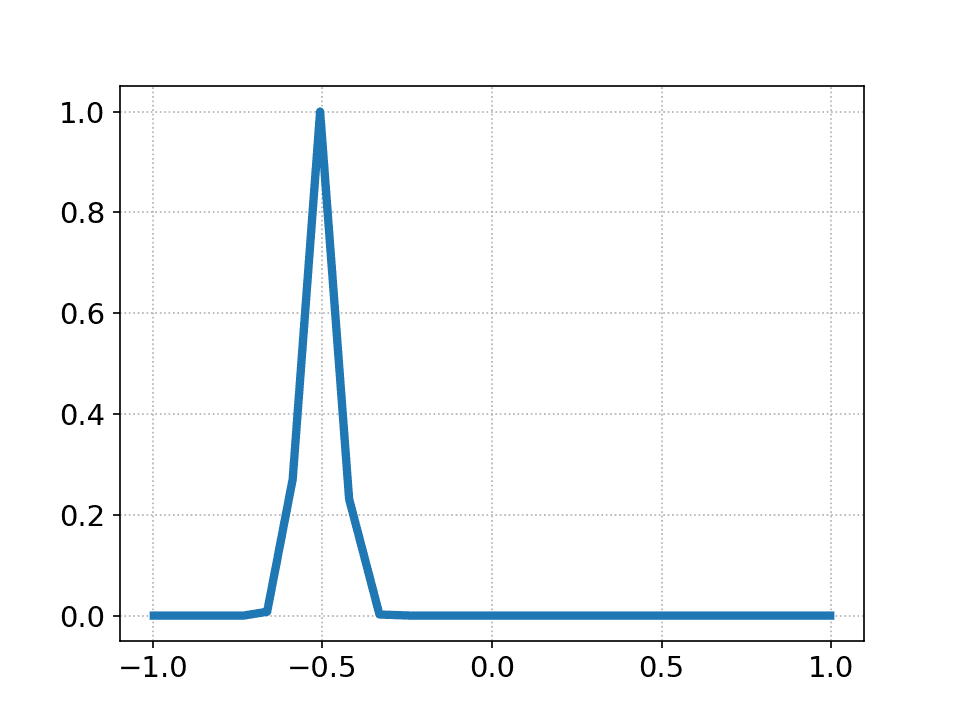}};}
}%
\hspace*{0.5cm}
\subfloat[{$u\big|_{\Gamma_{+}}$}]{\tikz[remember picture]{\node(3ALb){\includegraphics[width=0.4\textwidth]{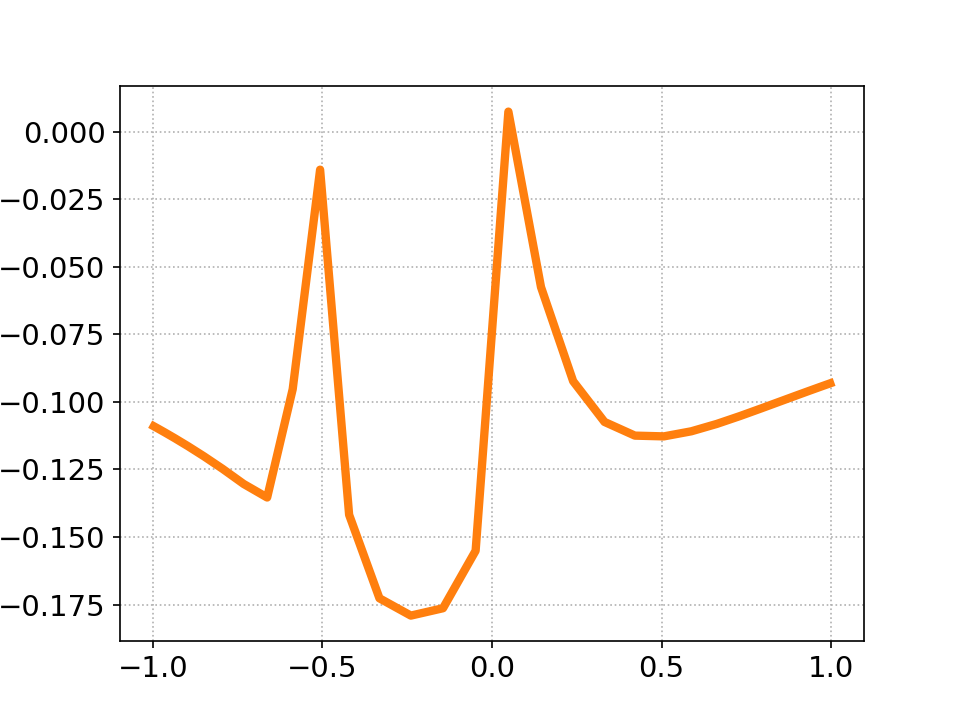}};}
}%
\tikz[overlay,remember picture]{\draw[-{Triangle[width=4pt,length=2pt]}, line width=2pt, ,draw=gray](1ALa) -- (1ALa-|1ALb.west) node[midway,above]{ {$\Lambda_a$}};}
\tikz[overlay,remember picture]{\draw[-{Triangle[width=4pt,length=2pt]}, line width=2pt, ,draw=gray](2ALa) -- (2ALa-|2ALb.west) node[midway,above,]{ {$\Lambda_a$}};}
\tikz[overlay,remember picture]{\draw[-{Triangle[width=4pt,length=2pt]}, line width=2pt, ,draw=gray](3ALa) -- (3ALa-|3ALb.west) node[midway,above,]{ {$\Lambda_a$}};}
\end{mdframed}
\end{subfigure}
%\hspace*{2cm}
\hspace{-0.6cm}
\qquad\tikz[baseline=-\baselineskip]\draw[-{Triangle[width=18pt,length=8pt]}, line width=8pt, ,draw=gray] (0,0) -- ++ (1,0) node[midway,above]{\pmb {$\eF^{-1}$}};
\captionsetup[subfigure]{font=Large,labelformat=empty,skip=10pt}
\hspace{-0.15cm}
\begin{subfigure}[ht]{0.38\linewidth}
\centering
{\includegraphics[width=\textwidth]{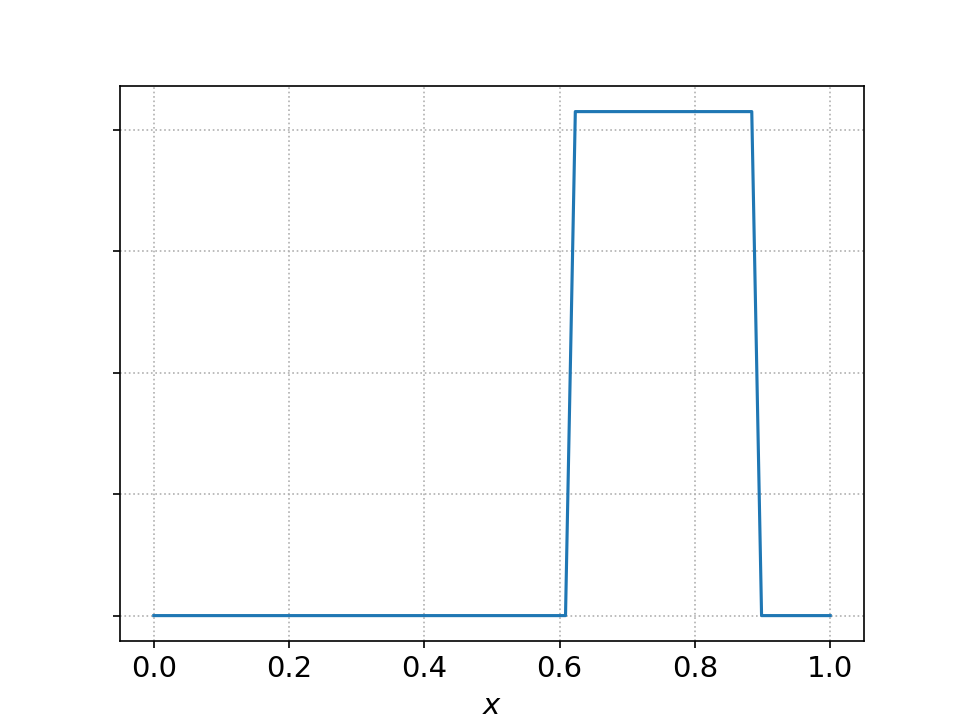}}
\caption{$a$}
\end{subfigure}
\caption{Illustration of Optimal Imaging through the Radiative Transport Equation. Left: Input is the Albedo operator \eqref{eq:albop} illustrated with three mappings between the inflow and outflow boundaries. Right: Output is the Scattering coefficient.}
\label{fig:rad_sample}
\end{figure*}

\begin{figure*}[ht!]
\captionsetup[subfloat]{font=large,labelformat=empty,skip=0pt}
\begin{subfigure}[ht]{0.5\linewidth}
\begin{mdframed}[backgroundcolor=gray!10,linecolor=gray!10]
\centering
\subfloat[]{\tikz[remember picture]{\node(1ALa){\includegraphics[width=0.4\textwidth]{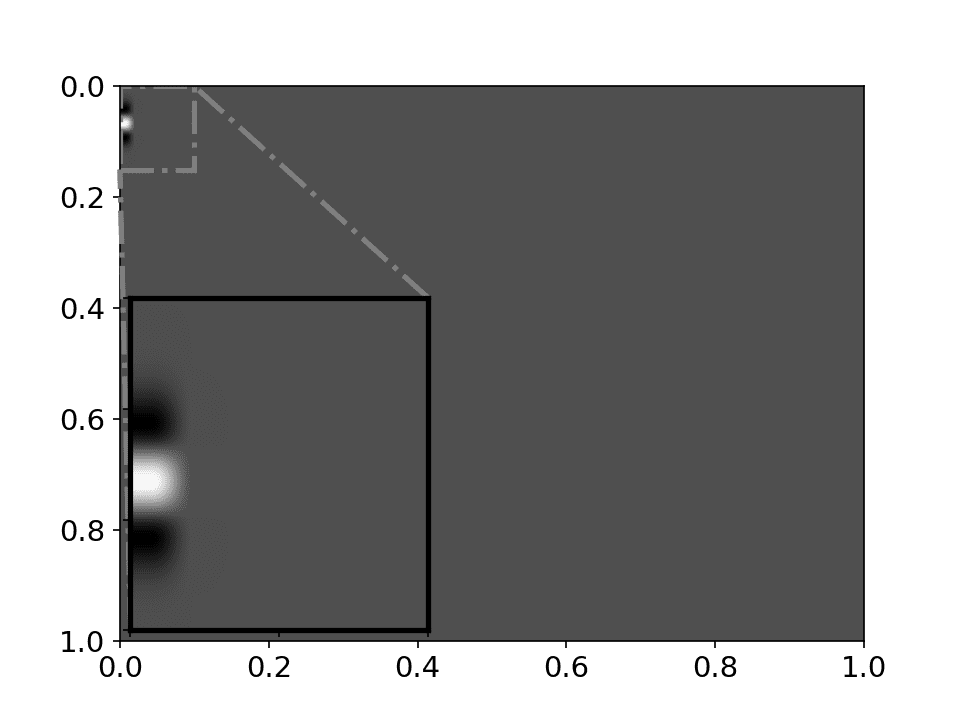}};}
}%
\hspace*{0.5cm}
\subfloat[]{\tikz[remember picture]{\node(1ALb){\includegraphics[width=0.4\textwidth]{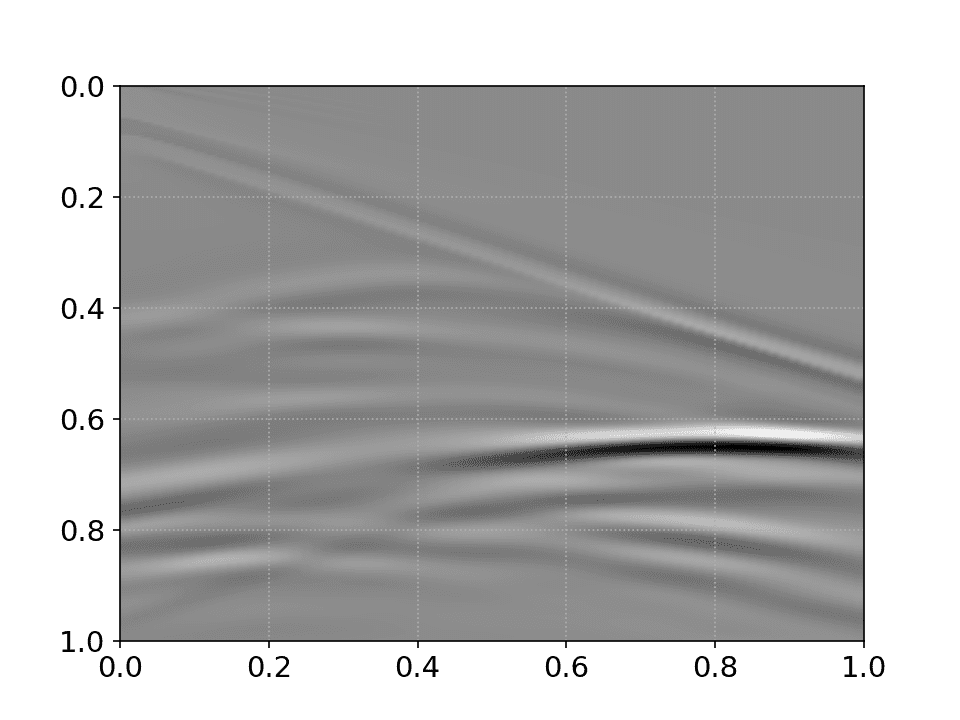}};}
}%
\vspace*{-0.8cm}
\subfloat[]{\tikz[remember picture]{\node(2ALa){\includegraphics[width=0.4\textwidth]{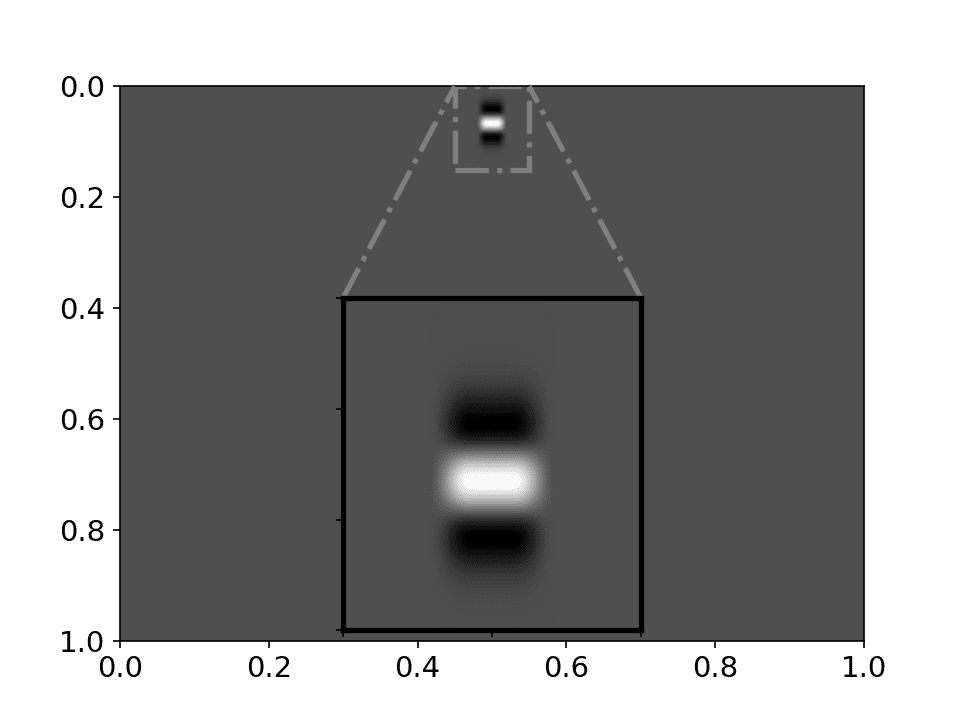}};}
}%
\hspace*{0.5cm}
\subfloat[]{\tikz[remember picture]{\node(2ALb){\includegraphics[width=0.4\textwidth]{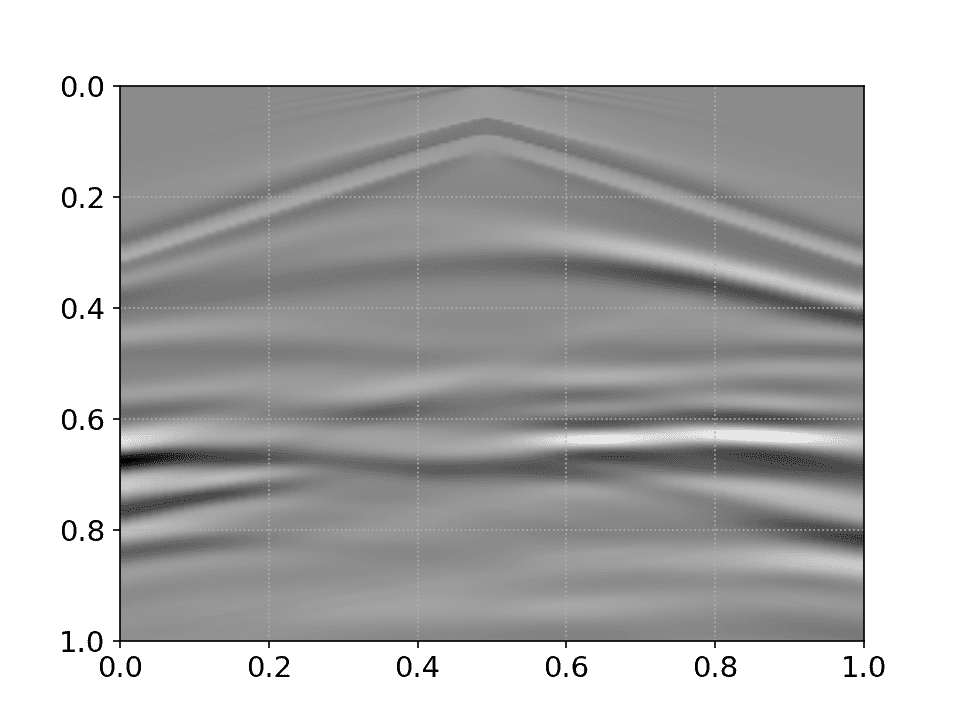}};}
}%
\vspace*{-0.8cm}
\subfloat[{$s$}]{\tikz[remember picture]{\node(3ALa){\includegraphics[width=0.4\textwidth]{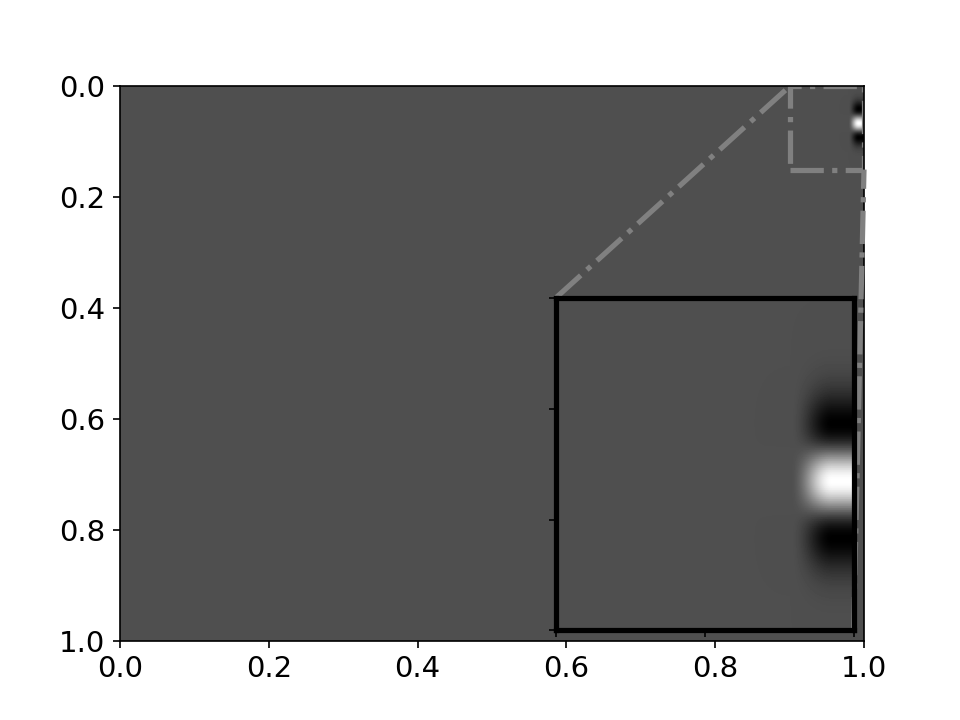}};}
}%
\hspace*{0.5cm}
\subfloat[{$ u\big|_{\eR} $}]{\tikz[remember picture]{\node(3ALb){\includegraphics[width=0.4\textwidth]{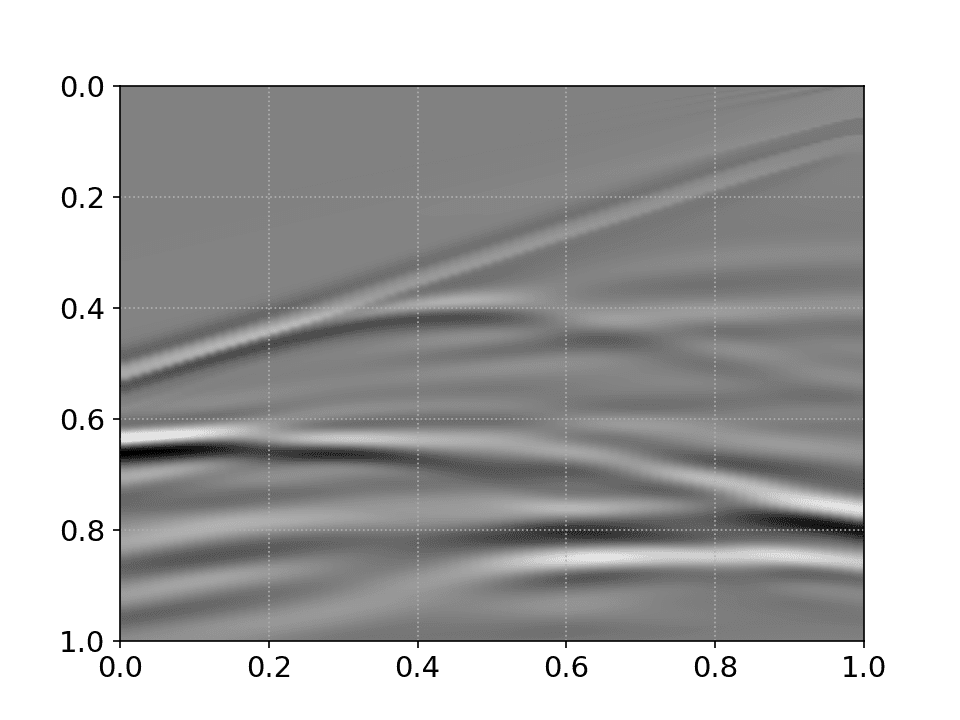}};}
}%
\tikz[overlay,remember picture]{\draw[-{Triangle[width=4pt,length=2pt]}, line width=2pt, ,draw=gray](1ALa) -- (1ALa-|1ALb.west) node[midway,above]{ {$\Lambda_a$}};}
\tikz[overlay,remember picture]{\draw[-{Triangle[width=4pt,length=2pt]}, line width=2pt, ,draw=gray](2ALa) -- (2ALa-|2ALb.west) node[midway,above,]{ {$\Lambda_a$}};}
\tikz[overlay,remember picture]{\draw[-{Triangle[width=4pt,length=2pt]}, line width=2pt, ,draw=gray](3ALa) -- (3ALa-|3ALb.west) node[midway,above,]{ {$\Lambda_a$}};}
\end{mdframed}
\end{subfigure}
%\hspace*{2cm}
\hspace{-0.6cm}
\qquad\tikz[baseline=-\baselineskip]\draw[-{Triangle[width=18pt,length=8pt]}, line width=8pt, ,draw=gray] (0,0) -- ++ (1,0) node[midway,above]{\pmb {$\eF^{-1}$}};
\captionsetup[subfigure]{font=Large,labelformat=empty,skip=10pt}
\hspace{-0.15cm}
\begin{subfigure}[ht]{0.38\linewidth}
\centering
{\includegraphics[width=\textwidth]{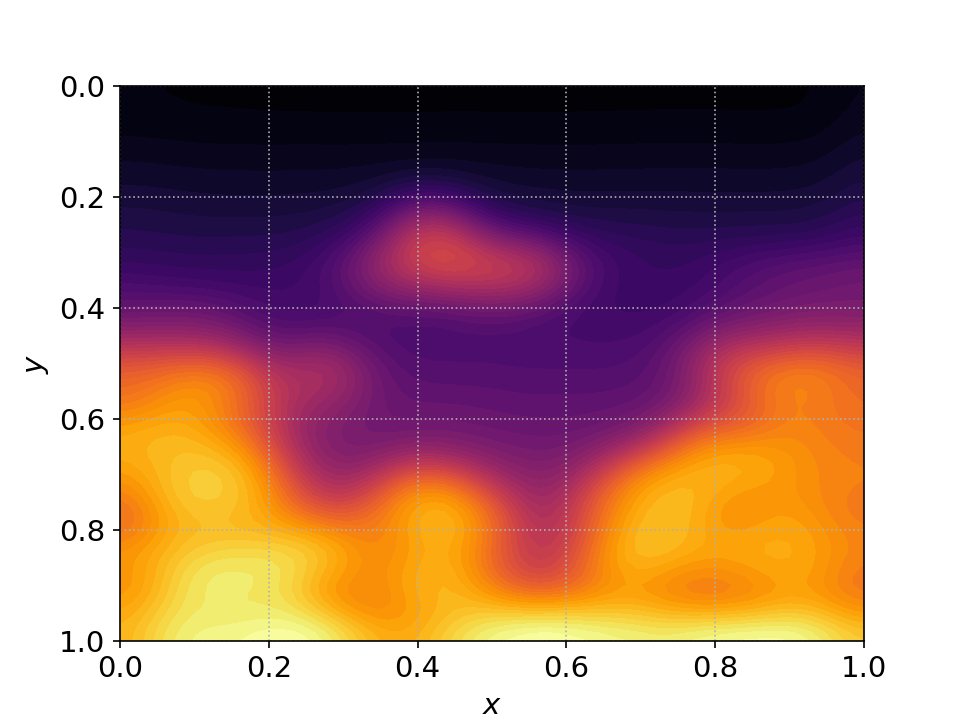}}
\caption{$a$}
\end{subfigure}
\caption{Illustration of Seismic Imaging. Left: Input is Source-to-Receiver map \eqref{eq:str} between Incident waves generated at Sources to Temporal signals recorder at Receivers. Right: Output is the velocity coefficient, corresponding to \emph{Style A} dataset of \cite{deng2021openfwi}.}
\label{fig:style_sample}
\end{figure*}

\noindent
\begin{figure*}[ht!]
\captionsetup[subfloat]{font=large,labelformat=empty,skip=0pt}
\begin{subfigure}[ht]{0.5\linewidth}
\begin{mdframed}[backgroundcolor=gray!10,linecolor=gray!10]
\centering
\subfloat[]{\tikz[remember picture]{\node(1ALa){\includegraphics[width=0.4\textwidth]{ImagesSamples/G_curve0.png}};}
}%
\hspace*{0.5cm}
\subfloat[]{\tikz[remember picture]{\node(1ALb){\includegraphics[width=0.4\textwidth]{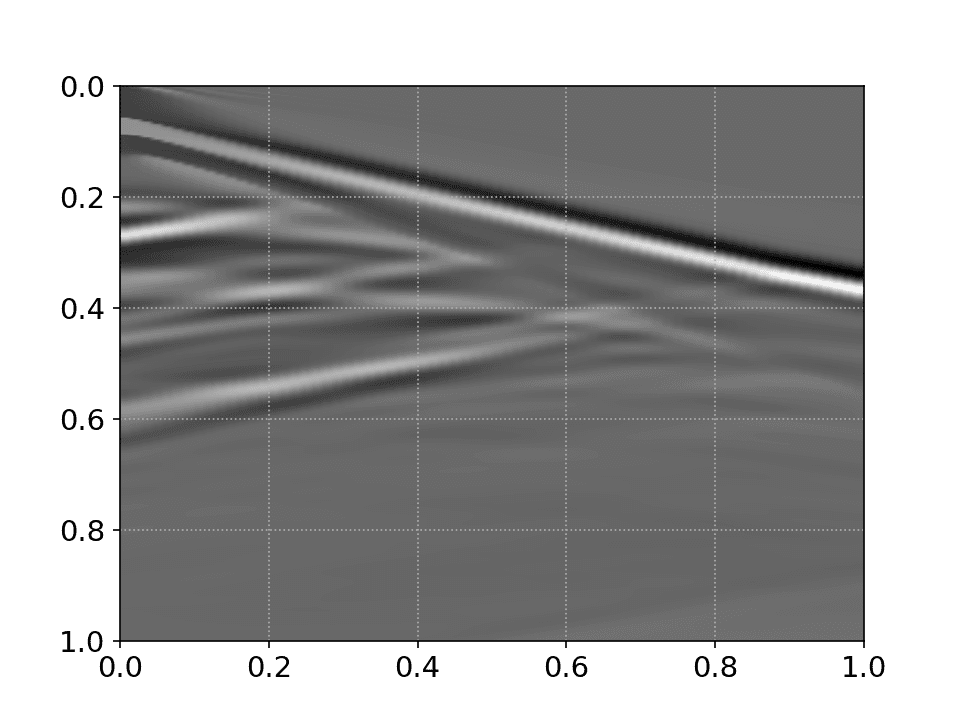}};}
}%
\vspace*{-0.8cm}
\subfloat[]{\tikz[remember picture]{\node(2ALa){\includegraphics[width=0.4\textwidth]{ImagesSamples/G_curve2.png}};}
}%
\hspace*{0.5cm}
\subfloat[]{\tikz[remember picture]{\node(2ALb){\includegraphics[width=0.4\textwidth]{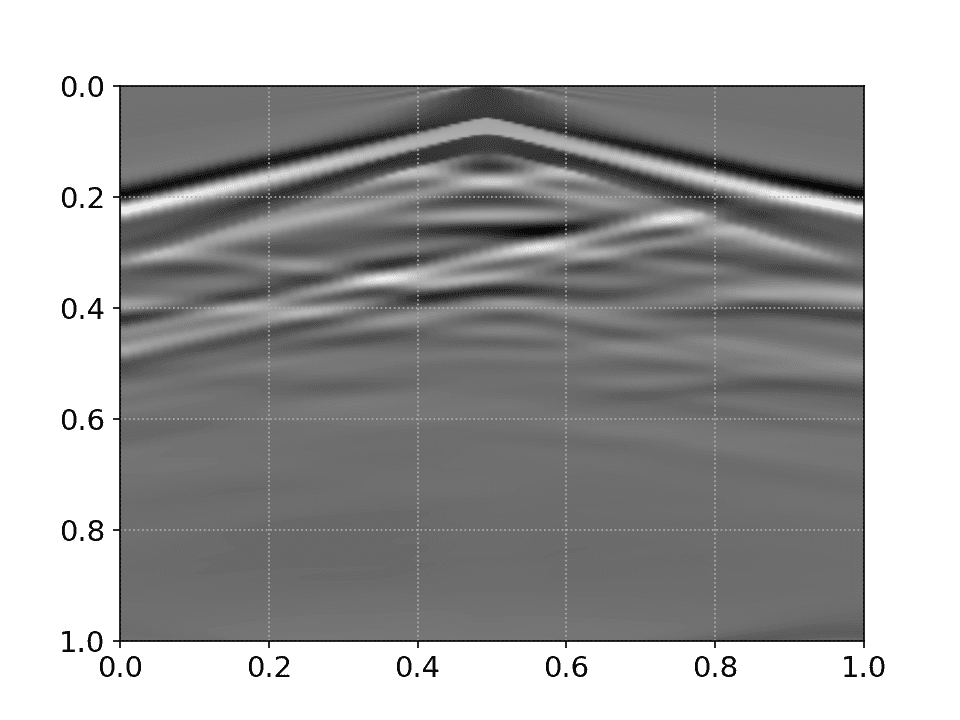}};}
}%
\vspace*{-0.8cm}
\subfloat[{$s$} ]{\tikz[remember picture]{\node(3ALa){\includegraphics[width=0.4\textwidth]{ImagesSamples/G_curve4.png}};}
}%
\hspace*{0.5cm}
\subfloat[{$ u\big|_{\eR} $} ]{\tikz[remember picture]{\node(3ALb){\includegraphics[width=0.4\textwidth]{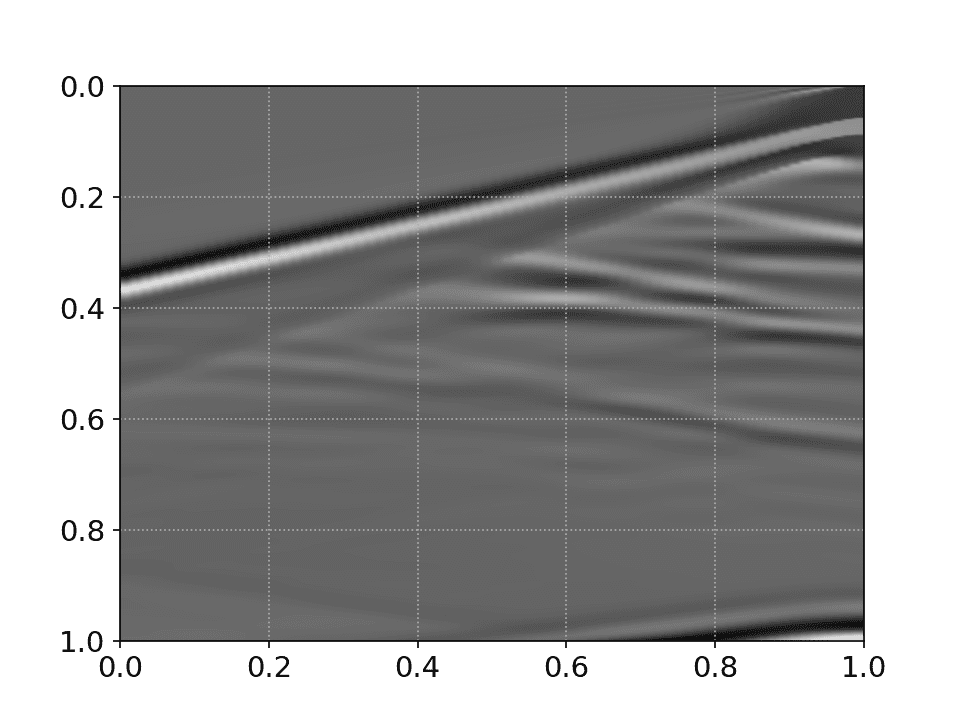}};}
}%
\tikz[overlay,remember picture]{\draw[-{Triangle[width=4pt,length=2pt]}, line width=2pt, ,draw=gray](1ALa) -- (1ALa-|1ALb.west) node[midway,above]{ {$\Lambda_a$}};}
\tikz[overlay,remember picture]{\draw[-{Triangle[width=4pt,length=2pt]}, line width=2pt, ,draw=gray](2ALa) -- (2ALa-|2ALb.west) node[midway,above,]{ {$\Lambda_a$}};}
\tikz[overlay,remember picture]{\draw[-{Triangle[width=4pt,length=2pt]}, line width=2pt, ,draw=gray](3ALa) -- (3ALa-|3ALb.west) node[midway,above,]{ {$\Lambda_a$}};}
\end{mdframed}
\end{subfigure}
%\hspace*{2cm}
\hspace{-0.6cm}
\qquad\tikz[baseline=-\baselineskip]\draw[-{Triangle[width=18pt,length=8pt]}, line width=8pt, ,draw=gray] (0,0) -- ++ (1,0) node[midway,above]{\pmb {$\eF^{-1}$}};
\captionsetup[subfigure]{font=Large,labelformat=empty,skip=10pt}
\hspace{-0.15cm}
\begin{subfigure}[ht]{0.38\linewidth}
\centering
{\includegraphics[width=\textwidth]{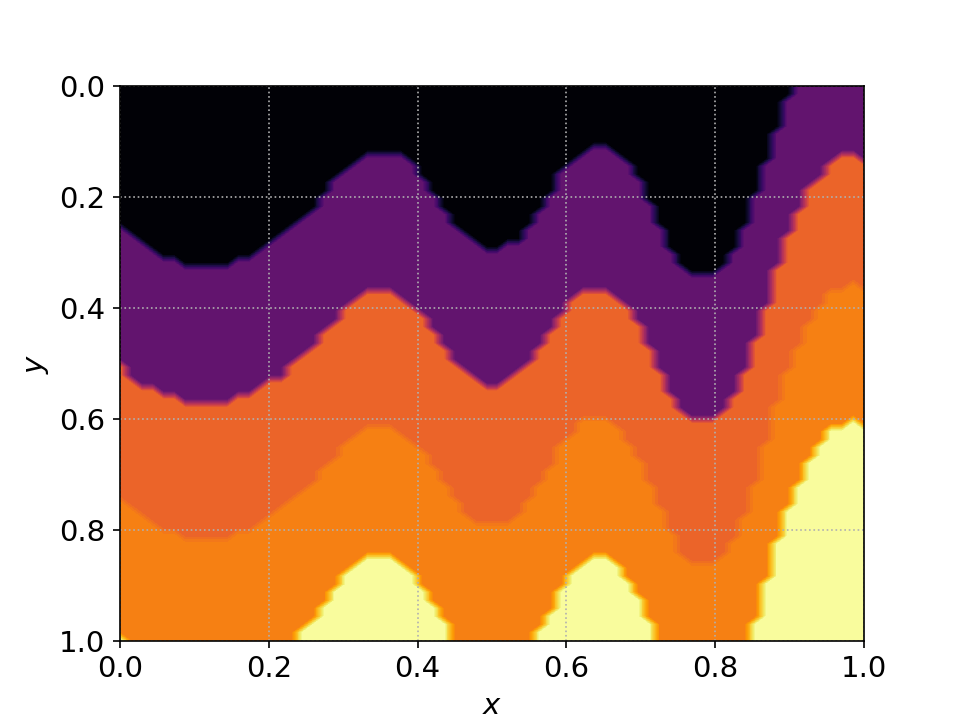}}
\caption{$a$}
\end{subfigure}
\caption{Illustration of Seismic Imaging. Left: Input is Source-to-Receiver map \eqref{eq:str} between Incident waves generated at Sources to Temporal signals recorder at Receivers. Right: Output is the velocity coefficient, corresponding to \emph{CurveVel A} dataset of \cite{deng2021openfwi}}
\label{fig:curve_sample}
\end{figure*}
\newpage
\section{Proof of Formula \eqref{eq:arep} in Main Text.}
\label{app:pf}
Below, we prove the representation formula \eqref{eq:arep} in the main text.
\begin{proof}
Multiplying $u_k$ (the solution of \eqref{eq:helm1}) to Eqn \eqref{eq:nep} and integrating over space, we obtain,
\begin{equation*}
\int\limits_D u_k \Delta \varphi_k dz + \lambda_k\int\limits_D \varphi_k dz = 0
\end{equation*}
Integrating by parts in the above equation and using the Gauss-Green formula yields,
\begin{equation}
\label{eq:pf1}
-\int\limits_D \langle \nabla u_k, \nabla \varphi_k \rangle dz +\int\limits_{\partial D} u_k \underbrace{\frac{\partial \varphi_k}{\partial \nu}}_{=0} ds(z) + \lambda_k\int\limits_D u_k \varphi_k dz = 0.
\end{equation}
Note that the Neumann boundary conditions from \eqref{eq:nep} in the above.

Similarly, multiplying the solution $\varphi_k$ of the Neumann problem \eqref{eq:nep} to the Eqn \eqref{eq:helm1} and repeating the above integration parts yields, 
\begin{equation}
\label{eq:pf2}
-\int\limits_D \langle \nabla u_k, \nabla \varphi_k \rangle dz +\int\limits_{\partial D} g_k \frac{\partial u_k}{\partial \nu} ds(z) + \int\limits_D a(z) u_k \varphi_k dz = 0.
\end{equation}
Formula \eqref{eq:arep} follows by subtracting \eqref{eq:pf1} from \eqref{eq:pf2}. 
\end{proof}

\section{Mathematical Description of Heart and Lungs Phantom}
To describe the phantom for the heart and lungs, we define the following sets of points on the domain $D$:
\begin{equation}
\begin{aligned}
     &s_h  = \left\{ (x,y)\in D ~\text{s.t}~ \sqrt{e_{h,1}(x-c_{h,1})^2 + e_{h,2}(y-c_{h,2})^2 }<0.2 \right\} \\\
    &s_{l_1}   = \left\{ (x,y)\in D ~\text{s.t}~ \sqrt{e_{l_1,1}\left(\left(\cos(\alpha)x_1+\sin(\alpha)x_2\right)-c_{l_1,1}\right)^2 + e_{l_1,2}\left(\left(-\sin(\alpha)x+\cos(\alpha)y\right) -c_{l_1,2}\right)^2} <0.5\right\}\\
    &s_{l_2}   = \left\{ (x,y)\in D ~\text{s.t}~ \sqrt{e_{l_2,1}\left(\left(\cos(\alpha)x_1+\sin(\alpha)x_2\right)-c_{l_2,1}\right)^2 + e_{l_2,2}\left(\left(-\sin(\alpha)x+\cos(\alpha)y\right) -c_{l_2,2}\right)^2} <0.4\right\}
\end{aligned}
\end{equation}
Here, $e_{h,1}=0.8$, $e_{h,2}=1$, $e_{l_1,1}=3$, $e_{l_1,2}=1$, $e_{l_2,1}=3$, and $e_{l_2,2}=1$ represent the eccentricities of the ellipses describing the heart and lungs. The center locations of the heart and lungs are given by $c_{h,1}=-0.1$, $c_{h,2}=0.4$, $c_{l_1,1}=0.5$, $c_{l_1,2}=0.2$, $c_{l_2,1}=-0.6$, and $c_{l_2,2}=0.1$ and the orientation of the lungs by $\alpha=\frac{\pi}{7}$.
Then the body conductivity is defined as,
\[ a(x,y)=
\begin{cases} 
      a_h & (x,y)\in s_h \\
      a_{l_1} & (x,y)\in s_{l_1} \\
      a_{l_2} & (x,y)\in s_{l_2} \\
      a_b  & \text{else}
   \end{cases}
\]
with $a_h=2$, $a_{l_1}=a_{l_2}=0.7$, $a_b=1$.

The training coefficients are obtained by adding $8\%$ white noise to all the parameters above. Specifically, given $y=[e_{h,1}, e_{h,2}, e_{l_1,1}, e_{l_1,2},e_{l_2,1}, e_{l_2,2}, c_{h,1}, c_{h,2},c_{l_1,1},c_{l_1,2}, c_{l_2,1}, c_{l_2,2}, a_h, a_{l_1}, a_{l_2}]$, we define the perturbed version of the parameter vector $y$ as $\Tilde{y}=y(1+0.08\xi)$, where $\xi\sim \mathcal{N}(0,1)$ is a random variable drawn from the standard normal distribution. 
The coefficient is then defined as:
\[ \Tilde{a}(x,y)=
\begin{cases} 
      \Tilde{a}_h & (x,y)\in \Tilde{s}_h \\
      \Tilde{a}_{l_1} & (x,y)\in \Tilde{s}_{l_1} \\
      \Tilde{a}_{l_2} & (x,y)\in \Tilde{s}_{l_2} \\
      \Tilde{a}_b  & \text{else}
   \end{cases}
\]
where $\Tilde{s}_h$, $\Tilde{s}_{l_1}$ and $\Tilde{s}_{l_2}$ are defined as above, but with the parameters replaced by their corresponding perturbed values in $\Tilde{y}$.

\section{Architecture and Training Details}
\label{app:tadet}
Below, details concerning the model architectures and training are discussed.
The implementation of the experiments of the paper is realized within the PyTorch framework and available at 
\begin{itemize}
\item[$\blacksquare$] \url{https://github.com/mroberto166/nio.git}
\end{itemize}

\subsection{Architecture Details}
\label{app:numex}
    \subsubsection{Feed Forward Dense Neural Networks}
    Given an input $y \in \R^m$, a feed-forward neural network (also termed as a multi-layer perceptron) transforms it to an output, through a layer of units (neurons) which compose of either affine-linear maps between units (in successive layers) or scalar nonlinear activation functions within units \cite{DLbook}, resulting in the representation,
    \begin{equation}
    \label{eq:ann1}
    u_{\theta}(y) = C_{L_t} \circ\sigma \circ C_{{L_t}-1}\ldots\circ\sigma \circ C_2 \circ \sigma \circ C_1(y).
    \end{equation} 
    Here, $\circ$ refers to the composition of functions, and $\sigma$ is a scalar (nonlinear) activation function. 
    For any $1 \leq \ell \leq L_t$, we define
    \begin{equation}
    \label{eq:C}
    C_\ell z_\ell = W_\ell z_\ell + b_\ell, ~ {\rm for} ~ W_\ell \in \R^{d_{\ell+1} \times d_\ell}, z_\ell \in \R^{d_\ell}, b_\ell \in \R^{d_{\ell+1}},\end{equation}
    and denote, 
    \begin{equation}
    \label{eq:theta}
    \theta = \{W_\ell, b_\ell\}_{\ell=1}^{L_t},
    \end{equation} 
    to be the concatenated set of (tunable) weights for the network. 
    Thus, in the machine learning terminology, a feed-forward neural network \eqref{eq:ann1} consists of an input layer, an output layer, and $L_t$ hidden layers with $d_\ell$ neurons, $1<\ell<L_t$. In all numerical experiments, the trunk net of DeepONet is a feed-forward neural network. Moreover, we consider a uniform number of neurons across all the layers of the network
    $d_\ell = d_{\ell-1} = d $, $ 1<\ell<L_t$.
    
    \subsubsection{Fully Convolutional Neural Network}
    Fully convolutional neural networks are a special class of convolutional networks which can be evaluated for virtually any resolution of the input. We use them as a strong baseline for PDE inverse problems in the results presented in Table \ref{tab:res}. The networks consist of an \textit{encoder} and \textit{decoder}, both defined by a composition of linear and nonlinear transformations:
    \begin{equation}
    \begin{aligned}
        E_{\theta_e}(y) &= C^e_L \circ\sigma \circ C^e_{L-1}\ldots\circ\sigma \circ C^e_2 \circ \sigma \circ C^e_1(y),\\
        D_{\theta_d}(z) &= C^d_L \circ\sigma \circ C^d_{L-1}\ldots\circ\sigma \circ C^d_2 \circ \sigma \circ C^d_1(z),\\
        u_\theta(y) &= D_{\theta_d} \circ  E_{\theta_e}(y).
    \end{aligned}
    \end{equation}
    The affine transformation $C_\ell$ commonly corresponds to a \textit{convolution} operation in the encoder and  \textit{transposed convolution} (also known as \textit{deconvolution}) in the decoder.
    
    The (de)convolution is performed with a kernel  $W_\ell\in \R^{k_\ell \times k_\ell}$, stride $s$, and padding $p$. It takes as input a tensor  $z_\ell \in \R^{w_\ell\times h_\ell\times c_\ell }$ with $c_\ell$ being the number of input channels, and computes $z_{\ell+1} \in \R^{w_{\ell+1}\times h_{\ell+1}\times c_{\ell+1} }$. Therefore, a (de)convolutional affine transformation can be uniquely identified with the tuple $(k_\ell, s, p, c_\ell, c_{\ell+1})$.
    
\begin{figure}[ht!]
\begin{center}
\centerline{\includegraphics[width=\columnwidth]{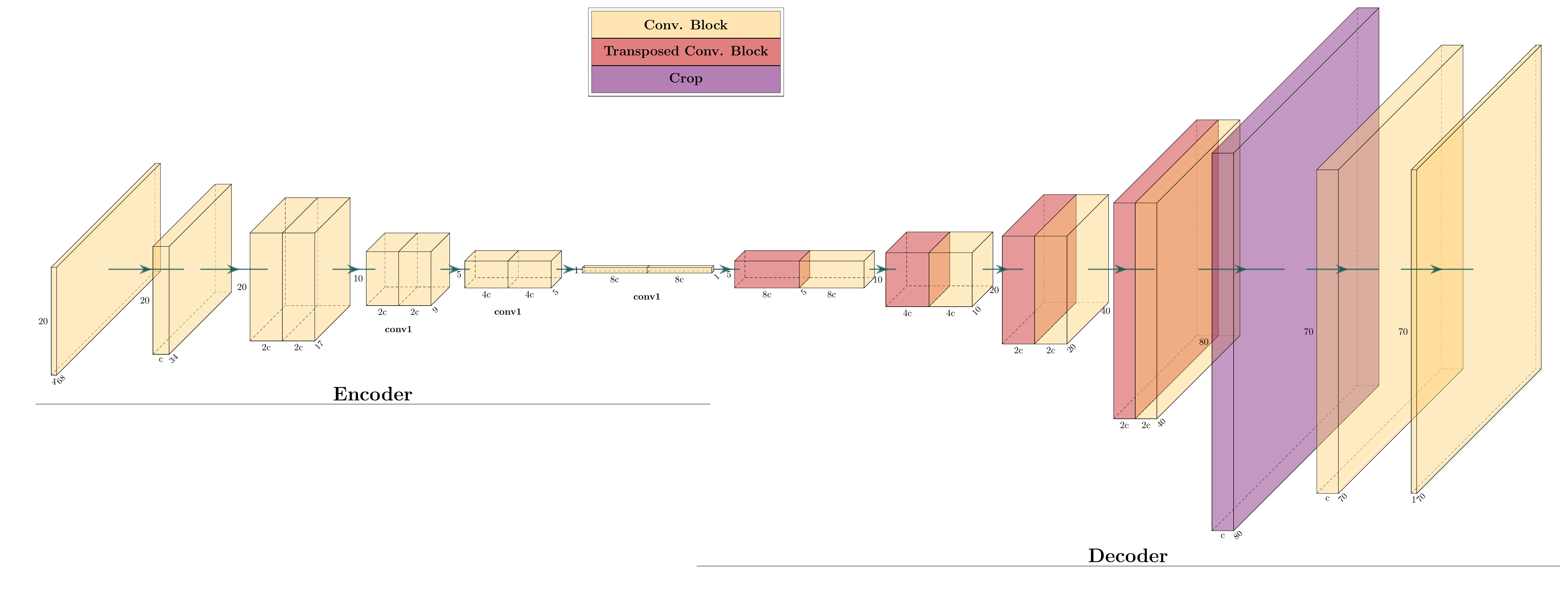}}
\vspace{-0.5cm}
    \caption{Schematic representation of the Fully-Convolutional Neural Network (FCNN) architecture used for the Calder\'{o}n problem with Trigononmetric coefficients and for the Inverse Wave Scattering with Helmholtz Equation.}
\label{fig:PoissonFCNN}
\end{center}
\end{figure}

\begin{figure}[ht!]
\begin{center}
\centerline{\includegraphics[width=\columnwidth]{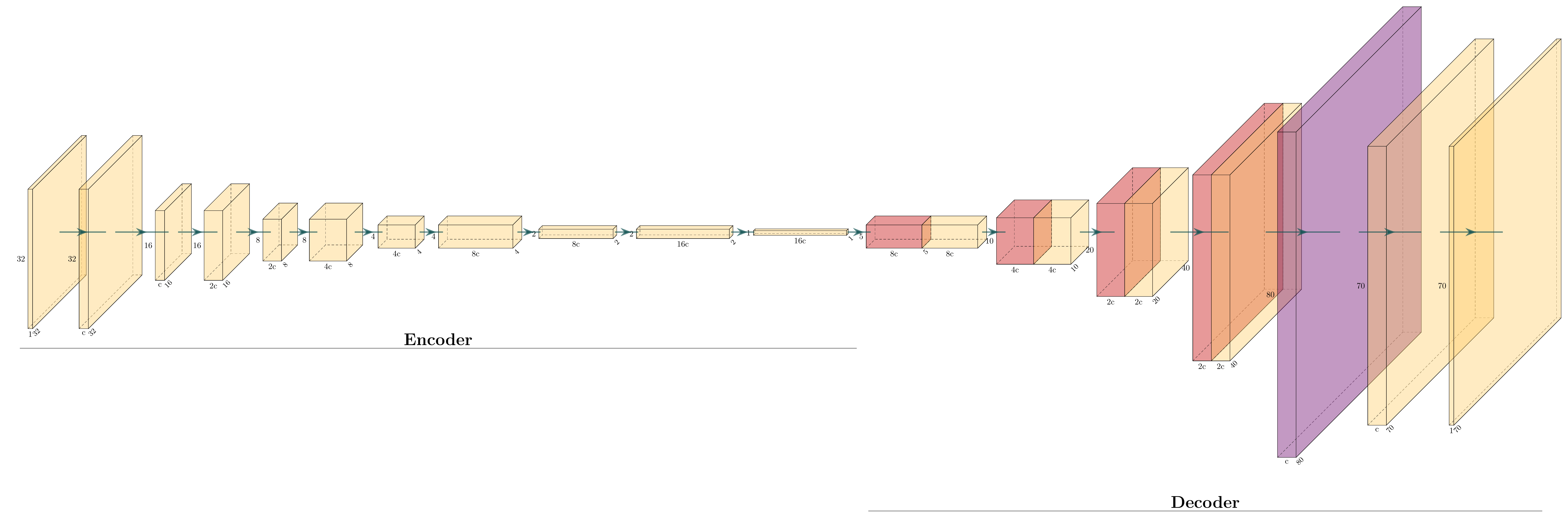}}

    \caption{Schematic representation of the Fully-Convolutional Neural Network architecture used for the optical imaging for the Calder\'on Problem with Heart\&Lungs phantom.}
\label{fig:EITFCNN}
\end{center}
\end{figure}

\begin{figure}[ht!]
\begin{center}
\centerline{\includegraphics[width=\columnwidth]{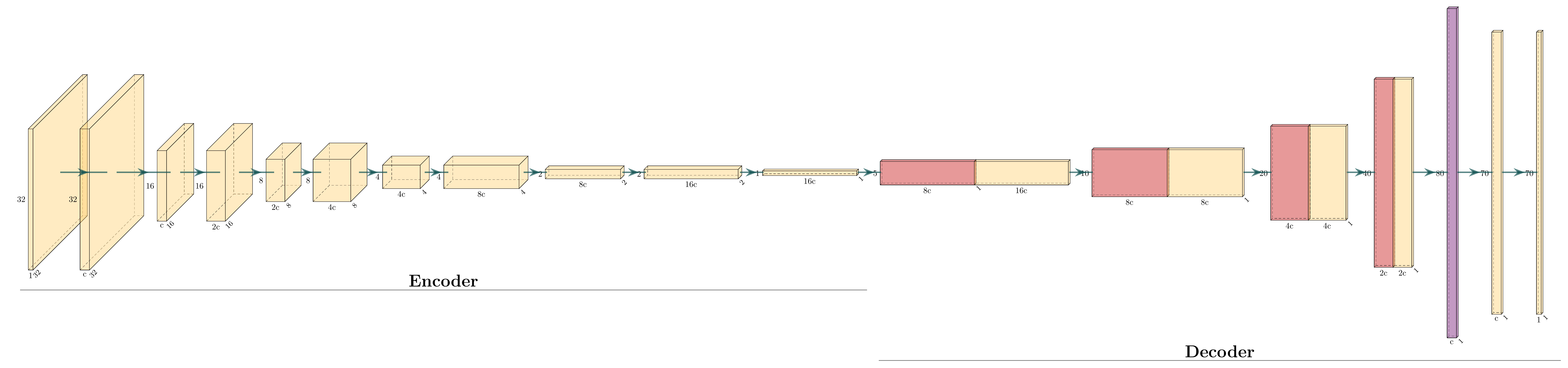}}

    \caption{Schematic representation of the Fully-Convolutional Neural Network architecture used for the optical imaging for the radiative transport Equation.}
\label{fig:RadFCNN}
\end{center}
\end{figure}

\begin{figure}[ht!]
\begin{center}
\centerline{\includegraphics[width=\columnwidth]{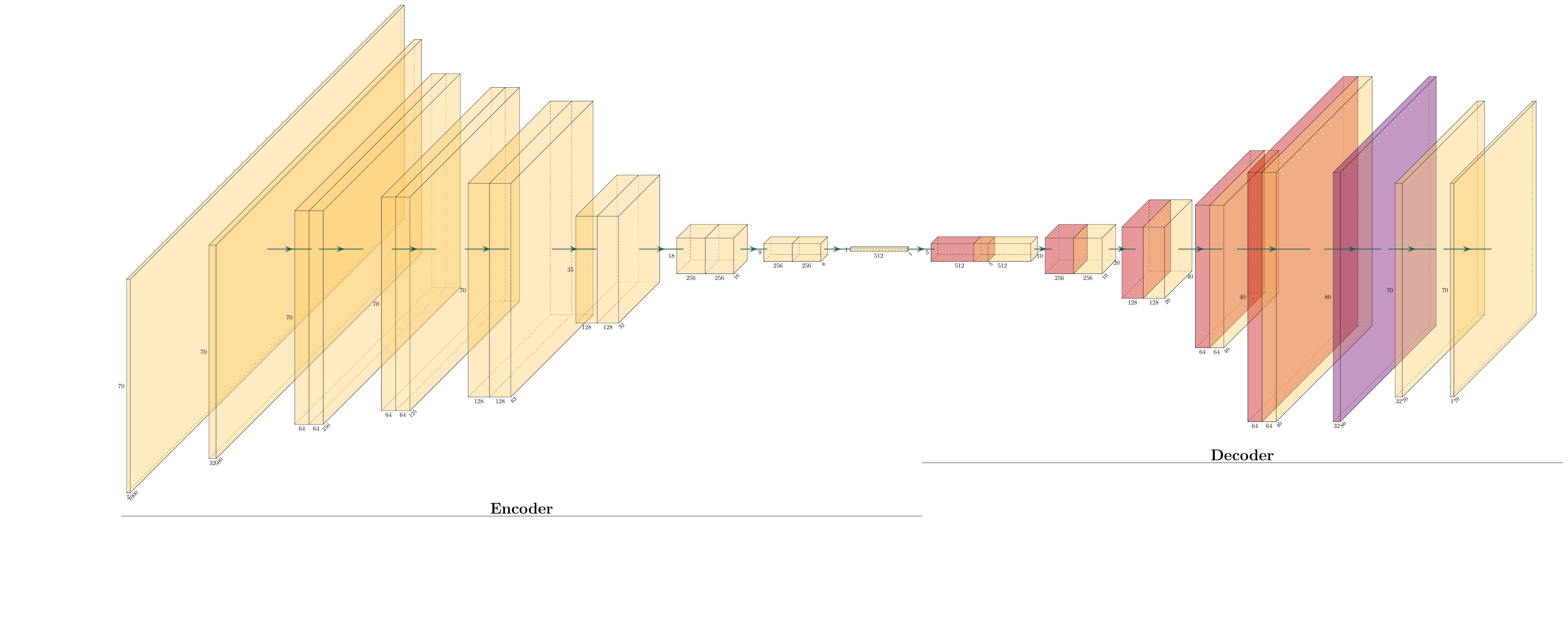}}
\vspace{-1cm}
    \caption{Schematic representation of the Fully-Convolutional Neural Network architecture used for the seismic imaging problems.}
\label{fig:WfiFCNN}
\end{center}
\end{figure}

    A visual representation of the convolutional architectures used for the benchmark problems is depicted in Figures \ref{fig:PoissonFCNN},  \ref{fig:EITFCNN}, \ref{fig:RadFCNN},\ref{fig:WfiFCNN}. The \textit{convolutional block} (or \textit{transposed convolution block}) is the composition of a convolution (or transposed convolution) operation, batch normalization, and activation function (Leaky ReLU).  The cropping operation involves adding negative padding to the edges of a tensor to achieve the desired output width and height. The number of channels $c$ is selected with cross-validation. The architecture used for seismic imagining is referred to as InversionNet in~\cite{deng2021openfwi}.

\begin{figure}[ht!]
\begin{center}
\centerline{\includegraphics[width=0.6\columnwidth]{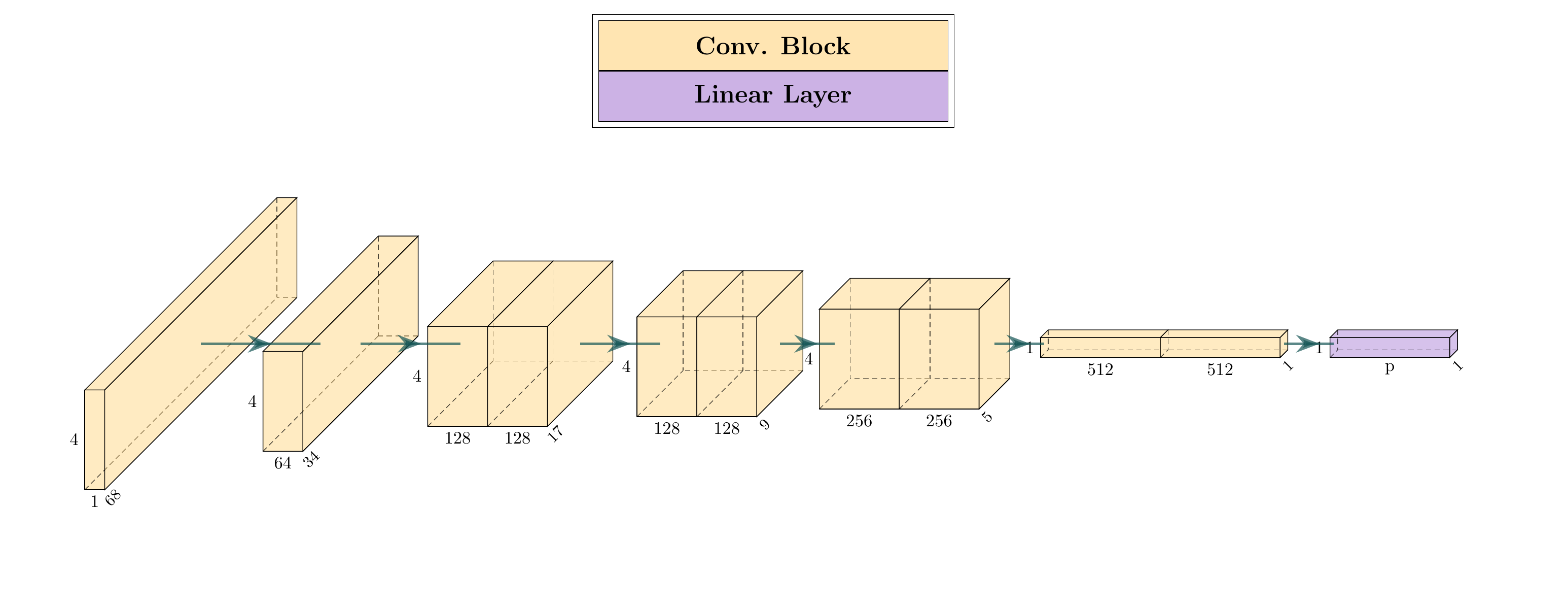}}
\vspace{-0.5cm}
    \caption{Schematic representation of the NIO-BranchNet architecture used for the Calder\'{o}n problem with Trigononmetric coefficients and for the Inverse Wave Scattering with Helmholtz Equation}
\label{fig:PoissonBranch}
\end{center}
\end{figure}

\begin{figure}[ht!]
\begin{center}
\centerline{\includegraphics[width=\columnwidth]{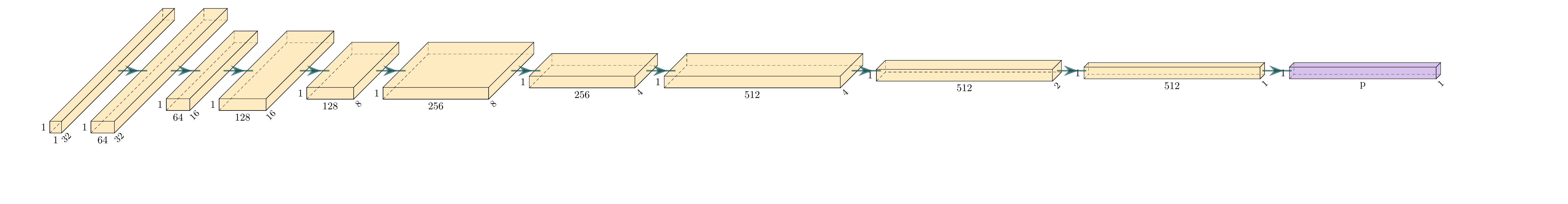}}
\vspace{-0.5cm}
    \caption{Schematic representation of the NIO-BranchNet architecture used for the optical Imaging with Radiative transport Equation and Calder\'on Problem with Heart\&Lungs phantom.}
\label{fig:RadBranch}
\end{center}
\end{figure}

\begin{figure}[ht!]
\begin{center}
\centerline{\includegraphics[width=0.8\columnwidth]{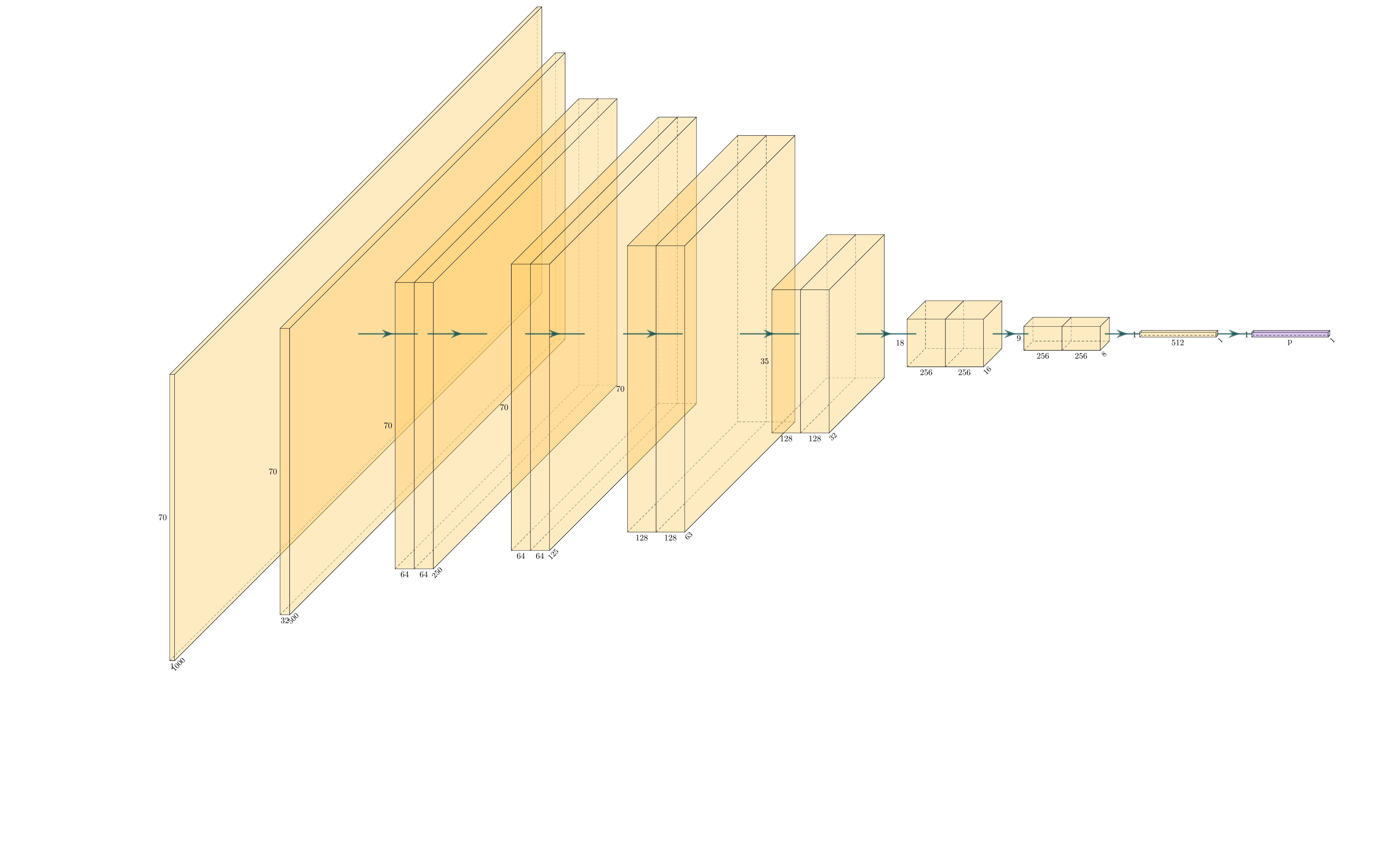}}
\vspace{-1.5cm}
    \caption{Schematic representation of the NIO-BranchNet architecture used for the seismic imaging.}
\label{fig:BranchWave}
\end{center}
\end{figure}

    \subsubsection{DeepONet}
    \label{sec:don}
    The architectures of the branch and trunk are chosen according to the benchmark addressed. 
    In particular, we employ standard feed-forward neural networks as trunk-net in all the experiments. In contrast, the branch is obtained as a composition of the \textit{encoder} of the fully convolutional networks depicted in figures \ref{fig:PoissonFCNN}, \ref{fig:EITFCNN} \ref{fig:RadFCNN} and \ref{fig:WfiFCNN}, and a linear transformation from $\R^n$ to $\R^p$, where $n$ denotes the number of channels in the last layer of the encoder and $p$ the number of basis functions. Moreover, $c=32$ for the seismic imaging and $c=64$ for all the other benchmarks. 
    
    Hence, the architecture of the branch is fixed. The number of the trunk hidden layers $L_t$, units $d$, and $p$ are chosen through cross-validation. On the other hand, the activation function $\sigma$ is chosen to be a leaky ReLU for both the branch and the trunk.
    
    \subsubsection{Fourier Neural Operator}
    We use the implementation of the FNO model provided by the authors of \cite{FNO}.
    Specifically,  the projection $Q$ to the target space is performed by a shallow neural network with a single hidden layer with $128$ neurons and $GeLU$ activation function. The same activation function is also used for all the Fourier layers. 
    Moreover, $b_\ell(x)=0$, for all $\ell=1,\ldots, T$ and the weight matrix $W_\ell$ used in the residual connection derives from a convolutional layer defined by $(k_\ell=1, s=1, p=0, c_{\ell}=d_v, c_{\ell+1}=d_v)$, for all $1<\ell<T$.

    \subsubsection{Neural Inverse Operator}
    In all numerical experiments, the proposed architecture is constructed by combining the DeepONet and Fourier Neural Operator by means of the lifting operator $R$ defined in equation \eqref{eq:r2}.

    The implementation of DeepONet follows the same description as outlined in Section \ref{sec:don}. However, the branch structure differs from the \textit{encoder} structure depicted in figures \ref{fig:PoissonFCNN}, \ref{fig:EITFCNN}, \ref{fig:RadFCNN}, and \ref{fig:WfiFCNN}. In the proposed NIO architecture, the channel mixing is performed downstream through the operator $\eM$. Specific details regarding the branch-net architectures used in NIO for the benchmark problems can be found in Figures \ref{fig:PoissonBranch}, \ref{fig:RadBranch}, and \ref{fig:BranchWave}. For instance, for the seismic imaging, the branch architecture is the same as the \textit{encoder} shown in Figure \ref{fig:WfiFCNN}, but only a single input channel is used instead of five, following the same rationale as mentioned above.
    Overall, the model includes the following hyperparameters: the number of layers $L_t$ and neurons $d$ of the DeepONet trunk, the number of basis functions $p$, and the lifting dimension $d_v$, the number of Fourier layers $T$ and number of (truncated)-Fourier coefficients $k$, of FNO.

\subsection{Training Details}
The training of the models, including the baselines, is performed with the ADAM optimizer, with a learning rate $\eta$ for 1000 epochs (250 epochs in the Seismic imaging problem) and minimizing the $L^1$-loss function. We also use a step learning rate scheduler and reduce the learning rate of each parameter group by a factor $\gamma$ every epoch. We train the models in mini-batches of size $256$, and a weight decay of magnitude $w$ is used. Moreover, the input and output data are transformed with a suitable map before training. Observe that the testing error reported in Table~\ref{tab:res} has been obtained on the non-transformed output data.
We consider two different data transformations to preprocess the data:
\begin{enumerate}
    \item \textit{MinMax}.  This transformation involves scaling both the inputs and outputs to a range between -1 and 1:
    \begin{equation}
         \Tilde{f} = 2\frac{{f}  - m }{M - m} -1,
    \end{equation}
    where $M$ and $m$ are the maximum and the minimum value of $f$ across all the \textit{training samples}.
    \item \textit{log-MinMax}. This transformation is specifically used for Seismic Imaging problems. The input data are transformed according to the following equation:
    \begin{equation}
        \Tilde{f} = \log\left(\left|{f}\right|\right)\text{sign}\left({f}\right),
    \end{equation} and then, the obtained input and output scaled between $-1$ and $1$.
\end{enumerate}
All the parameters mentioned above, including the type of data transformation (\textit{Identity}, \textit{MinMax}, \textit{log-MinMax}, are chosen through cross-validation.

At every epoch, the relative $ L^1$ error is computed on the validation set, and the set of trainable parameters resulting in the lowest error during the entire process is saved for testing. Early stopping is used to interrupt the training if the best validation error does not improve after $50$ epochs. 

The cross-validation is performed by running a random search over a chosen range of hyperparameters values and selecting the configuration, realizing the lowest relative $ L^1$ error on the validation set. Overall, 50 hyperparameter configurations are tested for NIO and 30 for the baselines. The model size (minimum and maximum number of trainable parameters) covered in this search are reported in Table \ref{tab:params}.

The results of the random search, i.e., the best-performing hyperparameter configurations for each model and each benchmark, are reported in tables \ref{tab:don_params}, \ref{tab:fcc_params}, and \ref{tab:nio_params}.  
The FCNN hyperparameters reported in the table for the seismic imaging problem are those used in \cite{deng2021openfwi}. 

\begin{table}[ht!]
\centering
    \renewcommand{\arraystretch}{1.3} 
    \footnotesize{
        \begin{tabular}{ c  c c c c c c } 
        \toprule
          & \bfseries\makecell{Calder\'{o}n \\Problem \\ Trigonometric} 
            &\bfseries\makecell{Calder\'{o}n \\Problem \\ Heart\&Lungs } 
           & \bfseries\makecell{Inverse Wave \\Scattering} 
           & \bfseries\makecell{Radiative \\Transport} 
           & \bfseries\makecell{Seismic \\ Imaging \\ CurveVel - A} 
           & \bfseries\makecell{Seismic \\ Imaging \\ Style - A}\\
            \midrule
            \midrule
           \textbf{DONet} &\makecell{4.6M \\ 9.07M}&\makecell{4.6M \\ 9.07M}&\makecell{9.54M \\ 14.01M}&\makecell{9.54M \\ 14.01M}&\makecell{12.01M \\ 15.85M}&\makecell{12.01M \\ 15.85M}\\
            \midrule
            \textbf{FCNN}&\makecell{1.07M \\ 68.32M}&\makecell{1.07M \\ 68.32M}&\makecell{4.31M \\ 275.37M}&\makecell{2.48M \\ 39.53M} & \makecell{24.4M \\ 24.4M } & \makecell{24.4M \\ 24.4M }  \\
            \midrule
            \textbf{NIO}&\makecell{7.95M \\ 27.79M}&\makecell{10.6M \\ 50.76M}&\makecell{7.95M \\ 27.79M}&\makecell{9.6M \\ 10.3M}&\makecell{13.07M \\ 32.91M}&\makecell{13.07M \\ 32.91M}  \\
        \bottomrule
        \end{tabular}
        \caption{Minimum (Top sub-row) and maximum (Bottom sub-row) number of trainable parameters among the random-search hyperparameters configurations for all the models in every problem reported in Table \ref{tab:res} in the main text. }
        \label{tab:params}
    }
\end{table}

\begin{table}[t]
\centering
    \renewcommand{\arraystretch}{1.1} 
    \footnotesize{
                \begin{tabular}{ c  c c c c c c c c} 
        \toprule
          & $\eta$ 
            & $\gamma$
           & $w$
           & Data Trans
           & $p$
           & $L_t$
           &$d$
           & \makecell{Trainable\\ Params } 
           \\
            \midrule
            \midrule
            \bfseries\makecell{Calder\'{o}n Problem \\ Trigonometric}&0.001 &1.0 &0.0 &Identity &25 &8 &200 &4.84M\\ \midrule
\bfseries\makecell{Calder\'{o}n Problem \\ Heart\&Lungs}&0.001 &1.0 &0.0 &MinMax &100 &15 &500 &13.1M\\ \midrule
\bfseries\makecell{Inverse Wave Scattering }&0.001 &1.0 &1e-06 &Identity &1000 &12 &500 &8.32M\\ \midrule
\bfseries\makecell {Radiative transport} &0.001 &1.0 &0.0 &MinMax &100 &15 &500 &13.1M\\ \midrule
\bfseries\makecell {Seismic Imaging \\ CurveVel - A}&0.001 &0.98 &1e-06 &log-MinMax &400 &12 &500 &15.09M\\ \midrule
\bfseries\makecell{Seismic Imaging \\ Style - A}&0.001 &0.98 &1e-06 &log-MinMax &400 &12 &500 &15.09M\\

        \bottomrule
        \end{tabular}
        \caption{DeepONet best-performing hyperparameters configuration for different benchmark problems.}
        \label{tab:don_params}
    }
\end{table}

\begin{table}[t]
\centering
    \renewcommand{\arraystretch}{1.1} 
    \footnotesize{
        \begin{tabular}{ c  c c c c c c} 
        \toprule
          & $\eta$ 
            & $\gamma$
           & $w$
           & Data Trans
           & $c$
           & \makecell{Trainable\\ Params } 
           \\
            \midrule
            \midrule
            \bfseries\makecell{Calder\'{o}n Problem \\ Trigonometric}&0.001 &1.0 &0.0 &MinMax &16 &1.07M\\ \midrule
            \bfseries\makecell{Calder\'{o}n Problem \\ Heart\&Lungs}&0.001 &1.0 &0.0 &MinMax &128 &275.37M\\ \midrule
            \bfseries\makecell{Inverse Wave Scattering }&0.001 &1.0 &0.0 &MinMax &128 &68.32M\\ \midrule
            \bfseries\makecell {Radiative transport} &0.001 &1.0 &1e-06 &MinMax &16 &2.48M\\ \midrule
             \bfseries\makecell {Seismic Imaging \\ CurveVel - A}&0.001 &1 &1e-04 &log-MinMax & 64 &24.4M \\ \midrule
            \bfseries\makecell{Seismic Imaging \\ Style - A}&0.001 &1 &1e-04 &log-MinMax & 64 &24.4M\\ 
        \bottomrule
        \end{tabular}
        \caption{Fully convolutional neural network best-performing hyperparameters configuration for different benchmark problems.}
        \label{tab:fcc_params}
    }
\end{table}

\begin{table}[t]
\centering
    \renewcommand{\arraystretch}{1.1} 
    \footnotesize{
        \begin{tabular}{ c  c c c c c c c c c c c} 
        \toprule
          & $\eta$ 
            & $\gamma$
           & $w$
           & Data Trans
           & $p$
           & $L_t$
           &$d$ &$k$ &$d_v$&$L$
           & \makecell{Trainable\\ Params } 
           \\
            \midrule
            \midrule
            \bfseries\makecell{Calder\'{o}n Problem \\ Trigonometric}&0.001 &1.0 &1e-06 &Identity &100 &8 &100 &25 &32 &4 &12.06M\\ \midrule
            \bfseries\makecell{Calder\'{o}n Problem \\ Heart\&Lungs}&0.001 &1.0 &1e-06 &MinMax &100 &8 &100 &25 &32 &4 &14.74M\\ \midrule
            \bfseries\makecell{Inverse Wave Scattering }&0.001 &0.98 &0.0 &Identity &100 &8 &200 &16 &64 &4 &15.57M\\ \midrule
            \bfseries\makecell {Radiative transport} &0.001 &0.98 &1e-06 &MinMax &400 &4 &100 &32 &64 &4 &10.3M\\ \midrule
            \bfseries\makecell {Seismic Imaging \\ CurveVel - A}&0.001 &0.98 &1e-06 &MinMax &25 &4 &200 &16 &32 &3 &16.49M\\ \midrule
            \bfseries\makecell{Seismic Imaging \\ Style - A}&0.001 &0.98 &1e-06 &log-MinMax &100 &8 &200 &16 &64 &2 &16.49M\\ 

        \bottomrule
        \end{tabular}
        \caption{ Neural Inverse Operator best-performing hyperparameters configuration for different benchmark problems.}
        \label{tab:nio_params}
    }
\end{table}

\subsubsection{Sensitivity to Initialization of the Trainable Parameters}
Next, we consider the best performing NIO  model corresponding to the hyperparameters reported in Table \ref{tab:nio_params}. We then train the model again using multiple initializations of the model parameters, and each model is trained a total of 10 times. Table \ref{tab:initialization} presents the means and standard deviations of the 10 test median errors for various benchmark experiments.

From the results shown in Table \ref{tab:initialization}, we observe that NIO demonstrates remarkable robustness with respect to random initializations, exhibiting a significantly low standard deviation to mean ratio for the chosen benchmarks. However, it should be noted that the Calder\'on problem with trigonometric functions is an exception to this trend, as it appears to be quite sensitive to different initializations of the model parameters.

\begin{table}[ht]
\centering
    \renewcommand{\arraystretch}{1.3} 
    \footnotesize{
        \begin{tabular}{   c c c c  } 
        \toprule
          \bfseries\makecell{Calder\'{o}n \\Problem \\ Trigonometric} 
            &\bfseries\makecell{Calder\'{o}n \\Problem \\ Heart\&Lungs } 
           & \bfseries\makecell{Inverse Wave \\Scattering} 
           & \bfseries\makecell{Radiative \\Transport} \\
            \midrule
            \midrule
         $ 1.03 \pm 0.45\% $&$ 0.19 \pm 0.03\% $&$ 1.08 \pm 0.16\% $&$ 1.14 \pm 0.18\% $ \\
        \bottomrule
        \end{tabular}
        \caption{Means $\pm$ standard deviations of relative median $L^1$ test errors computed over 10 different NIO retrainings for different benchmarks. }
        \label{tab:initialization}
    }
\end{table}

\subsubsection{Sensitivity to the Number of Training Samples}
\begin{figure*}[htbp]
    \begin{subfigure}{0.5\textwidth}
        \centering
        \includegraphics[width=1\linewidth]{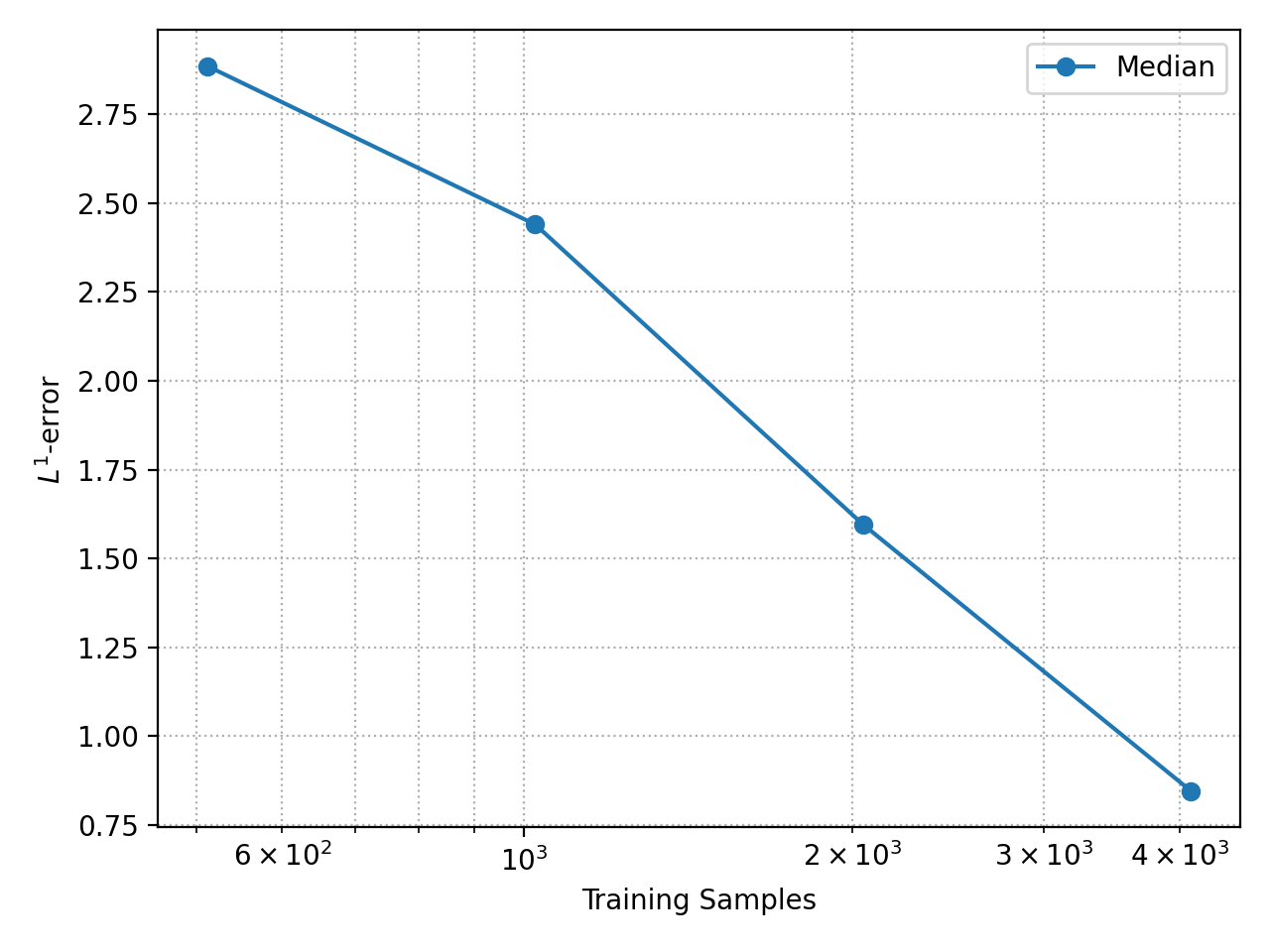}
         \caption{Calder\'{o}n Problem Trigonometric}
         \label{fig:ltilde:sine}
    \end{subfigure}
    \begin{subfigure}{0.5\textwidth}
        \centering
        \includegraphics[width=1\linewidth]{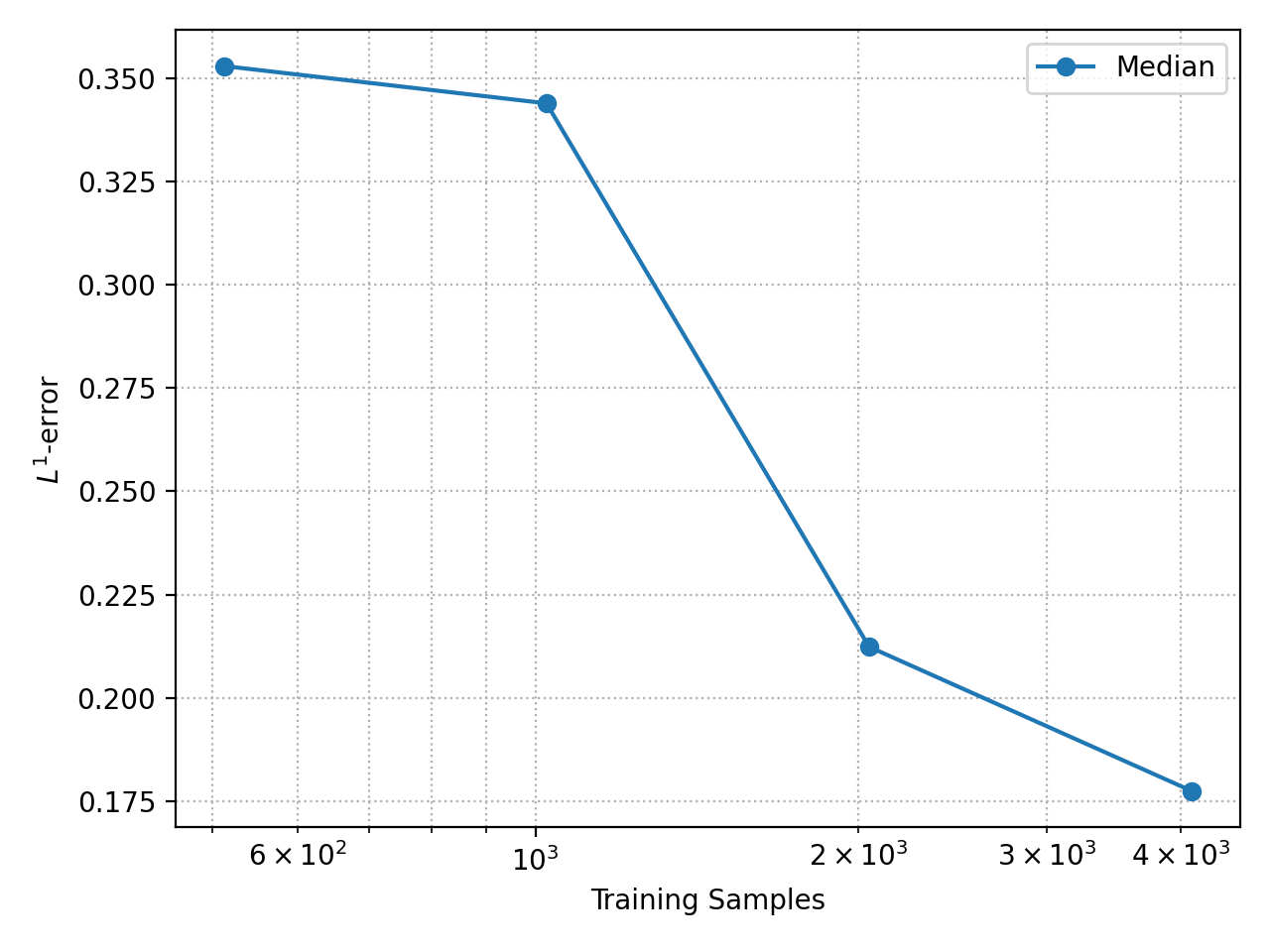}
        \caption{Calder\'{o}n Problem Heart\&Lungs}
        \label{fig:diffTS:eit}
    \end{subfigure}
    
    \begin{subfigure}{0.5\textwidth}
        \centering
        \includegraphics[width=1\linewidth]{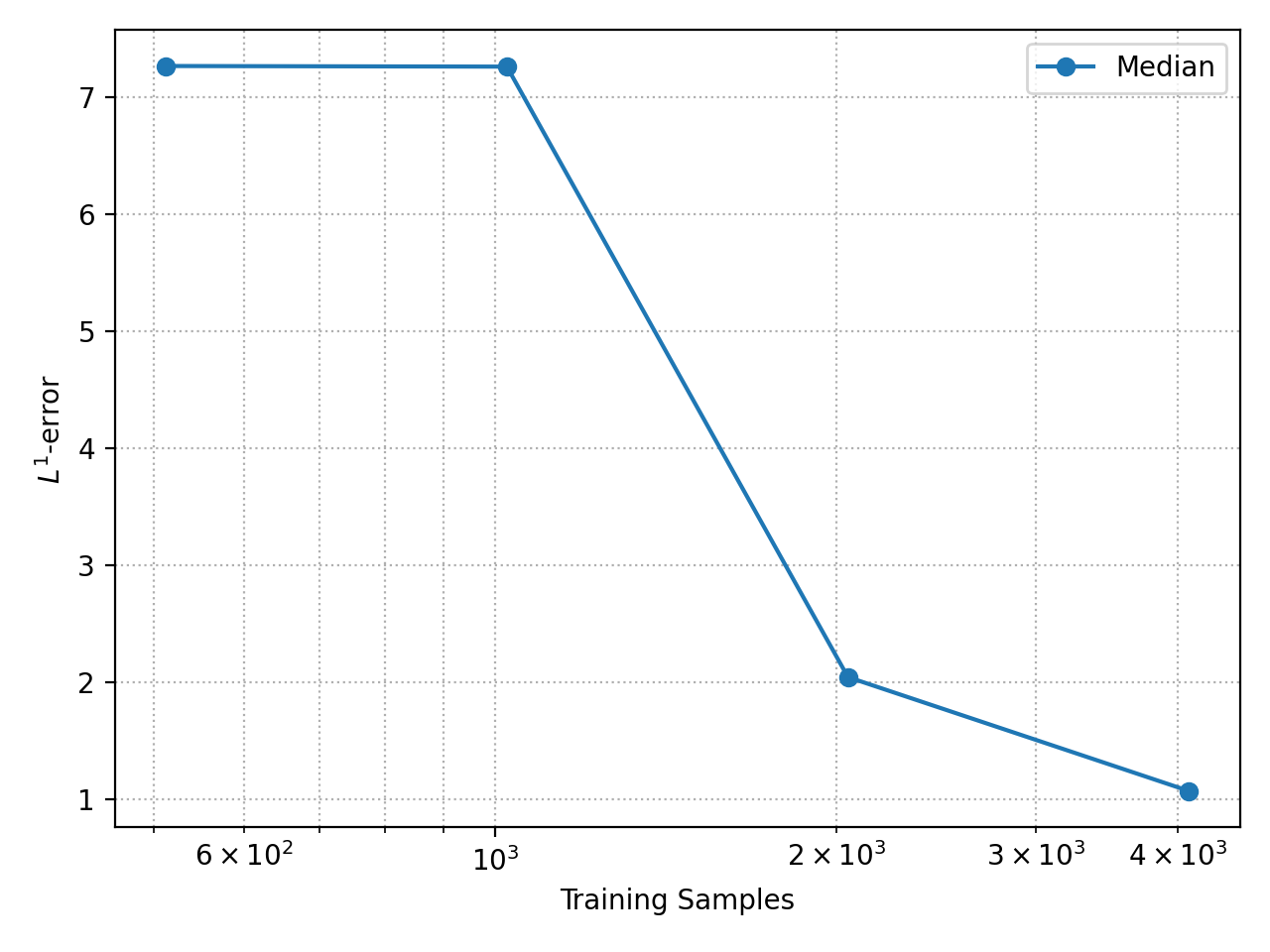}
         \caption{Inverse Wave Scattering }
         \label{fig:diffTS:helm}
    \end{subfigure}
    \begin{subfigure}{0.5\textwidth}
        \centering
        \includegraphics[width=1\linewidth]{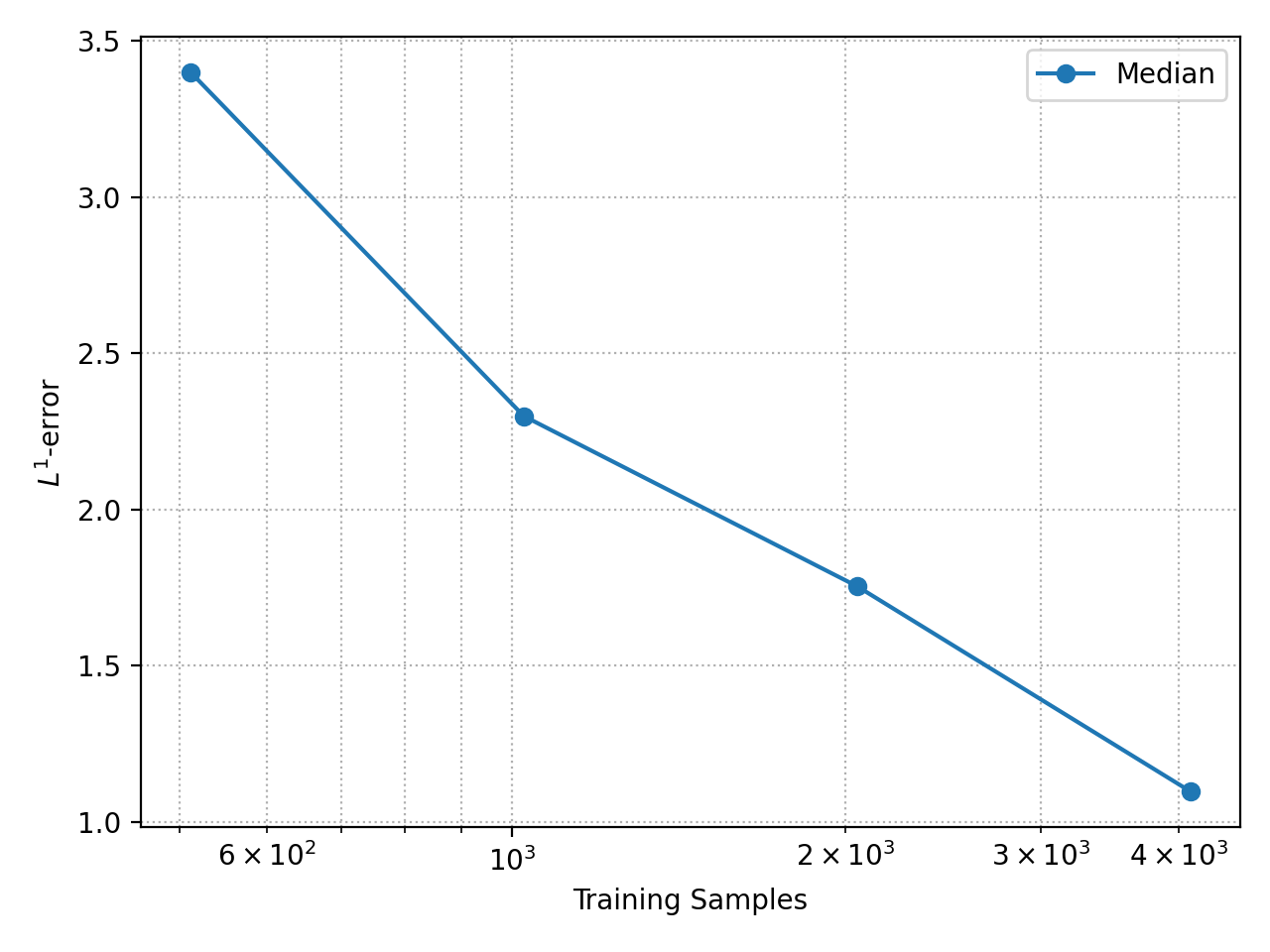}
         \caption{Radiative transport}
         \label{fig:diffTS:rad}
    \end{subfigure}

    \caption{Median of the $L^1$-error computed over testing samples  VS number of training samples for different benchmarks with NIO.}
\label{fig:diffTS}
\end{figure*}

Once again, we revisit the best-performing NIO  architectures for different benchmark problems and focus on varying the number of training samples and retraining the selected models accordingly. We then plot in Figure\ref{fig:diffTS} the median of the testing error as we change the cardinality of the training set. We observe that even for 512 training samples, the accuracy of the models, except for the Inverse Wave Scattering, remains very satisfactory. It should be observed that the number of training samples used in the benchmarks is very small compared to the state-of-the-art data-driven models used in this context.

\section{Further Experimental Results}
\label{app:addnum}
\subsection{Illustration of Results reported in Table \ref{tab:res}.}

\begin{figure*}[ht!]
    \begin{subfigure}{1\textwidth}
        \centering
        \includegraphics[width=1\linewidth]{{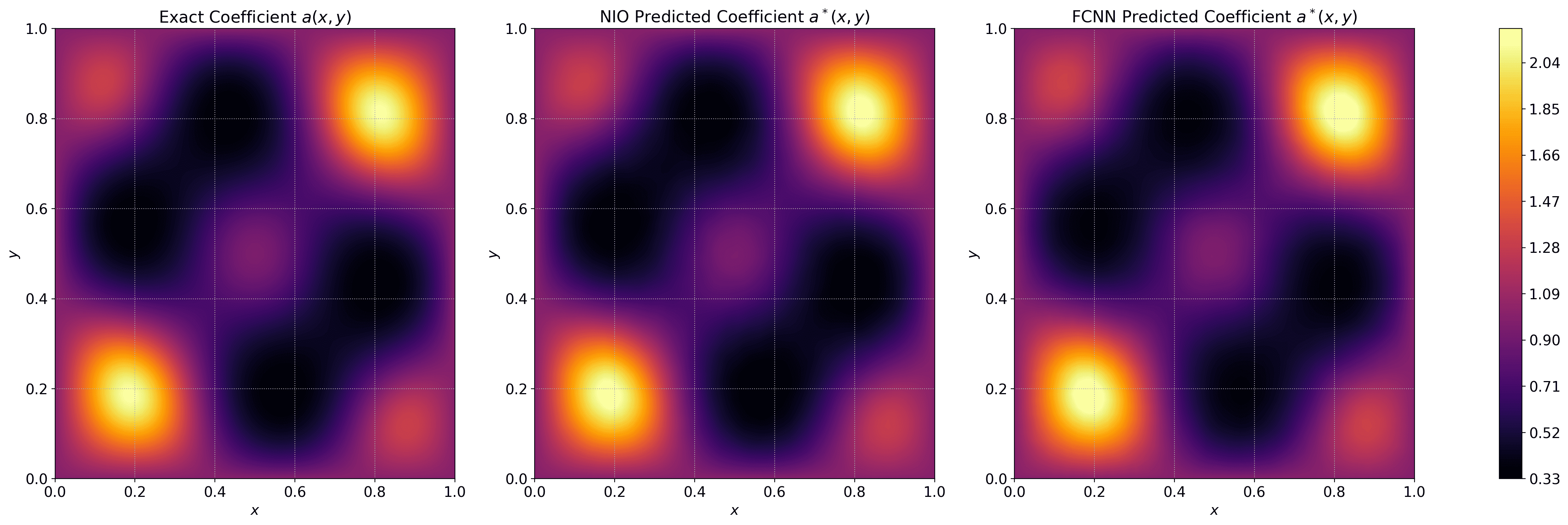}}
        \caption{Test Sample 1}
        \label{fig:calderon:sine:0}
    \end{subfigure}
    \begin{subfigure}{1\textwidth}
        \centering
        \includegraphics[width=1\linewidth]{{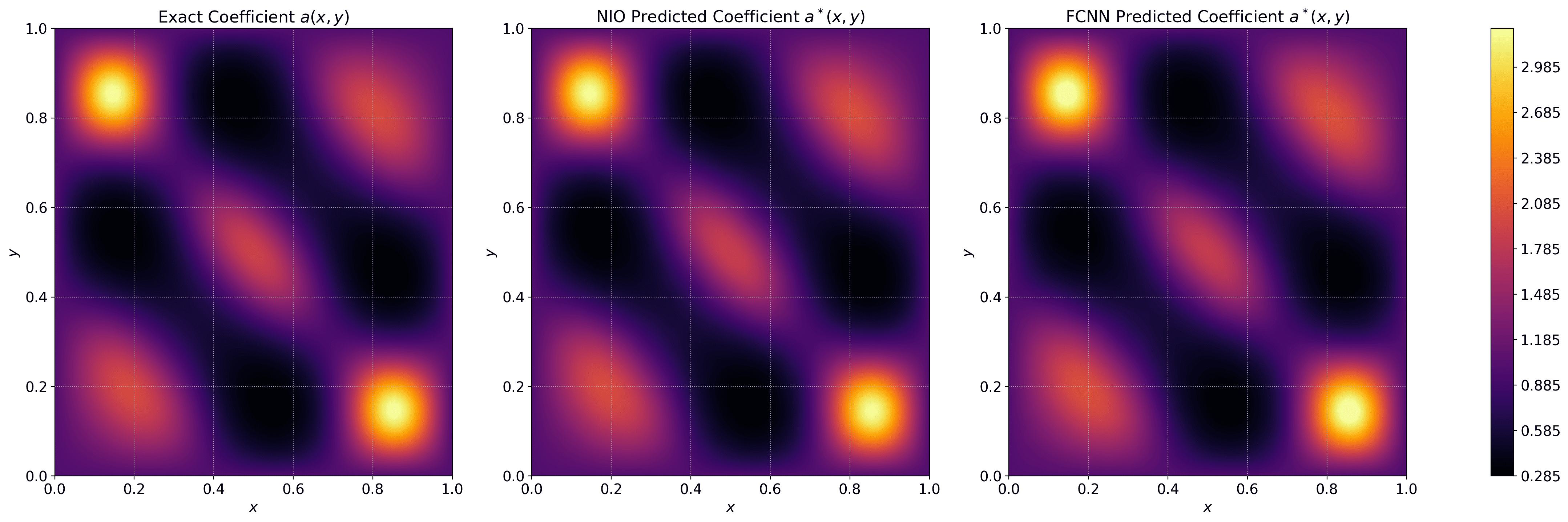}}
        \caption{Test Sample 2}
        \label{fig:calderon:sine:2}
    \end{subfigure}
    \caption{Exact and predicted coefficients for two different test samples (Rows) for the Calder\'{o}n problem with Trigonometric coefficients. Left Column: Ground Truth. Middle Column: NIO reconstruction. Right Column: FCNN Reconstruction.}
\label{fig:calderon:sine}
\end{figure*} 
\begin{figure*}[ht!]
    \begin{subfigure}{1\textwidth}
        \centering
        \includegraphics[width=1\linewidth]{{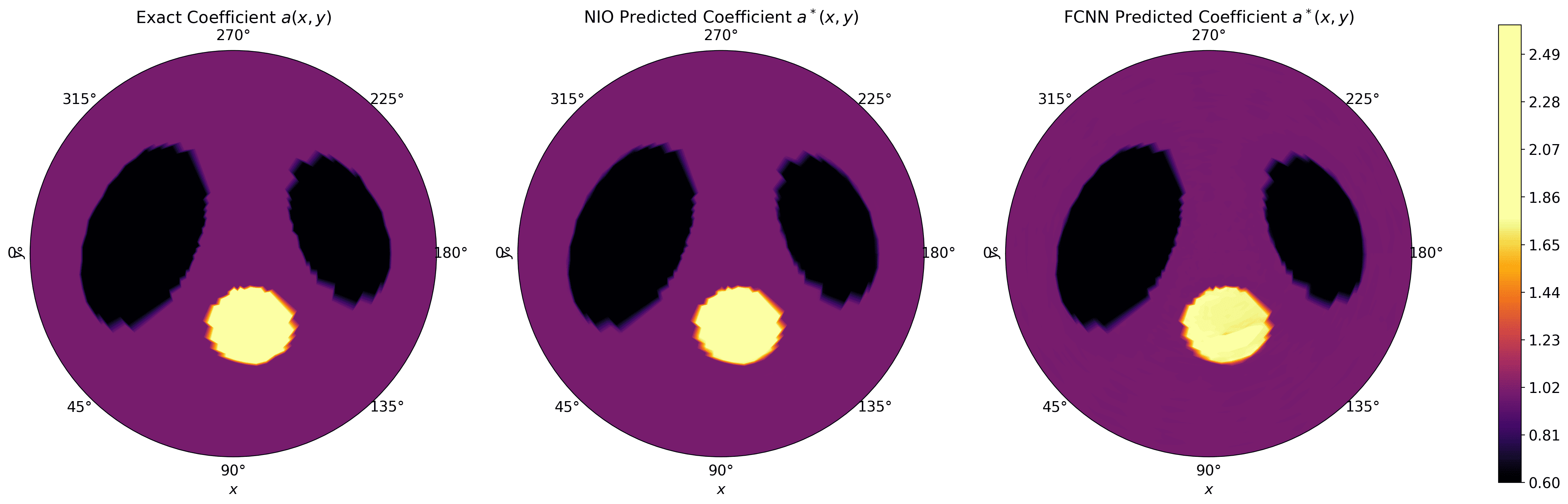}}
        \caption{Test Sample 1}
        \label{fig:calderon:eit:0}
    \end{subfigure}
    \begin{subfigure}{1\textwidth}
        \centering
        \includegraphics[width=1\linewidth]{{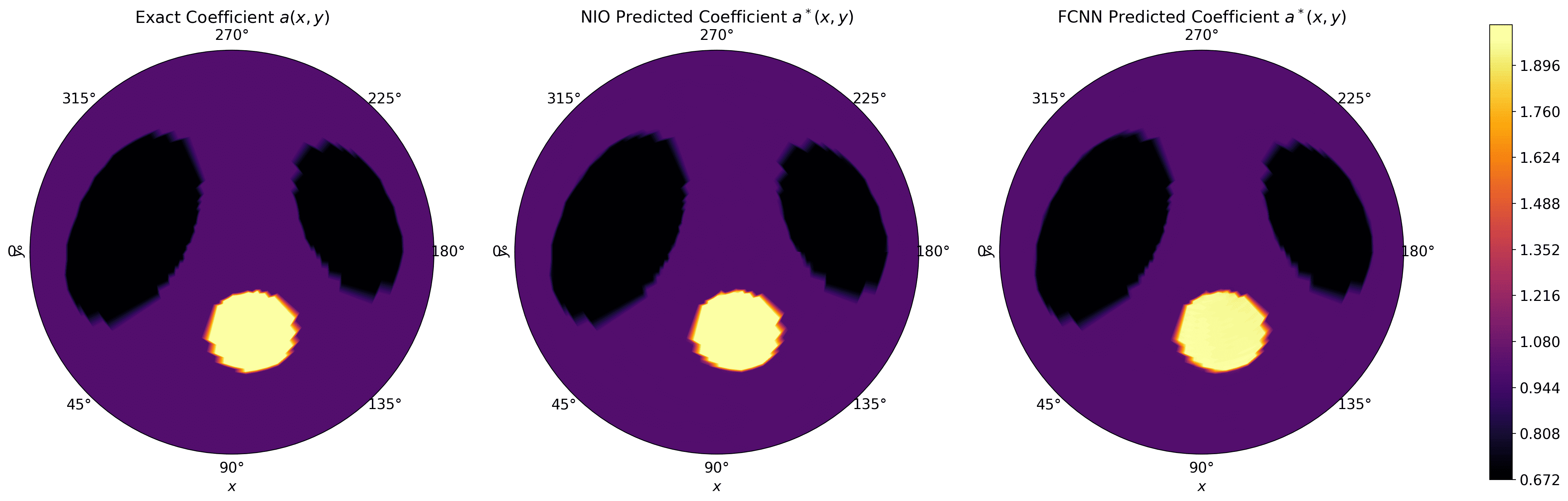}}
        \caption{Test Sample 2}
        \label{fig:calderon:eit:2}
    \end{subfigure}
    \caption{Exact and predicted coefficients for two different test samples (Rows) for the Calder\'{o}n problem with Heart\&Lungs Phantom. Left Column: Ground Truth. Middle Column: NIO reconstruction. Right Column: FCNN Reconstruction.}
\label{fig:calderon:eit}
\end{figure*} 

\begin{figure*}[ht!]

 \begin{subfigure}{1\textwidth}
        \centering
        \includegraphics[width=1\linewidth]{{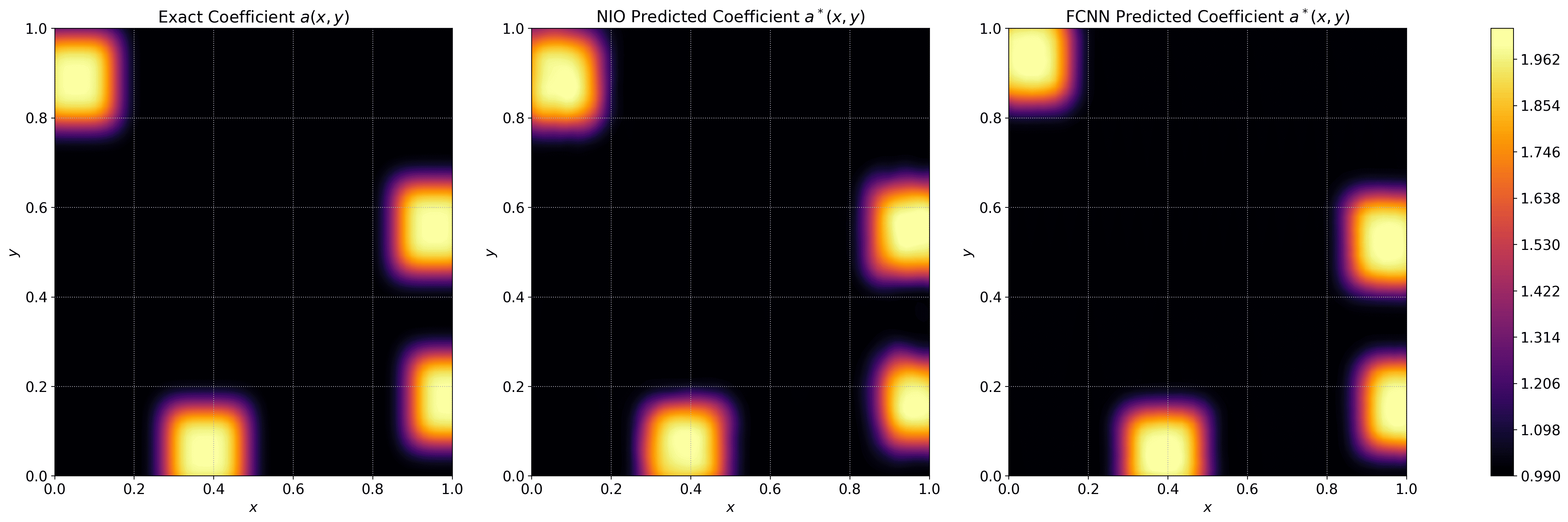}}
        \caption{Test Sample 1}
        \label{fig:calderon:helm:2}
    \end{subfigure}
    
    \begin{subfigure}{1\textwidth}
        \centering
        \includegraphics[width=1\linewidth]{{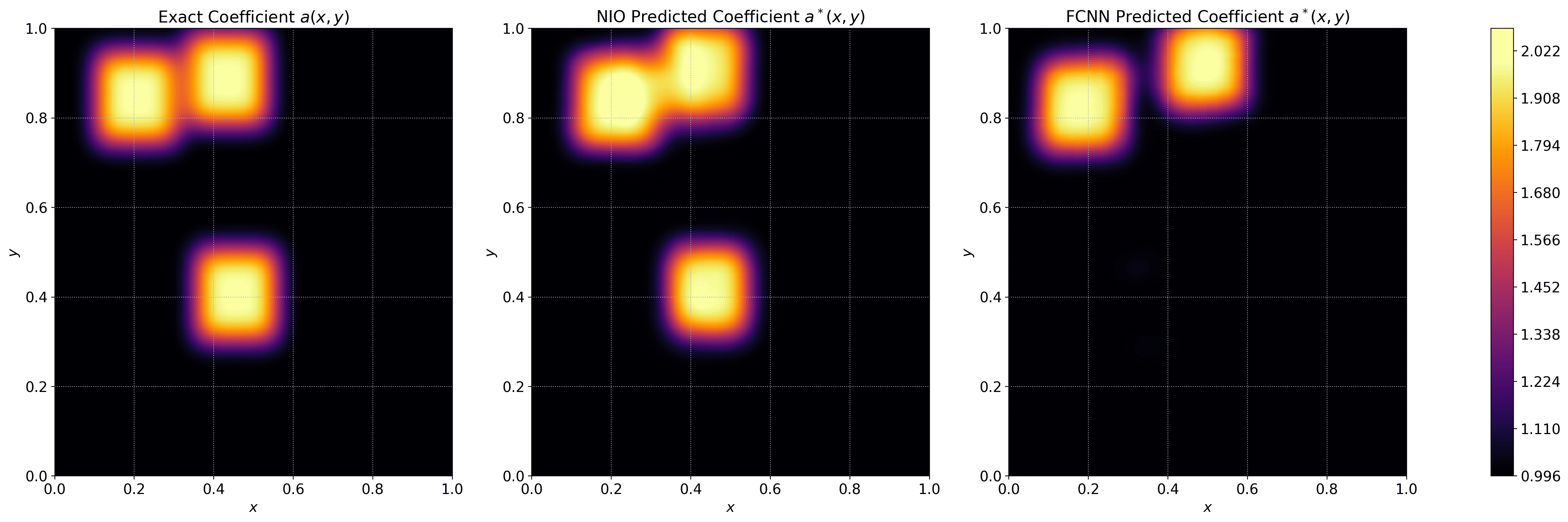}}
        \caption{Test Sample 2}
        \label{fig:calderon:helm:0}
    \end{subfigure}
    \caption{Exact and predicted coefficients for two different test samples (Rows) for the Inverse Wave Scattering with Helmholtz Equation. Left Column: Ground Truth. Middle Column: NIO reconstruction. Right Column: FCNN Reconstruction.}
\label{fig:helm}
\end{figure*} 

\begin{figure*}[ht!]
     \begin{subfigure}{0.48\textwidth}
        \centering
        \includegraphics[width=1\linewidth]{{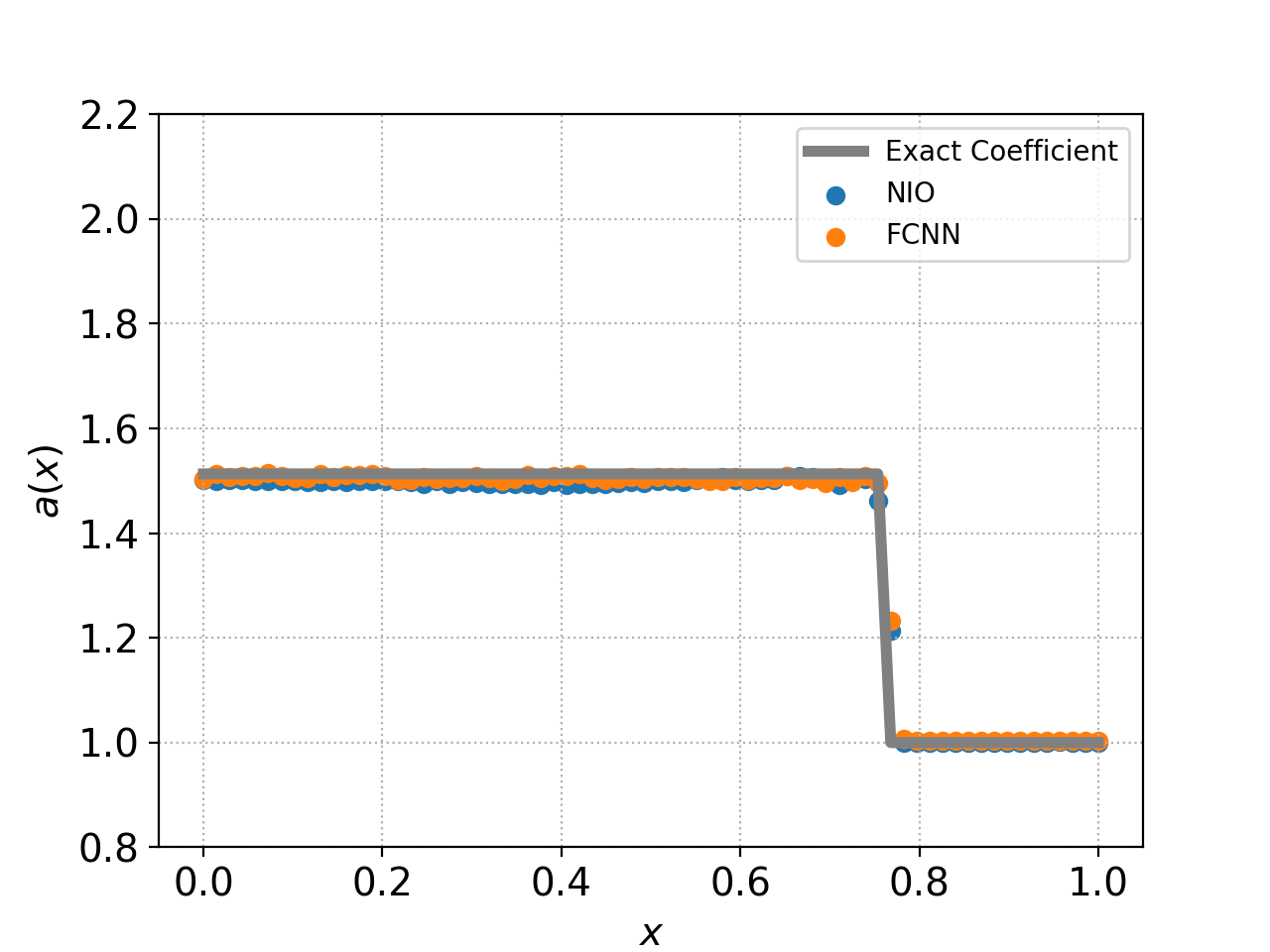}}
        \caption{Test Sample 1}
        \label{fig:radiative:2}
    \end{subfigure}
    \begin{subfigure}{0.48\textwidth}
        \centering
        \includegraphics[width=1\linewidth]{{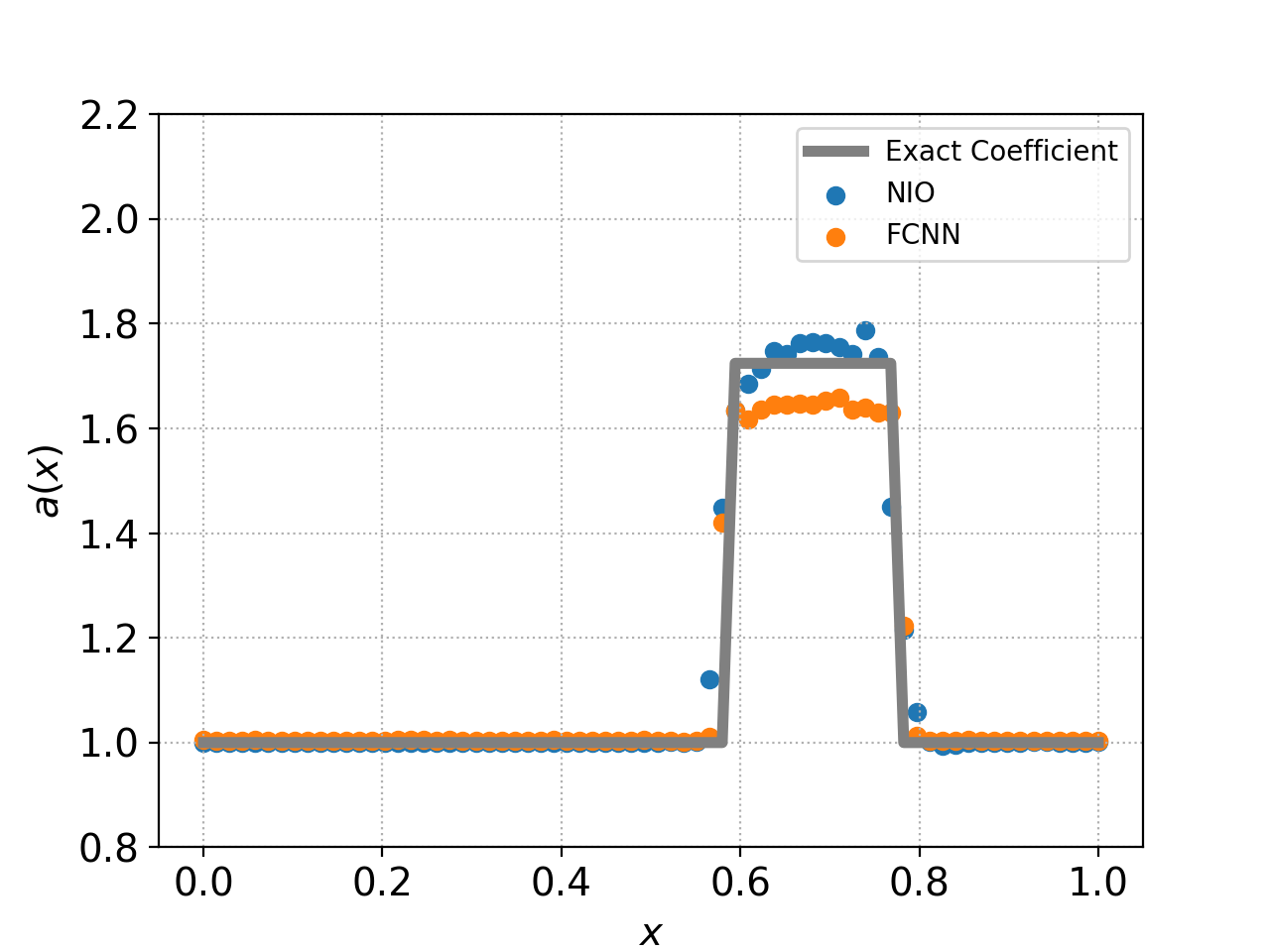}}
        \caption{Test Sample 2}
        \label{fig:radiative:1}
    \end{subfigure}
    \caption{Exact and predicted absorption coefficients for two different test samples, obtained with optical imaging for the radiative transport Equation.}
\label{fig:radiative}
\end{figure*} 

\begin{figure*}[htbp]
    \begin{subfigure}{1\textwidth}
        \centering
        \includegraphics[width=1\linewidth]{{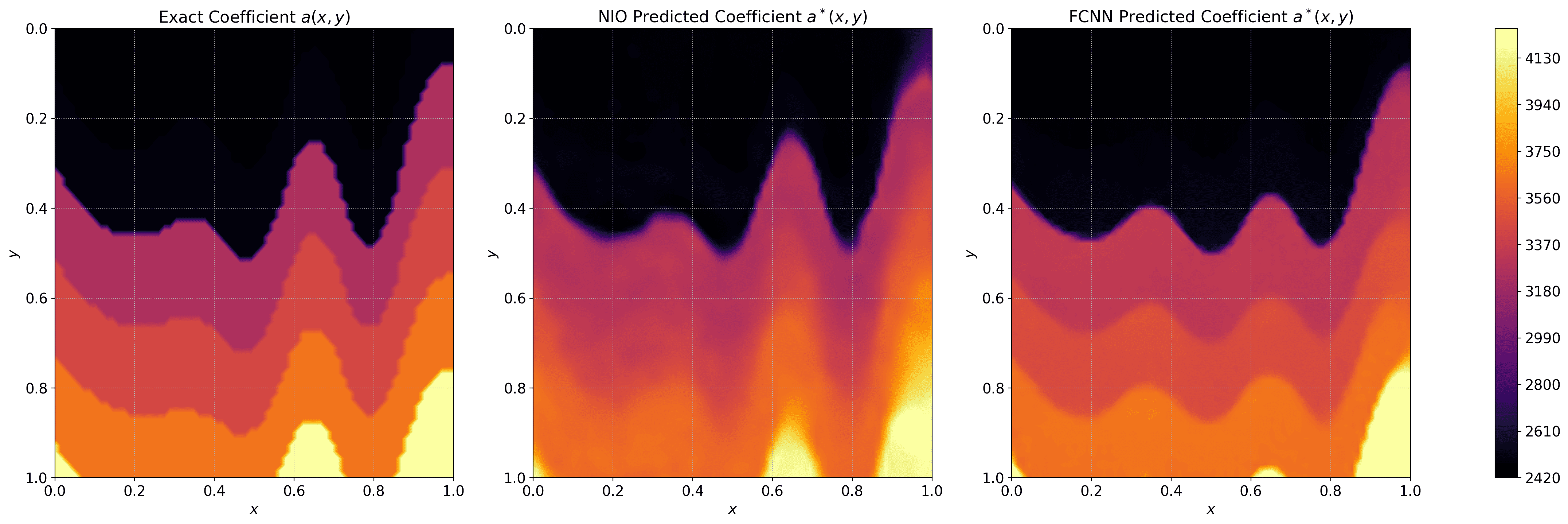}}
         \caption{Test Sample 1}
         \label{fig:wfi:curve:1}
    \end{subfigure}
    \begin{subfigure}{1\textwidth}
        \centering
        \includegraphics[width=1\linewidth]{{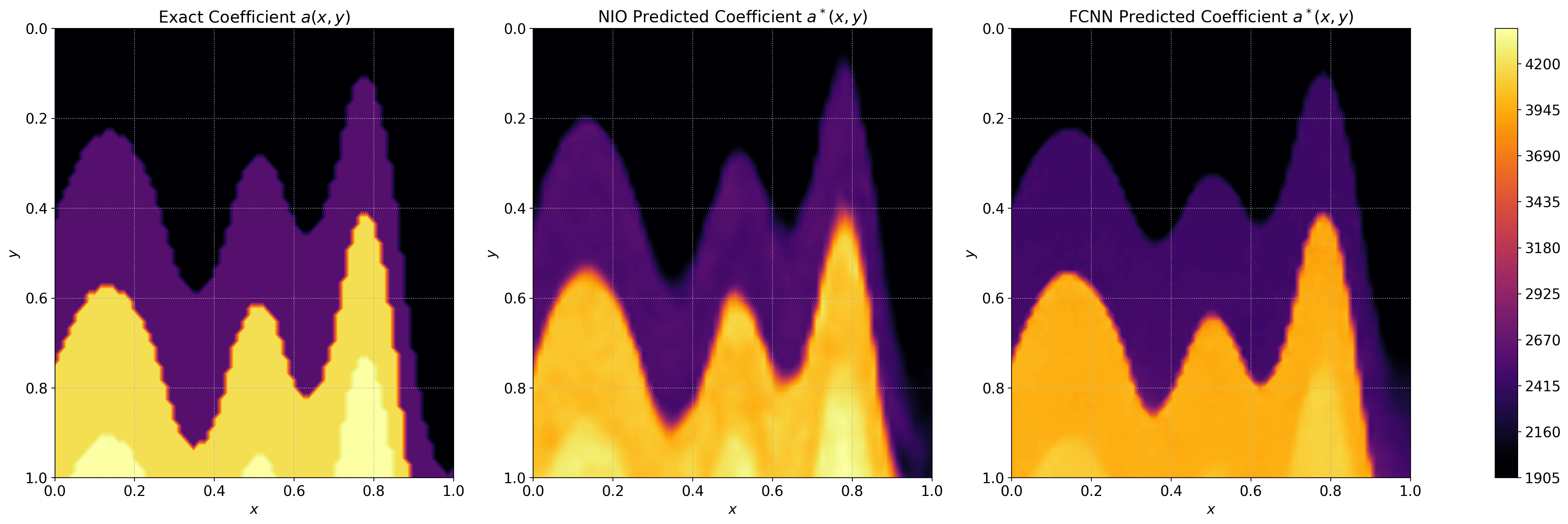}}
        \caption{Test Sample 2}
        \label{fig:wfi:curve:0}
    \end{subfigure}
    \caption{Exact and predicted coefficients for two different test samples (Rows) for the Seismic Imaging with the acoustic wave equation with \emph{CurveVel A} data set. Left Column: Ground Truth. Middle Column: NIO reconstruction. Right Column: FCNN Reconstruction.}
\label{fig:wfi:curve}
\end{figure*}
 \begin{figure*}[htbp]
    \begin{subfigure}{1\textwidth}
        \centering
        \includegraphics[width=1\linewidth]{{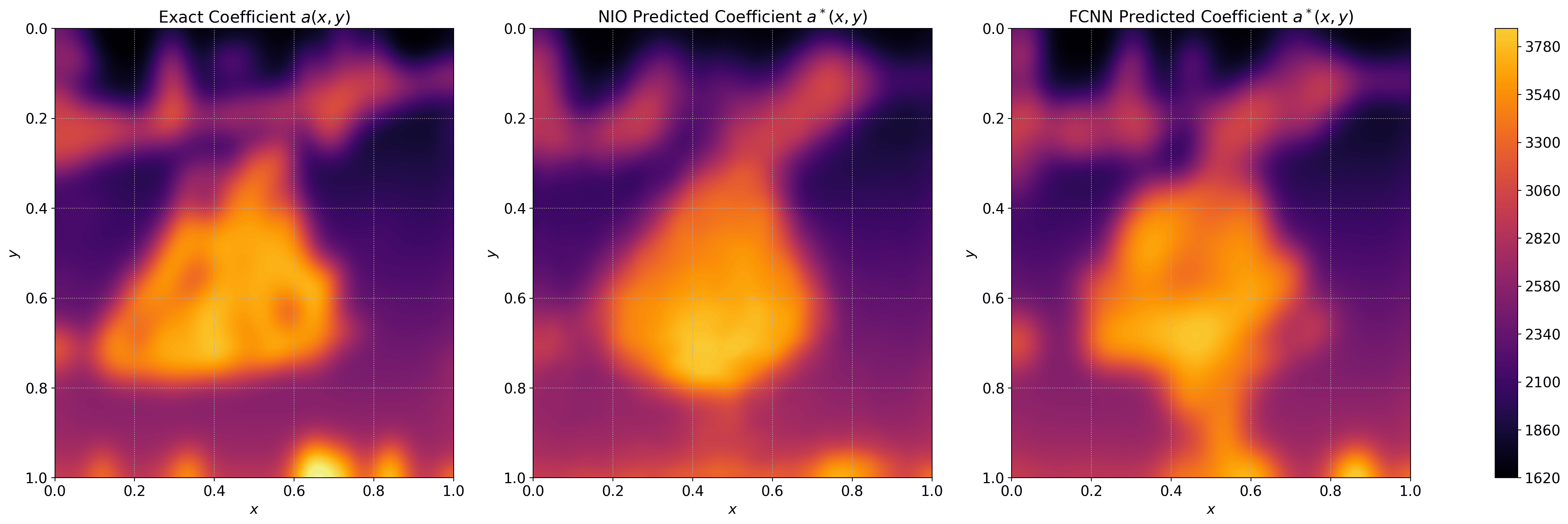}}
         \caption{Test Sample 1}
         \label{fig:wfi:style:1}
    \end{subfigure}
    \begin{subfigure}{1\textwidth}
        \centering
        \includegraphics[width=1\linewidth]{{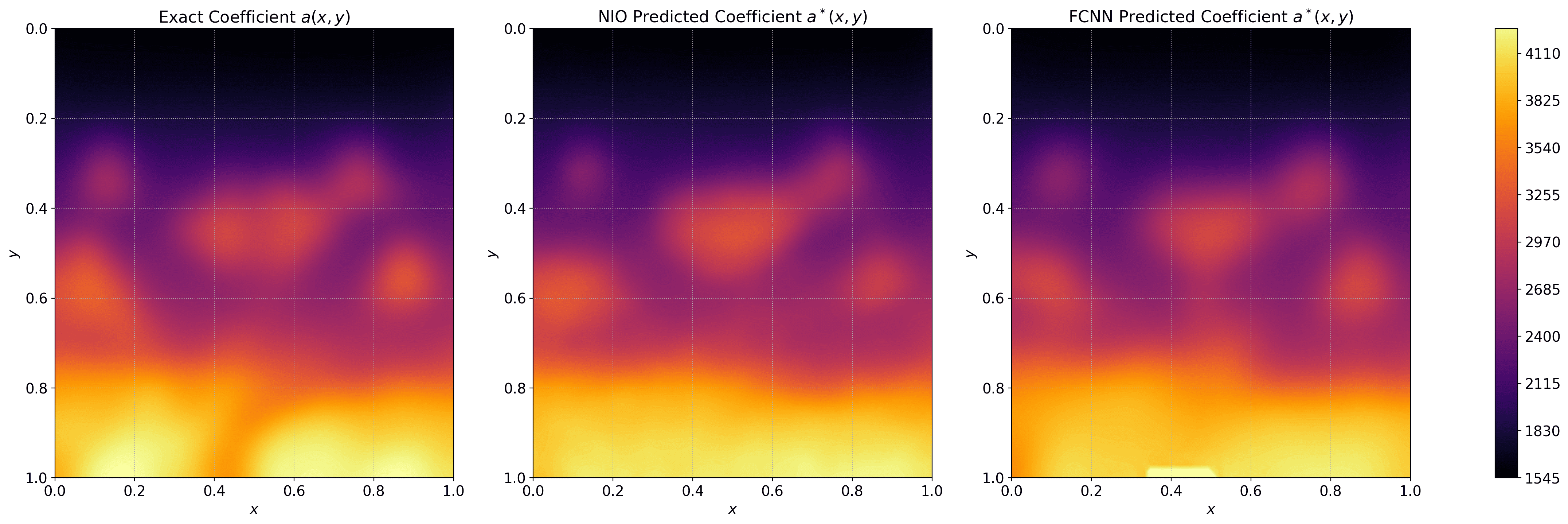}}
        \caption{Test Sample 2}
        \label{fig:wfi:style:0}
    \end{subfigure}
    \caption{Exact and predicted coefficients for two different test samples (Rows) for the Seismic Imaging with the acoustic wave equation with \emph{Style A} data set. Left Column: Ground Truth. Middle Column: NIO reconstruction. Right Column: FCNN Reconstruction.}
\label{fig:wfi:style}
\end{figure*}

We start by elaborating on the results obtained on the benchmarks and presented in Table \ref{tab:res}. In Figure \ref{fig:calderon:sine}, we show two randomly drawn test samples for the Calder\'on Problem for inferring conductivity with trigonometric coefficients by EIT. For both these test samples, we see that NIO (and FCNN) can accurately approximate the ground truth without any visible artifacts. This observation correlates with very small test errors with NIO. At least for these two samples, there appears to be little visible difference between NIO and FCNN. Nevertheless, the results from Table \ref{tab:res} demonstrate that NIO outperforms FCNN considerably on this problem by almost halving the test error. 

\begin{figure*}[ht!]
    \begin{subfigure}{1\textwidth}
        \centering
        \includegraphics[width=1\linewidth]{{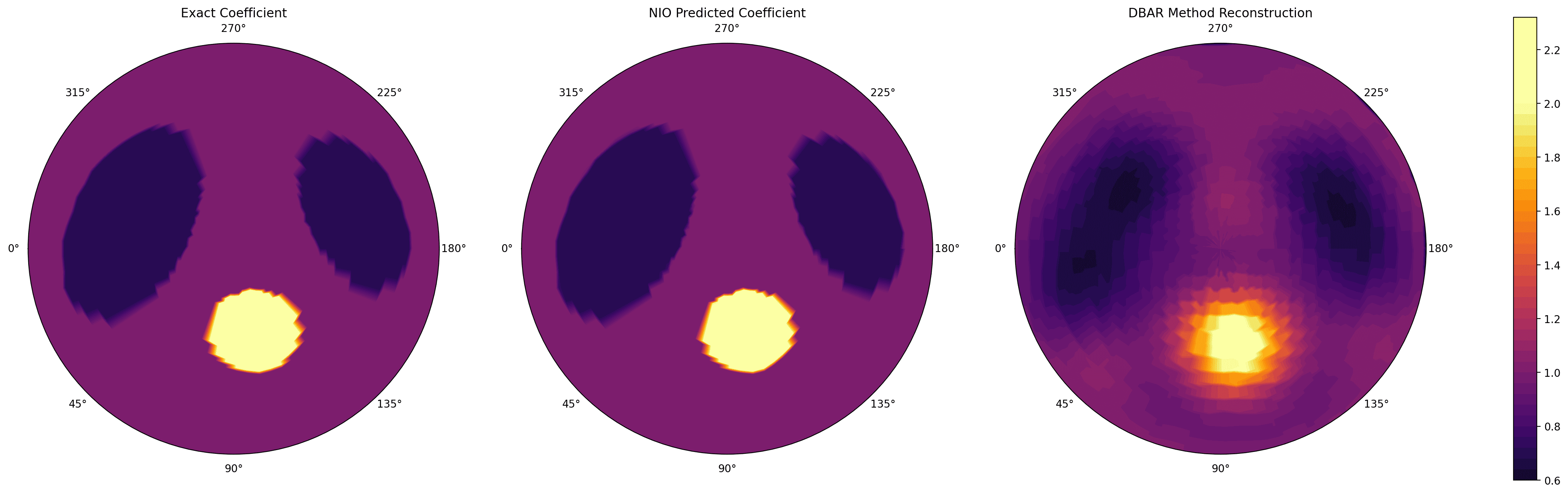}}
        \caption{Test Sample 1}
        \label{fig:calderon:heart:0}
    \end{subfigure}
    \begin{subfigure}{1\textwidth}
        \centering
        \includegraphics[width=1\linewidth]{{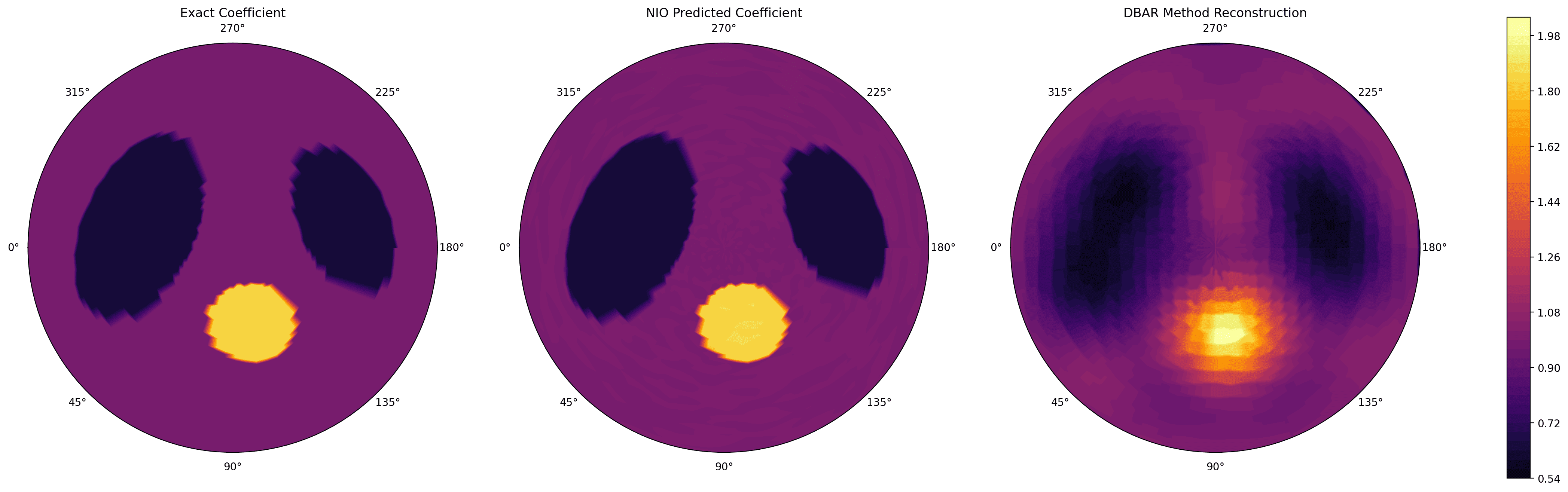}}
        \caption{Test Sample 2}
        \label{fig:calderon:heart:2}
    \end{subfigure}
    \caption{Exact and predicted coefficients for two different test samples (Rows) for the Calder\'{o}n problem with Trigononmetric -coefficients. Left Column: Ground Truth. Middle Column: NIO reconstruction. Right Column: Reconstruction with the D-bar Direct method of \cite{MSbook}.}
\label{fig:calderon:heart}
\end{figure*}

Next, in Figure \ref{fig:calderon:heart}, we focus on the discontinuous heart-lungs Phantom inferred with EIT. Also, in this case, there is no visual difference between the NIO and FCNN, which are both very accurate in reconstructing the ground truth, and this is indeed consistent with the very low generalization error achieved by both models.

In Figure \ref{fig:helm}, we plot the results of two randomly chosen test samples for the inverse wave scattering problem and compare the ground truth with the reconstruction with NIO and FCNN. In the first sample (top row), both models accurately reconstruct the ground truth coefficient with very little visible difference between the competing models. In contrast, in the second sample, the reconstruction with NIO and FCNN are noticeable differences. In particular, FCNN cannot reconstruct the small rectangular scatterer (at the top right of the square domain), whereas NIO can reconstruct it. This possibly explains why NIO is significantly more accurate (see Table \ref{tab:res}) for this experiment in reconstructing scatterers.  

In Figure \ref{fig:radiative}, we plot two randomly chosen test samples to recover the absorption coefficient with optical imaging for the Radiative transport equation \eqref{eq:RTE}. The ground truth and reconstructions obtained with NIO and FCNN are shown. For the first test sample, both models can provide an accurate reconstruction with a sharp resolution of the discontinuities in the absorption coefficient. On the other hand, for the second sample (Figure \ref{fig:radiative} Right), we see that FCNN gets the correct location but the wrong magnitude of the discontinuity, whereas NIO can approximate both accurately, probably accounting for the significant gain in accuracy on this problem (see Table \ref{tab:res} of main text).

In Figures \ref{fig:wfi:curve} and \ref{fig:wfi:style}, we show two randomly chosen test samples for Seismic imaging of the subsurface property (squared slowness) by the acoustic wave equation \eqref{eq:wave}, corresponding to the \emph{CurveVel-A} and \emph{Style A} datasets (considered in \cite{deng2021openfwi}), respectively. Both figures show that NIO and FCNN reconstruct the coefficient reasonably accurately, although slight differences exist between the models. Nevertheless, coupled with quantitative results from Table \ref{tab:res}, we can conclude that NIO is at least on par with FCNN, which was shown to be one of the state-of-the-art models in this context in~\cite{deng2021openfwi}.

\begin{table}[ht]

\begin{center}
\begin{small}
%\begin{sc}
  \begin{tabular}{ l l l l l l l l l l l}
    \toprule
    \multirow{3}{*}{} &
    \multicolumn{2}{c}{\bfseries DONet} &
      \multicolumn{2}{c}{\bfseries FCNN} &
      \multicolumn{2}{c}{\bfseries NIO}  \\
      \midrule
    & $L^1\downarrow$ & $L^2\downarrow$ & 
    $L^1\downarrow$ & $L^2\downarrow$ &  
    $L^1\downarrow$ & $L^2\downarrow$   \\
    \midrule\midrule
\bfseries\makecell{Calder\'{o}n Problem \\ Trigonometric}& \makecell{1.3\% \\ 3.19\%}& \makecell{1.58\% \\ 3.75\%}& \makecell{1.08\% \\ 2.24\%}& \makecell{1.33\% \\ 2.7\%}& \makecell{0.56\% \\ 1.21\%}& \makecell{0.7\% \\ 1.47\%}\\ \midrule
\bfseries\makecell{Calder\'{o}n Problem \\ Heart\&Lungs}& \makecell{0.87\% \\ 1.42\%}& \makecell{3.32\% \\ 4.11\%}& \makecell{0.25\% \\ 0.44\%}& \makecell{1.37\% \\ 3.5\%}& \makecell{0.16\% \\ 0.2\%}& \makecell{0.88\% \\ 1.53\%}\\ \midrule
\bfseries\makecell{Inverse Wave Scattering }& \makecell{2.28\% \\ 5.55\%}& \makecell{4.37\% \\ 11.0\%}& \makecell{0.84\% \\ 4.98\%}& \makecell{2.08\% \\ 13.47\%}& \makecell{0.3\% \\ 2.15\%}& \makecell{1.08\% \\ 5.92\%}\\ \midrule
\bfseries\makecell {Radiative transport} & \makecell{1.56\% \\ 3.38\%}& \makecell{2.72\% \\ 6.5\%}& \makecell{1.0\% \\ 2.26\%}& \makecell{2.02\% \\ 6.0\%}& \makecell{0.74\% \\ 1.63\%}& \makecell{1.82\% \\ 4.86\%}\\ \midrule
\bfseries\makecell {Seismic Imaging \\ CurveVel - A}& \makecell{3.25\% \\ 4.93\%}& \makecell{4.79\% \\ 7.4\%}& \makecell{2.01\% \\ 3.53\%}& \makecell{3.98\% \\ 6.52\%}& \makecell{2.16\% \\ 3.36\%}& \makecell{3.83\% \\ 5.81\%}\\ \midrule
\bfseries\makecell{Seismic Imaging \\ Style - A}& \makecell{2.97\% \\ 4.88\%}& \makecell{4.02\% \\ 6.61\%}& \makecell{2.39\% \\ 4.1\%}& \makecell{3.49\% \\ 6.09\%}& \makecell{2.4\% \\ 3.83\%}& \makecell{3.4\% \\ 5.6\%}\\
    
    \bottomrule
  \end{tabular}
  \end{small}
\end{center}
\caption{0.25 and 0.75 quantiles of the relative $L^1$-error and $L^2$-error computed over testing samples for different benchmarks and models.}
\label{tab:res_quant}
\end{table}

\subsection{Robustness of Reconstruction to $\Lambda_a$-Discretizazion.}
\label{sec:rob_disc}
As outlined in Section \ref{sec:nio}, one crucial property that the model should exhibit is robustness to the number of samples $\Tilde{L}$ used to approximate the pushforward measure $\mu_\Psi$. To assess this, we conduct two different experiments.

\begin{figure*}[htbp]
    \begin{subfigure}{0.5\textwidth}
        \centering
        \includegraphics[width=1\linewidth]{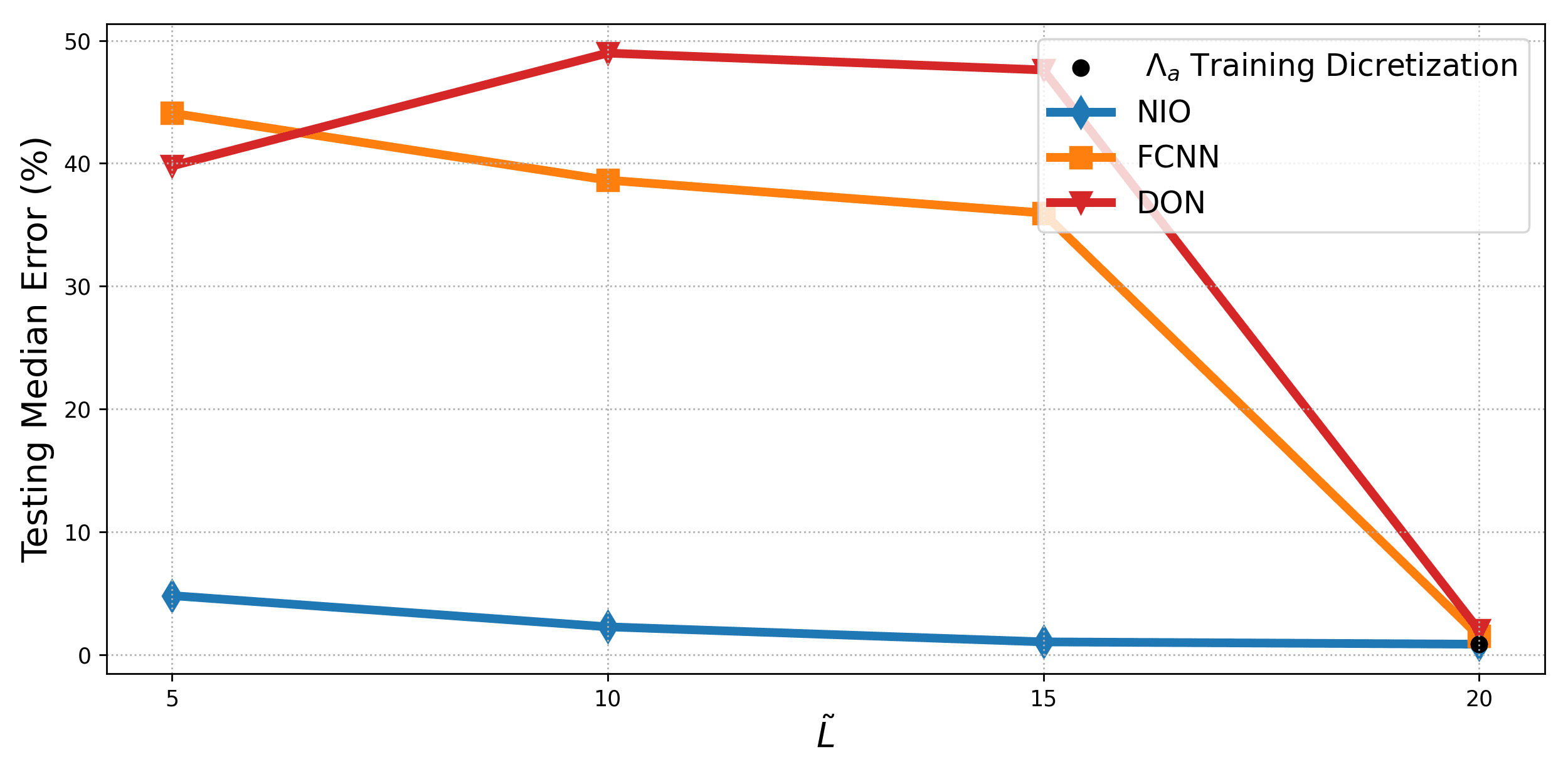}
         \caption{Calder\'{o}n Problem Trigonometric}
         \label{fig:diffL:sine}
    \end{subfigure}
    \begin{subfigure}{0.5\textwidth}
        \centering
        \includegraphics[width=1\linewidth]{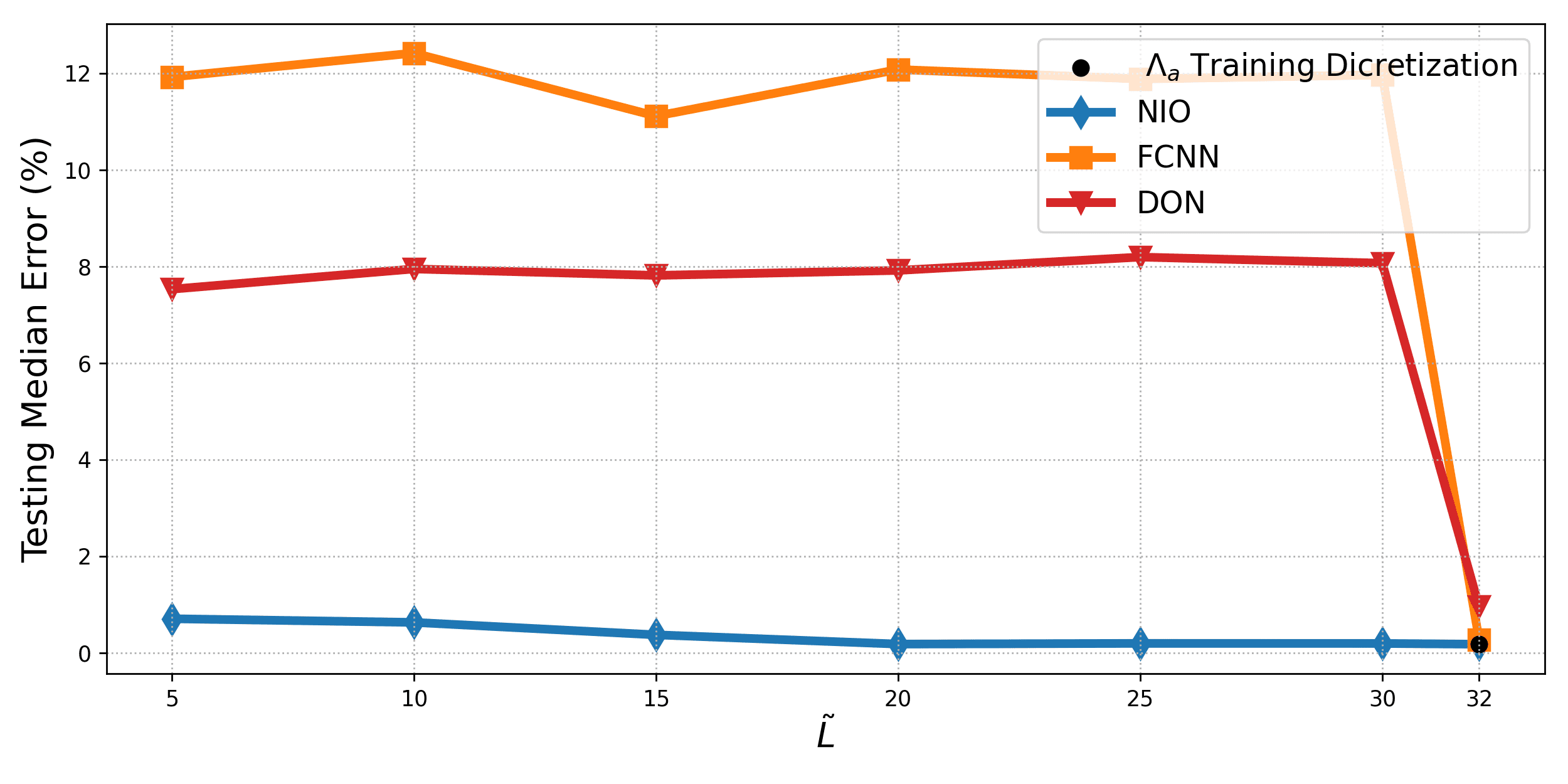}
        \caption{Calder\'{o}n Problem Heart\&Lungs}
        \label{fig:diffL:eit}
    \end{subfigure}
    
    \begin{subfigure}{0.5\textwidth}
        \centering
        \includegraphics[width=1\linewidth]{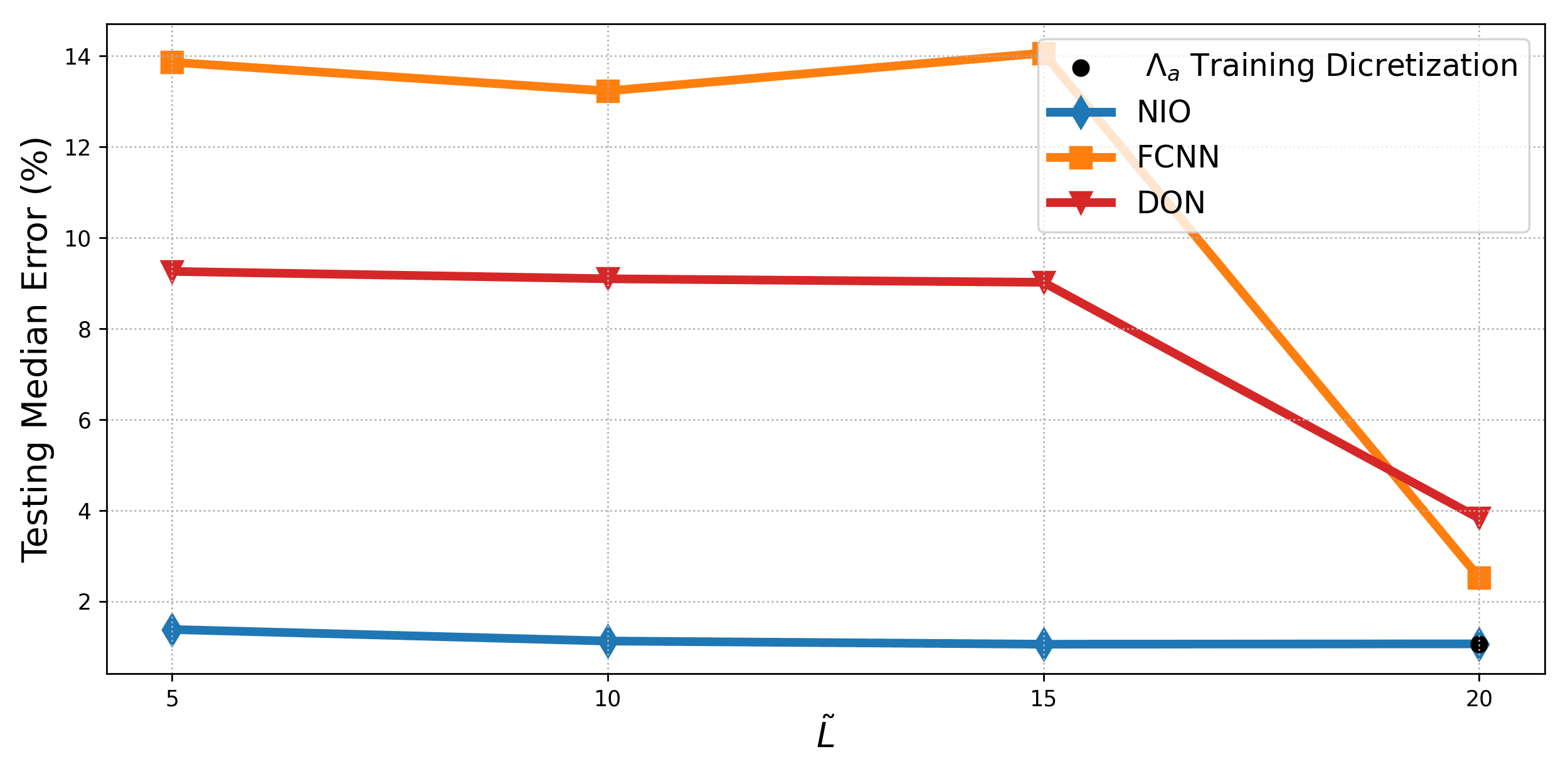}
         \caption{Inverse Wave Scattering }
         \label{fig:diffL:helm}
    \end{subfigure}
    \begin{subfigure}{0.5\textwidth}
        \centering
        \includegraphics[width=1\linewidth]{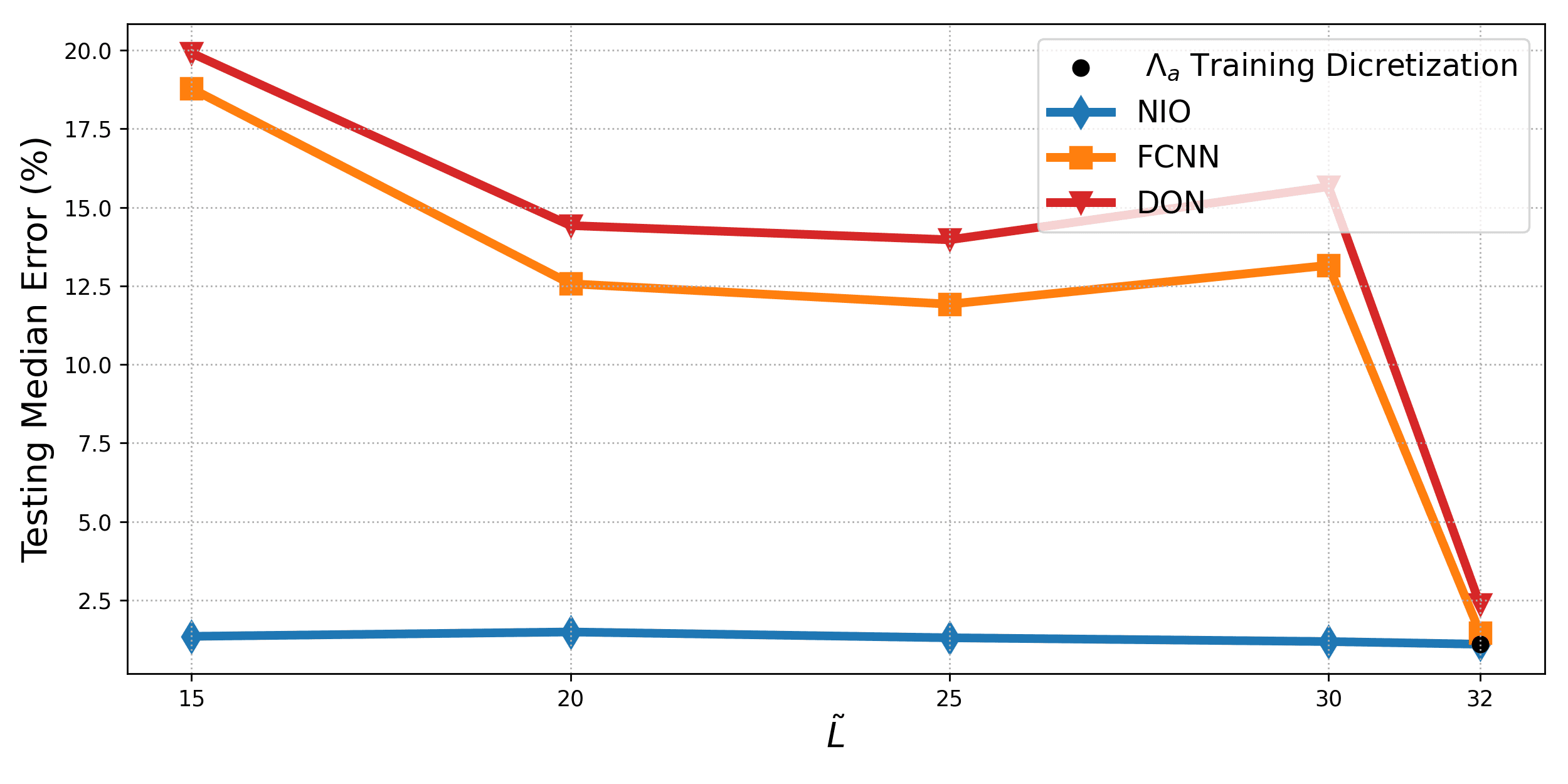}
         \caption{Radiative transport}
         \label{fig:diffL:rad}
    \end{subfigure}
    
    \begin{subfigure}{0.5\textwidth}
        \centering
        \includegraphics[width=1\linewidth]{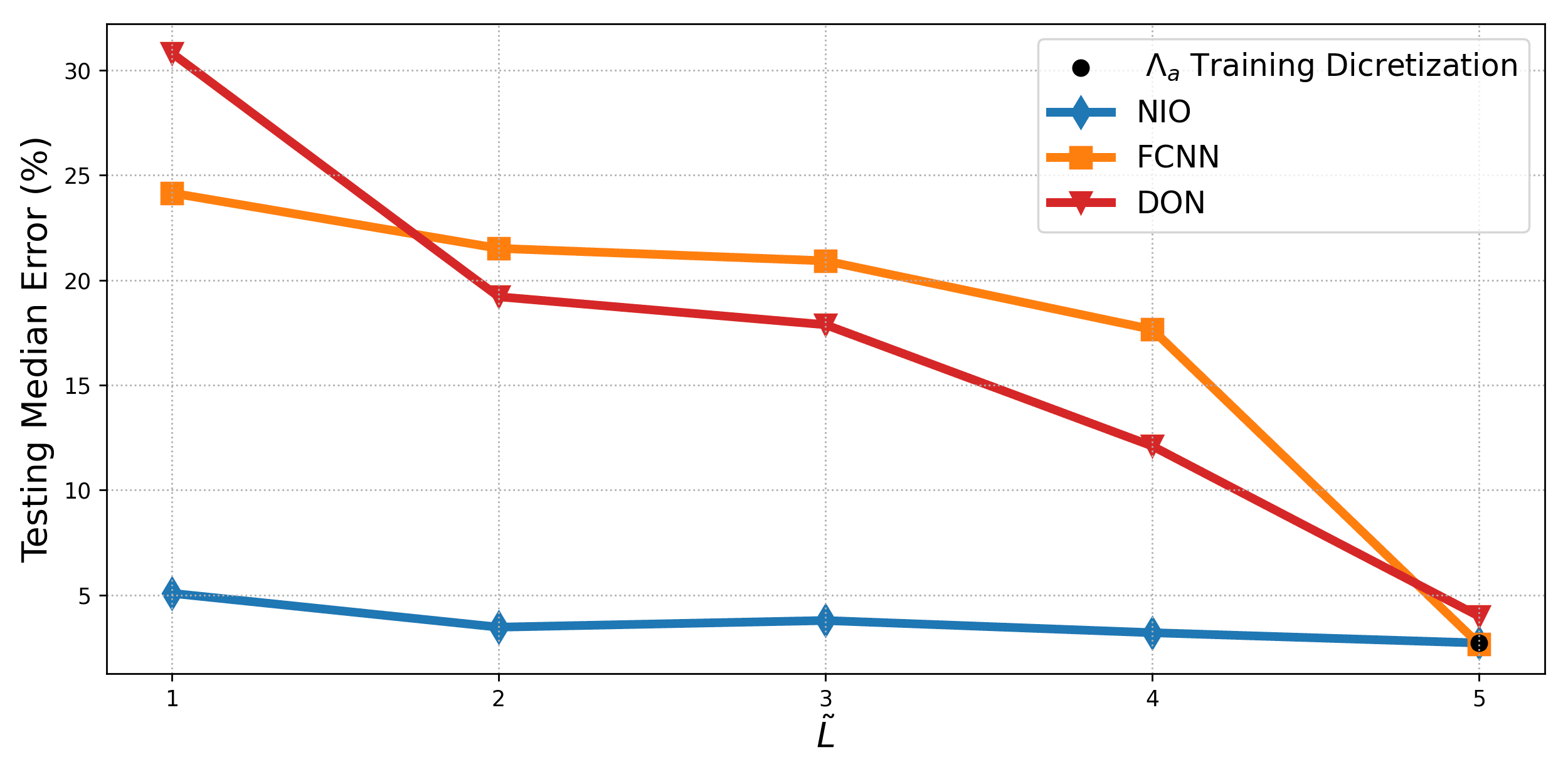}
        \caption{Seismic Imaging CurveVel - A}
        \label{fig:diffL:curve}
    \end{subfigure}
    \begin{subfigure}{0.5\textwidth}
        \centering
        \includegraphics[width=1\linewidth]{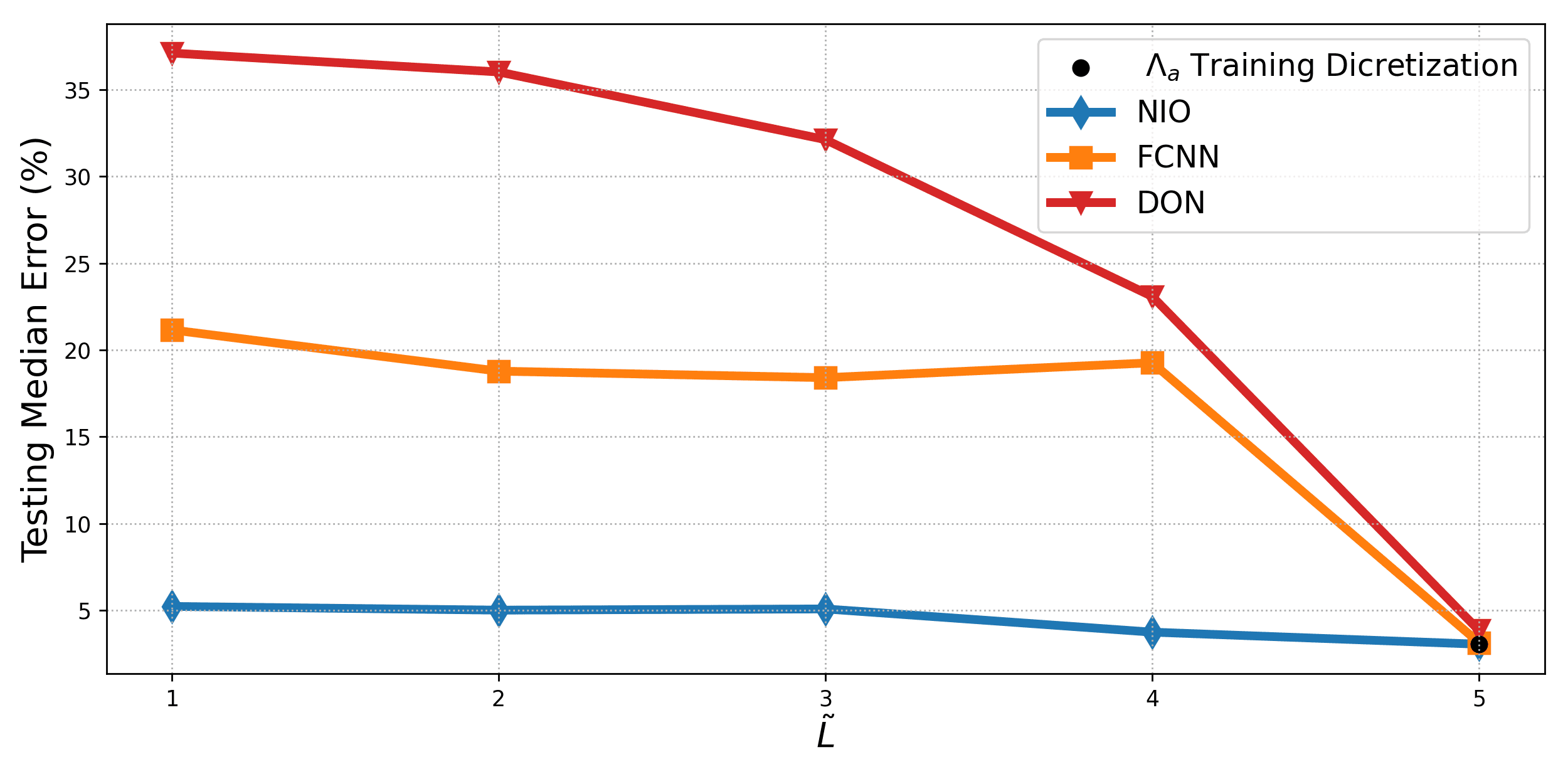}
         \caption{Seismic Imaging  StyleVel - A}
         \label{fig:diffL:style}
    \end{subfigure}
    
    \caption{Median of the $L^1$-error computed over testing samples $(\{\Psi_k\}_{k=1}^{\Tilde{L}}, a)$ VS $\Tilde{L}$ for different benchmarks with different models  ($\Tilde{L}<L$). }
\label{fig:diffL}
\end{figure*}

In the first experiment, we consider the test set used to compute the errors reported in the main table \ref{tab:res} and consisting of input-output pairs $(\{\Psi_\ell\}_{\ell=1}^L, a)$.
For each model (NIO and baselines) and each benchmark, we construct a new test set by picking at random $\Tilde{L}$ samples from $\{\Psi_\ell\}_{\ell=1}^L$, $\Tilde{L}\leq L$ and compute the corresponding testing error. As a remark, it should be noted that NIO can be evaluated directly for any input $\{\Psi_k\}_{k=1}^{\Tilde{L}}$, without no change in the architecture. On the other hand, in order to even evaluate the baselines, interpolation (we chose to use the nearest interpolation) must be used to obtain inputs consisting of exactly $L$ samples. 
In Figure \ref{fig:diffL}, we plot the median $L^1$-error obtained as a function of the number of samples $\Tilde{L}$ for different models and benchmarks. We observe that the performance of NIO remains invariant with respect to $\Tilde{L}$, with the testing error only showing a \textit{slight} increase as $\Tilde{L}$ decreases, which is typically expected since the approximation of the measure $\mu_\Psi$ by the empirical distribution becomes less accurate as the number of samples decreases. In contrast, the performance of the baselines deteriorates even when the number of samples $\Tilde{L}$ is extremely close to the training set size. This demonstrates that the baselines are \textit{not invariant }to permutations of the samples $\{\Psi_\ell\}_{\ell=1}^L$.

\begin{figure*}[htbp]
    \begin{subfigure}{0.5\textwidth}
        \centering
        \includegraphics[width=1\linewidth]{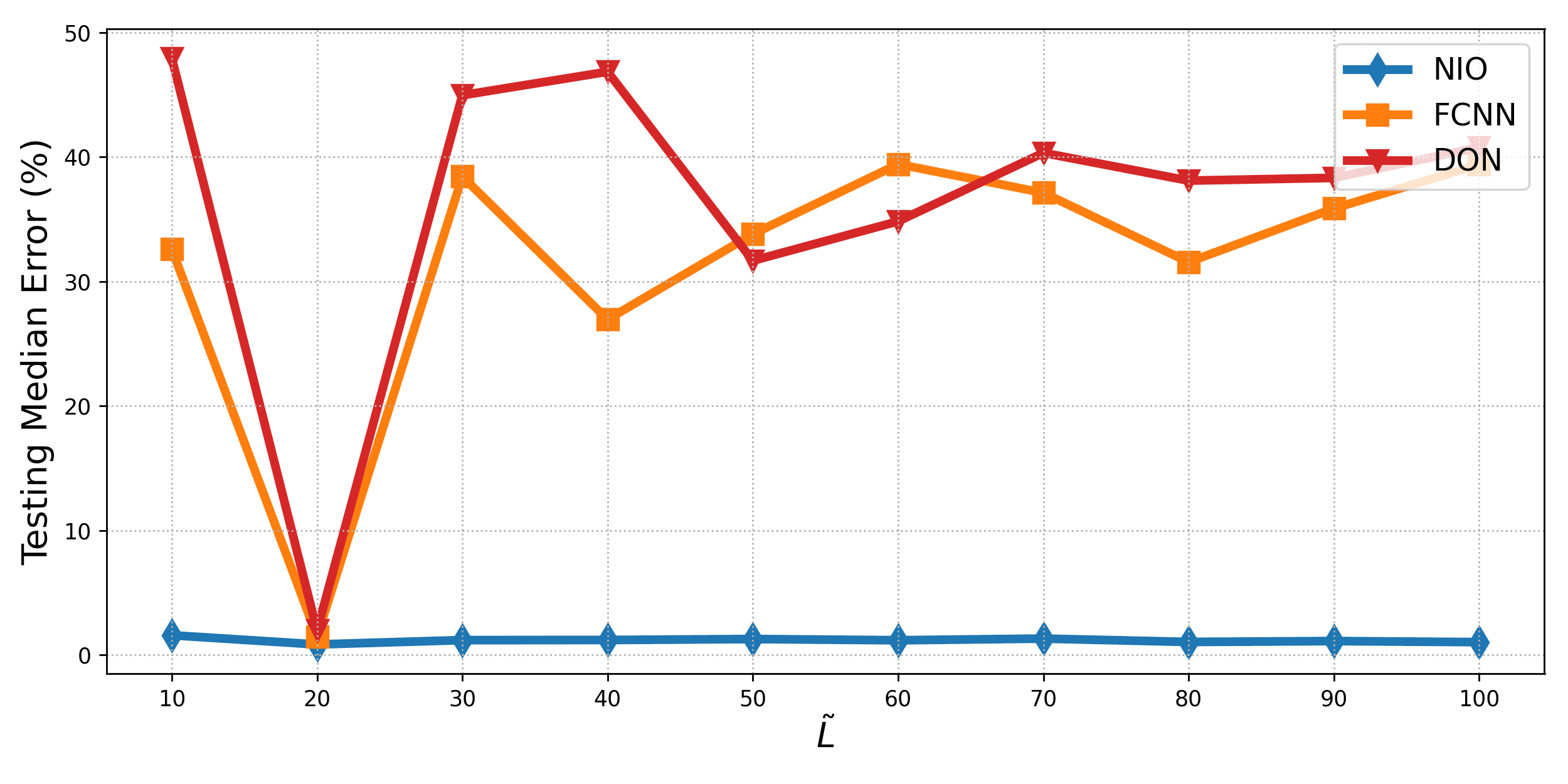}
         \caption{Calder\'{o}n Problem Trigonometric}
         \label{fig:moreL:sine}
    \end{subfigure}
    \begin{subfigure}{0.5\textwidth}
        \centering
        \includegraphics[width=1\linewidth]{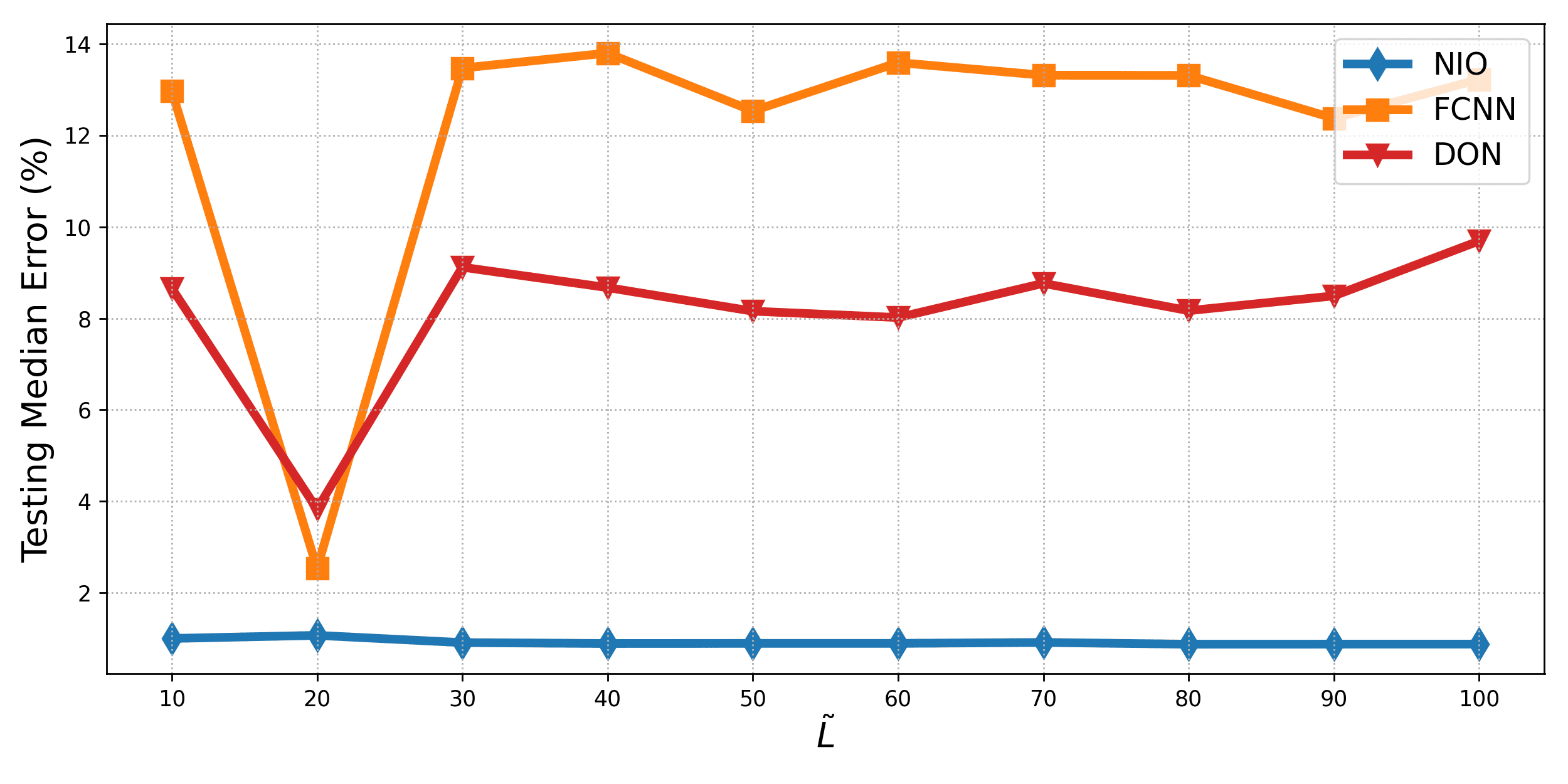}
        \caption{Inverse Wave Scattering }
        \label{fig:moreL:helm}
    \end{subfigure}
    \caption{Median of the $L^1$-error computed over testing samples $(\{\Psi_k\}_{k=1}^{\Tilde{L}}, a)$ VS $\Tilde{L}$ for different benchmarks with different models ($\Tilde{L}$ spans the entire range 10-100).}
\label{fig:moreL}
\end{figure*}

Next, we consider the Calder\'{o}n problem with trigonometric functions and the inverse wave scattering problem. We generate a new testing set from scratch, consisting of input-output pairs $(\{\Psi_\ell\}_{\ell=1}^L, a)$, where $L=100$ (compared to the $L=20$ samples used for model training). We then conduct the same experiments as before and present the results in Figure \ref{fig:moreL}. These results further reinforce the fact that NIO exhibits invariance with respect to the discretization of the input measure $\mu_\Psi$.

\subsection{Robustness of Reconstruction to Noise}
Inverse problems, such as the abstract PDE inverse problem \eqref{eq:imap},  can be very sensitive to noise as the stability estimate \eqref{eq:stab1} indicates, and reconstruction methods have to show some robustness with respect to noisy measurements in order to be practically useful. To test the robustness of NIO (and competing models) to noise, we take all the benchmark test problems reported in Table \ref{tab:res} of the main text and add $1\%$ noise to the inputs to each model at test time. Table \ref{tab:res_noise} presents the resulting test errors. This table shows that NIO (as well as DOnet and FCNN) is very robust to this measurement noise. 

Furthermore, upon closer examination of the results, it is evident that the models displaying the highest robustness with respect to additional noise are those trained using the \textit{log-MinMax} data scaling transformation. To validate this observation, we consider the inverse wave scattering problem and train NIO with the hyperparameters reported in Table \ref{tab:nio_params}, but employing the \textit{log-MinMax} scaling of the data. Additionally, instead of monitoring the validation error computed on the noiseless data, we monitor the validation error computed on data corrupted by $10\%$ noise and interrupt the training based on this metric. The final median testing error on $1\%$-noisy data is $1.64\%$, two times lower than the value reported in Table \ref{tab:res_noise}. Moreover, the testing error on the noise-free data only marginally increased to $1.61\%$. These findings suggest that utilizing \textit{log-MinMax} scaling and potentially monitoring the validation error on the corrupted data can significantly enhance the model's robustness to noise, with minimal loss in performance on the noise-free data.

\begin{table}[ht]
\begin{center}
\renewcommand{\arraystretch}{1.1} 
\begin{footnotesize}
%\begin{sc}
  \begin{tabular}{ l l l l }
    \toprule
    \multirow{1}{*}{} &
    \multicolumn{1}{c}{\bfseries DONet} &
      \multicolumn{1}{c}{\bfseries FCNN} &
      \multicolumn{1}{c}{\bfseries NIO}  \\
    \midrule\midrule
   \bfseries\makecell{Calder\'{o}n Problem \\ Trigonometric}& \makecell{2.02\%}& \makecell{1.51\%}& \makecell{0.91\%}\\ \midrule
\bfseries\makecell{Calder\'{o}n Problem \\ Heart\&Lungs}& \makecell{0.95\%}& \makecell{0.27\%}& \makecell{0.18\%}\\ \midrule
\bfseries\makecell{Inverse Wave Scattering }& \makecell{3.83\%}& \makecell{2.54\%}& \makecell{3.72\%}\\ \midrule
\bfseries\makecell {Radiative transport} & \makecell{2.38\%}& \makecell{1.47\%}& \makecell{1.1\%}\\ \midrule
\bfseries\makecell {Seismic Imaging \\ CurveVel - A}& \makecell{3.98\%}& \makecell{2.65\%}& \makecell{2.73\%}\\ \midrule
\bfseries\makecell{Seismic Imaging \\ Style - A}& \makecell{3.82\%}& \makecell{3.13\%}& \makecell{3.09\%}\\
    \bottomrule
  \end{tabular}
  \end{footnotesize}
  \caption{Median of the relative $L^1$-error computed over $1\%$-noisy testing samples for different benchmarks with different models.\label{tab:res_noise}}
\end{center}
\end{table}

\subsection{Robustness of Reconstructions to Varying Grid Sizes.}
Although the inputs and outputs to the inverse problem \eqref{eq:imap} are continuous objects in principle, in practice, one has to deal with discretized versions of both inputs and outputs. This is true when the ground truth is generated by numerical simulations and observed through other forms of measurement. It is highly desirable that an operator learning algorithm be robust to the resolutions at which it is tested; see \cite{NO} for further discussion on this topic. To test if the proposed NIO architecture is robust with respect to resolution, we focus on the inverse wave scattering with the Helmholtz equation example, where NIO was trained with data obtained from a finite difference scheme on a uniform $70 \times 70$ grid. To test the robustness with respect to resolution, we use this trained model to also infer at two different resolutions, namely at $50 \times 50$ and $100 \times 100$, and present the results, together with DeepONet and FCNN baselines in Table \ref{tab:res_res} to observe that NIO (and the baselines) is robust to varying resolutions. 
\begin{table}[ht]
\begin{center}
\renewcommand{\arraystretch}{1.1} 
\begin{footnotesize}
%\begin{sc}
  \begin{tabular}{ l l  l l l l l l l}
    \toprule
    \multirow{2}{*}{} &
    \multicolumn{2}{c}{\bfseries DONet} &
      \multicolumn{2}{c}{\bfseries FCNN} &
      \multicolumn{2}{c}{\bfseries NIO}  \\
      \midrule
    \makecell{Resolution} & $50\times50$  & $100\times100$  & $50\times50$ &  $100\times100$ & $50\times50$ & $100\times100$ \\
    \midrule\midrule
    \bfseries\makecell{Inverse Wave Scattering}& \makecell{3.74\%}& \makecell{3.63\%}& \makecell{1.81\%}& \makecell{1.66\%}& \makecell{0.93\%}& \makecell{0.95\%}\\
    \bottomrule
  \end{tabular}
  \end{footnotesize}
  \caption{Relative median $L^1$-error computed over testing samples generated at different resolutions (grid sizes).}
\label{tab:res_res}
\end{center}
\end{table}

\subsection{Robustness of Reconstruction to Random Sensors Location.}

While training data typically assumes equidistant placement of sensors along the boundaries of the square domain (as they are synthetically generated using standard numerical methods), real-world scenarios often involve sensors located randomly along the boundaries. Hence, the learning model must exhibit robustness to these random sensor placements. 

Our experiments to assess this robustness focus on two specific problems: the Calder\'{o}n problem with trigonometric function and inverse wave scattering. We perform testing with input data obtained from $200$ sensors randomly distributed along the domain boundary. Observe that the training data accounts for $272$ sensors. Therefore, before feeding the data to NIO (and baselines), we interpolate it onto the original equispaced set of points. For both problems, we examine the $ L^1$ error, as presented in Table \ref{tab:res_randomloc}.

The results show that the $L^1$-error increases only to $1.18\%$ and $1.43\%$, compared to the original setup where the errors were $0.86\%$ and $1.11\%$, respectively. These findings underscore the NIO model's ability to maintain robust performance even when the boundary sensors are placed at different locations.

\begin{table}[ht]
\begin{center}
\begin{small}
%\begin{sc}
  \begin{tabular}{ l l l l }
    \toprule
     &
    {\bfseries DONet} &
      {\bfseries FCNN} &
      {\bfseries NIO}  \\
      \midrule
    & $L^1\downarrow$ & 
    $L^1\downarrow$ &  
    $L^1\downarrow$   \\
    \midrule\midrule
    \bfseries\makecell{Calder\'{o}n Problem  Trigonometric}& \makecell{ 3.39\%}& \makecell{ 1.88\%}& \makecell{ 1.18\% }\\ \midrule
    \bfseries\makecell{Inverse Wave Scattering  }& \makecell{ 3.84\% }& \makecell{2.53\%}& \makecell{1.43\% }\\ 
    \bottomrule
  \end{tabular}
  \end{small}
  \caption{Relative median $L^1$-error computed over testing samples with random location of the boundary measurements with different models and different benchmarks. }
  \label{tab:res_randomloc}
\end{center}
\end{table}

\subsection{Out-of-Distribution Reconstruction.}
\label{app:ood}

In addition to \emph{in-distribution} testing, we also consider an \emph{out-of-distribution} testing task. This will enable us to evaluate the ability of the models to \emph{generalize} to inputs (and outputs) that possess different features from the training ones.

First, we considered the Calder\'on Problem (Trigonometric) benchmark. The coefficients in the training distribution were sampled from the exponential of a sum of sines, with up to 4 frequency modes (up to $8\pi$). We now test with the following distributions:
\begin{itemize}
    \item Distribution A:
    $a(x, y) = \exp\big(\sum_{k=1}^{6}c_{k}\sin(k\pi x)\sin(k\pi y)/k^{\frac{3}{2}} \big)$
    \item Distribution B:
    $a(x, y) = \exp\big(\sum_{k=1}^{6}c_{k}\sin(k\pi x)\sin(k\pi y)/k \big)$
\end{itemize}
Here, ${c_k}$ is a uniformly distributed random variable in the range $[0,1]^{m}$. The coefficients include up to $6$ frequency modes, with different decays of the higher order modes ($1.5$ and $1$). Thus, at test time, the model now has to infer data with significantly higher frequencies than the training data.

Next, we consider the Calderon problem of the Heart and Lungs. Here, the training distribution was based on a normally distributed perturbation of the Heart and Lungs Phantom, with the amplitude of the perturbation being at most $8\%$ of the Phantom values (See {\bf SM} \ref{fig:heart_sample}). At test time, we now consider perturbations with amplitudes being $12\%$ of the Phantom values, thus sampling from a different distribution. For instance, this higher amplitude could model individuals with some diseases. Note that even higher variations are probably unrealistic as this problem models body organs and one has to restrict to some biological constraints.

As a third benchmark, we consider the inverse scattering problem. Here, the training distribution was of a coefficient that consisted of between 1-4 scatterers, \emph{of identical shape}, whose locations were randomly chosen (See {\bf SM} \ref{fig:step_sample}). We chose to test this model now on two different test distributions 
\begin{itemize}
    \item Distribution A:
    $a(x, y) = \sum_{k=1}^{5} \exp\big(-c(x-c_{1,k} )^4 - c(y-c_{2,k})^4\big)$. In this case, we use a family of coefficients with a fixed number of inclusions equal to five.
    \item Distribution B: $a(x, y) = \sum_{k=1}^{m} \exp\big(-b_k^4(x-c_{1,k} )^4 - b_k^4(y-c_{2,k})^4\big),$
    with $b_k \sim \mathcal{U}[5, 15]$, $m = \mod(\bar{m})$, $m\sim \mathcal{U}([1,4])$ and $\{(c_{1,k}, c_{2,k})\} \sim \mathcal{U}([0,1]^{m\times 2})$. This corresponds to a medium with one to four scatterers with varying shapes.
\end{itemize}

The relative median $L^1$ error for different models and different out-of-distribution testing is reported in Table \ref{tab:res_outs}.  We observe that NIO generalizes well to unseen data, with test errors increasing at most by approximately a factor of $4$, and still outperforms the baselines in all cases.

\begin{table}[ht]
\begin{center}
\begin{small}
%\begin{sc}
  \begin{tabular}{ l l l l }
    \toprule
     &
    {\bfseries DONet} &
      {\bfseries FCNN} &
      {\bfseries NIO}  \\
      \midrule
    & $L^1\downarrow$ & 
    $L^1\downarrow$ &  
    $L^1\downarrow$   \\
    \midrule\midrule
    \bfseries\makecell{Calder\'{o}n Problem \\ Trigonometric Distribution A}& \makecell{1.37\%}& \makecell{1.27\%}& \makecell{1.2\%}\\ \midrule
    \bfseries\makecell{Calder\'{o}n Problem Trigonometric \\ Distribution B}& \makecell{1.62\%}& \makecell{1.28\%}& \makecell{0.91\%}\\ \midrule
    \bfseries\makecell{Calder\'{o}n Problem \\ Heart\&Lungs}& \makecell{1.05\%}& \makecell{0.28\%}& \makecell{0.19\%}\\ \midrule
    \bfseries\makecell{Inverse Wave Scattering \\ Distribution A }& \makecell{4.61\%}& \makecell{3.84\%}& \makecell{3.0\%}\\ \midrule
    \bfseries\makecell{Inverse Wave Scattering \\ Distribution B }& \makecell{8.61\%}& \makecell{8.98\%}& \makecell{4.54\%}\\
    \bottomrule
  \end{tabular}
  \end{small}
  \caption{Relative median $L^1$-error computed over out-of-distribution test samples with different models and different benchmarks. }
  \label{tab:res_outs}
\end{center}
\end{table}

\subsection{Ablation Studies.}
\label{app:abl}

 \begin{figure*}[ht!]
    \begin{subfigure}{0.5\textwidth}
        \centering
        \includegraphics[width=1\linewidth]{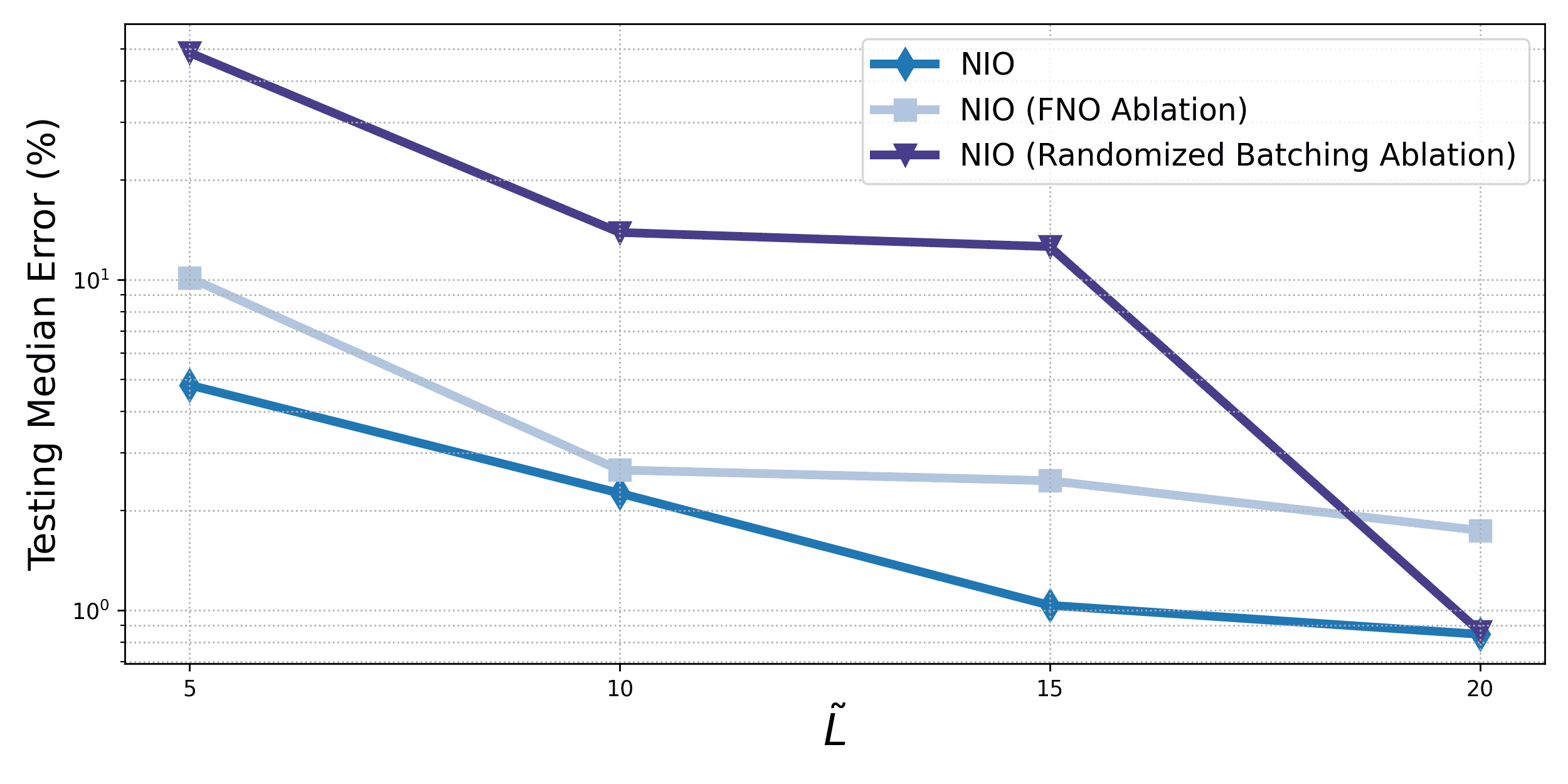}
         \caption{Calder\'{o}n Problem Trigonometric}
         \label{fig:abl:sine}
    \end{subfigure}
    \begin{subfigure}{0.5\textwidth}
        \centering
        \includegraphics[width=1\linewidth]{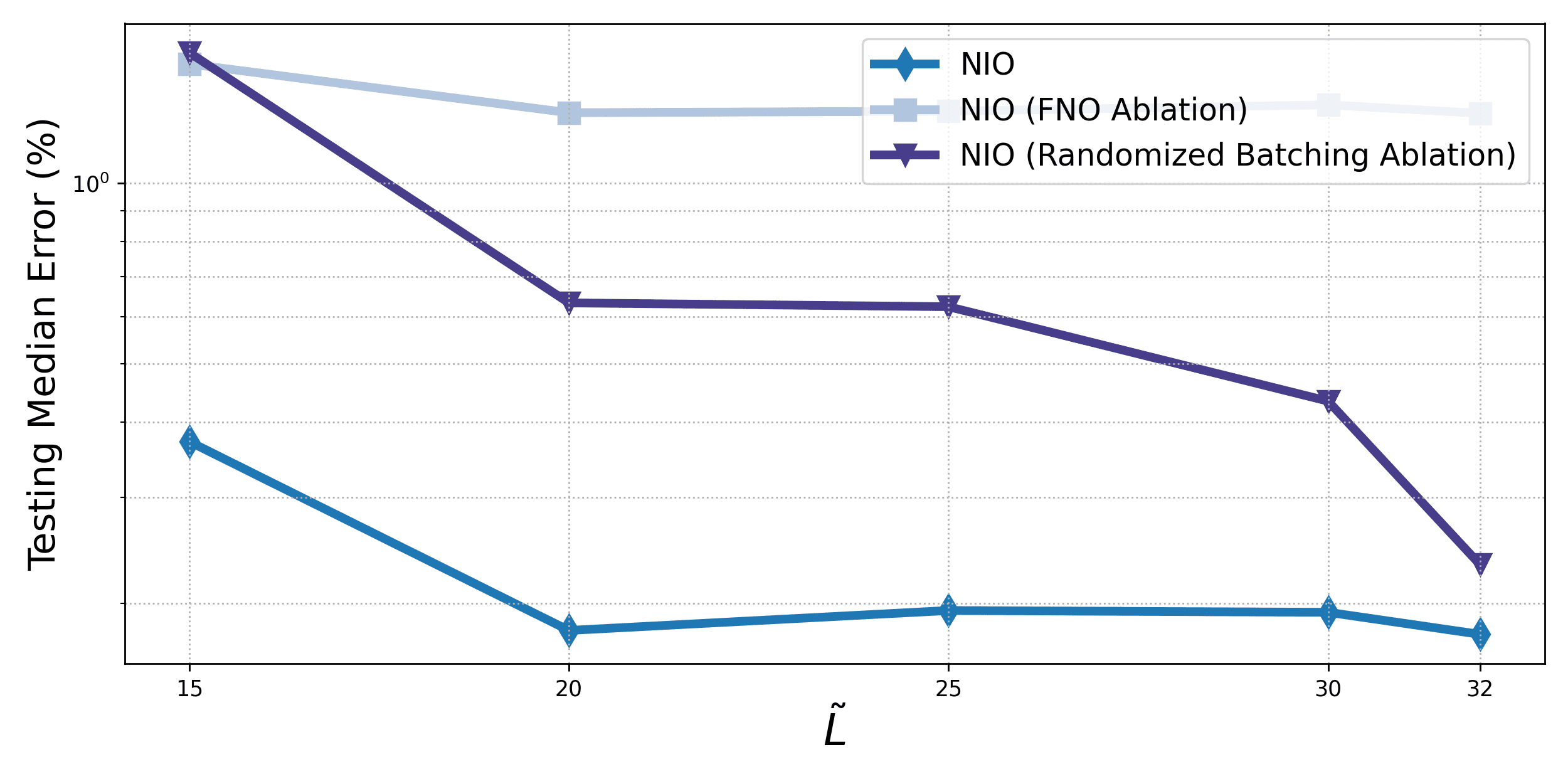}
         \caption{Calder\'{o}n Problem Heart\&Lungs}
         \label{fig:abl:eit}
    \end{subfigure}
    \begin{subfigure}{0.5\textwidth}
        \centering
        \includegraphics[width=1\linewidth]{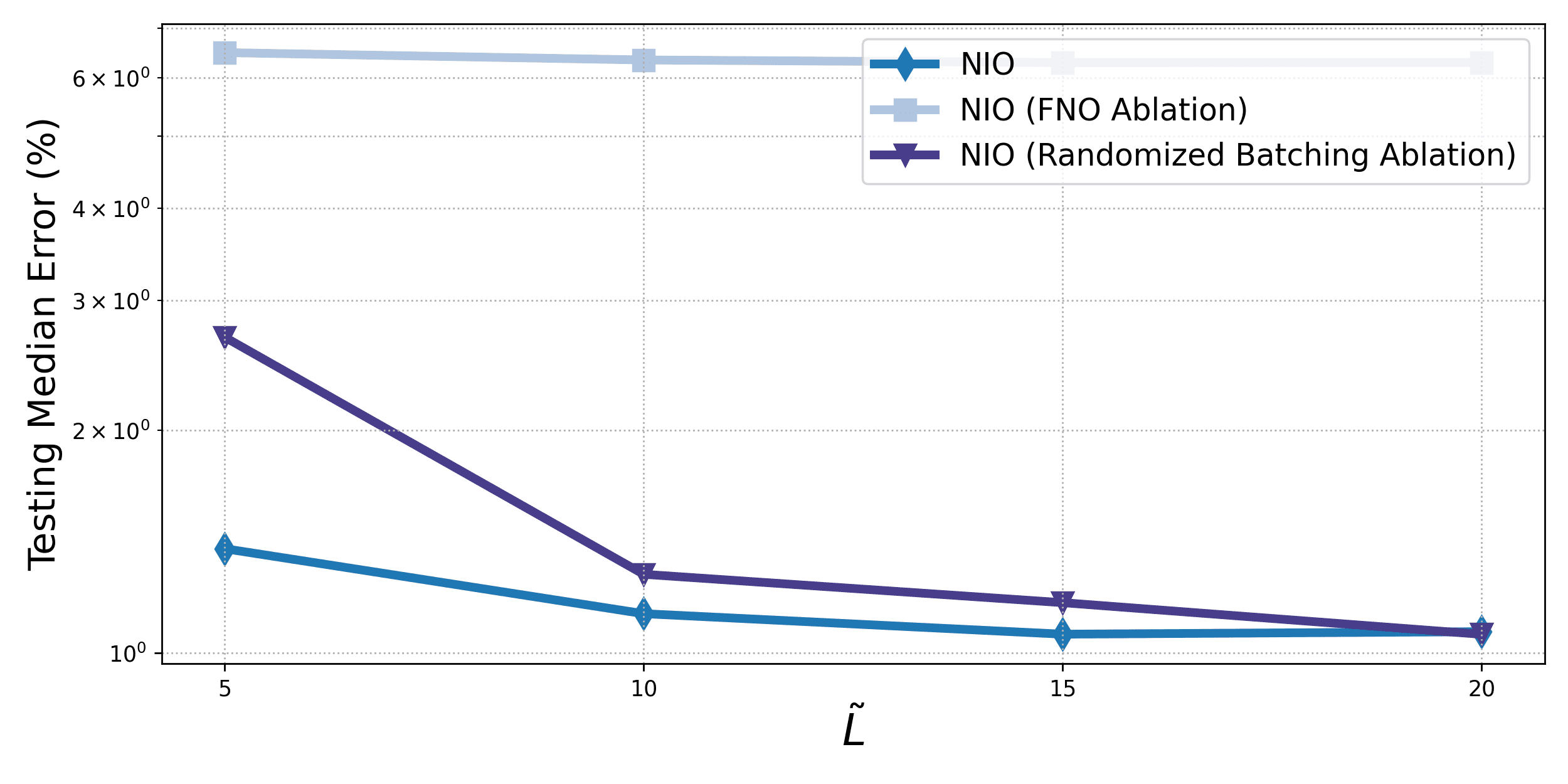}
        \caption{Inverse Wave Scattering}
        \label{fig:abl:helm}
    \end{subfigure}
    \begin{subfigure}{0.5\textwidth}
        \centering
        \includegraphics[width=1\linewidth]{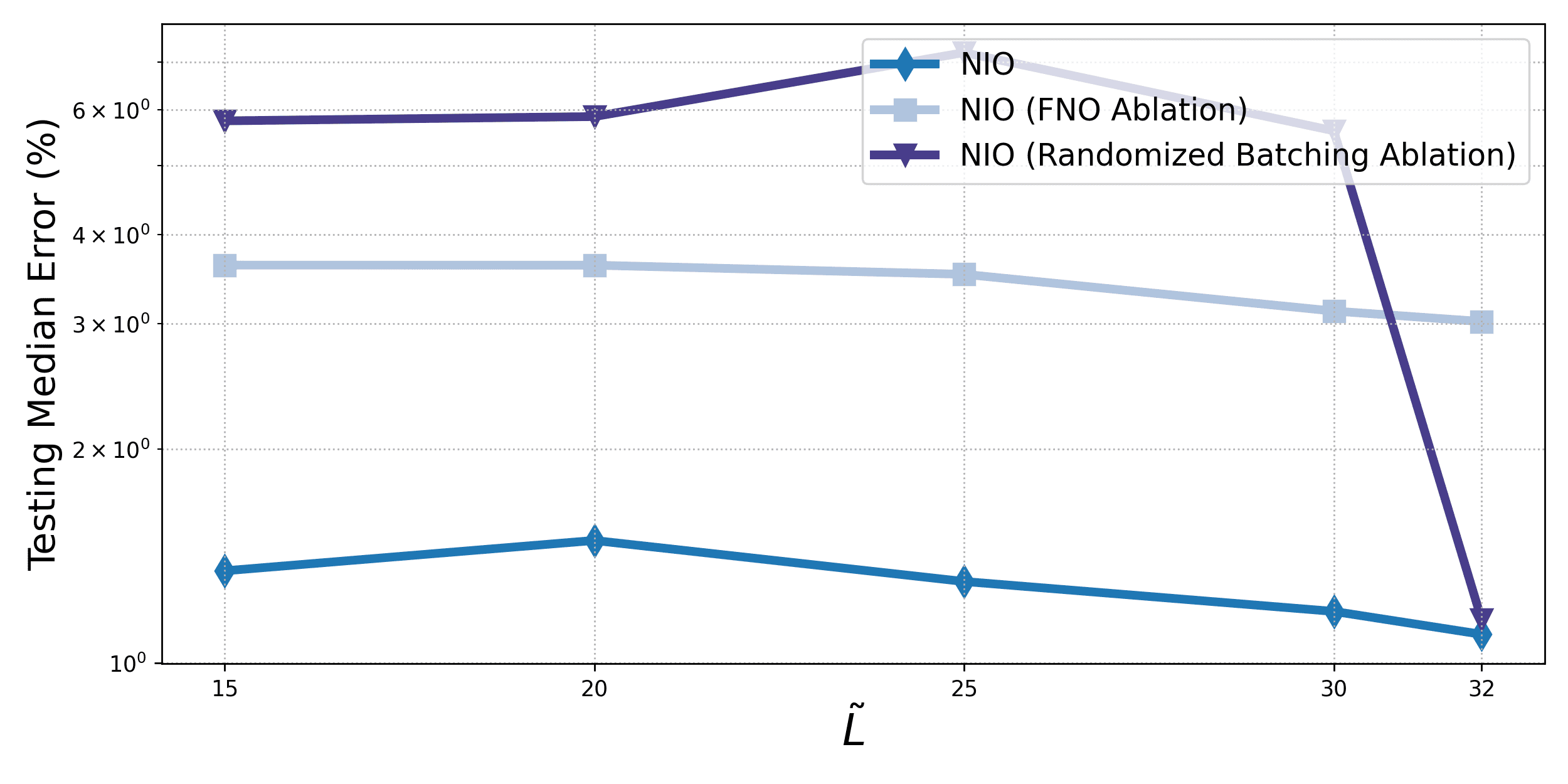}
        \caption{Radiative Transport}
        \label{fig:abl:rad}
    \end{subfigure}

    \caption{Median of the $L^1$-error computed over testing samples $(\{\Psi_k\}_{k=1}^{\Tilde{L}}, a)$, VS $\Tilde{L}$ for different benchmarks with different models (NIO and ablations). }
\label{fig:abl}
\end{figure*}

We conduct two ablation studies focusing on two key elements of the Neural Inverse Operator. Firstly, we focus on the architecture shown in Figure \ref{fig:ffDeepFNO}, where we remove the nonlinear part $\eM$ of FNO and set $d_v=1$. In this case, we have:
\begin{equation}
a^\ast(z) = R(f_1,\dots,f_L, z) = \frac{D}{L}\sum_{\ell=1}^L f_\ell + Ez,
\end{equation}
where $E$ and $D$ are real-valued parameters. With these experiments, we aim to assess if the channels' mixing realized with $\eM$ could improve the model's performance. Secondly, we examine the influence of randomized batching by training NIO without including the algorithm.

In order to maintain consistency, we use the same hyperparameter configurations for the ablation models as those of the best-performing NIO models (refer to Table \ref{tab:nio_params} for the specific values).

The ablation study is being carried out for the Calder\'{o}n problem with trigonometric function and heart\&lungs Phantom, inverse wave scattering, and radiative transport.  We consider the same experimental setup outlined in Section \ref{sec:rob_disc} and report the corresponding results in Figure \ref{fig:abl}. The figure shows that the nonlinear term $\eM$ and randomized batching significantly improve the model's performance. For the Calder\'on problem with trigonometric coefficients, removing $\eM$ resulted in a nearly 2-times increase of the generalization error. On the other hand, for the remaining problems, the improvement is considerably more relevant, up to a factor of six for the inverse wave scattering. 

It should be noted that in all experiments, only between $L=20$ and $L=32$ boundary measurements are used for training. In this scenario, the limited measurement data represents a bottleneck in accurately reconstructing the target coefficient. With a larger number, the nonlinear part may lead to an even greater reduction in the generalization error.

Regarding randomized batching, the improvements for the Calder\'on problem and radiative transport are substantial, ranging from 5 to 10 times compared to the ablated version. On the other hand, the improvements are more modest for the Inverse Wave Scattering, and the algorithm does not appear to considerably enhance generalization. However, these results are particularly impressive on account of the virtually zero cost associated with the randomized batching algorithm. 

\subsection{Comparison Standard Numerical Methods for Inverse Problems}
\label{sec:comparison_num}
In this section, we compare the performance of the proposed architecture, in terms of accuracy and inference time, with standard numerical methods, particularly PDE-constrained optimization techniques.

\subsubsection{Calder\'{o}n Problem Heart\&Lungs}
We begin by considering the Calder\'{o}n problem for the discontinuous heart-lungs Phantom and conduct a comparative analysis between the proposed approach and the well-known D-bar method \cite{MSbook}, which is commonly employed in the context of EIT. In Figure \ref{fig:heart_sample}, we present the ground truth along with the NIO reconstruction for a randomly selected set of test samples. For comparison, we also include the reconstruction obtained using the D-bar method. As the figure shows, NIO reconstructs the ground truth to very high accuracy, consistent with the very small errors presented in Table \ref{tab:res}.
On the other hand, the D-bar method is relatively inaccurate and provides a blurred and diffusive reconstruction of the shapes. In fact, the $L^1$-test error for the D-bar method is an unacceptably high $8.75\%$, compared to the almost $0.15\%$ test error with NIO. This is even more impressive when one looks at the run times. The D-bar method takes approximately $2$ hours to run for a single sample, whereas the inference time for NIO is only $0.1$ seconds (on CPU). Thus, we can provide a method which \emph{two orders of magnitude more accurate while being four orders of magnitude faster to run}. This highlights the massive gain in performance with machine learning-based methods, such as NIO, compared to traditional direct methods. 

\subsubsection{Inverse Wave Scattering}
Next, we investigate the Inverse Wave Scattering problem and compare the NIO reconstruction with the results obtained through PDE-constrained optimization. We examine an out-of-distribution (Distribution A) test sample to accomplish this.

For PDE-constrained optimization, we employ a feed-forward neural network with trainable parameters $\theta$ that parameterizes the coefficient $a$. The neural network architecture follows the form specified in Equation \ref{eq:ann1}, with $L$ layers, $d$ neurons, and activation function $\sigma$. The model parameters $\theta$ are initialized randomly, and the Helmholtz equation \eqref{eq:helmholtz} is numerically solved for $L=20$ realizations of the boundary conditions $g_\ell$, where $\ell=1,...,L$ and the $L^2$-loss
\begin{equation}\label{eq:L2 misfit}
    J(\theta) = \sum_{\ell=1}^L|| \Psi_\ell - \Tilde{\Psi}_\ell||^2_{2}
\end{equation} 
computed. Here, $\Tilde{\Psi}_\ell$, $\ell=1,...,L$ denotes the target data.  The trainable network parameters are updated using LBFG in the direction of the gradient of $J(\theta)$. The process is iterated until convergence (1000 iterations, corresponding to $1000\times20$ PDE solves). 

Observe that the hyperparameters of the network are chosen through a random search by picking the one minimizing $J(\theta)$.

The reconstructed coefficient obtained through PDE-constrained optimization, along with the NIO prediction obtained by directly feeding the model with $\Tilde{\Psi}_\ell$ for $\ell=1,\ldots, L$, are depicted in the right panel of Figure \ref{fig:diffL_main}. The NIO reconstruction is significantly more accurate than the one obtained with PDE-constrained optimization, further reinforced by the corresponding $L^1$-error, equal to $2.3\%$ and $11.1\%$, respectively. 
The corresponding total time required to reconstruct the coefficient, amounts to less than $1$ second (on CPU) for NIO and 8.5 hours for the traditional method. It is worth noting that the finite difference (FD) solver employed for solving the equation is implemented on GPU within the PyTorch framework. By solving the PDE in parallel for $L=20$ boundary measurements on 20 different GPUs, the computation time could potentially be reduced to 30 minutes, which is still three orders of magnitude slower than NIO's inference.

Finally, an alternative approach would be to replace the forward solver with a learning model instead of directly learning the inverse map. Although the surrogate model might achieve low errors in learning the forward map (compared to the inverse map), the inference time for the model could be approximately $0.1$ seconds (on a CPU). If we require $1000$ calls to the model for the optimization process, this will result in an overall inference time that is two orders of magnitude longer than NIO. Moreover, the reconstruction accuracy cannot surpass that obtained using the FD forward solver, which is already four times worse than the accuracy achieved with NIO.

\subsubsection{Seismic Imaging}
%\textcolor{red}{For Yunan, shall we add some details?}
We also tested the PDE-constrained optimization method in the context of Seismic imaging with the CurveVel-A dataset. Unlike the NIO approach, in the PDE-constrained optimization approach, we first discretize the velocity coefficient $a(z)$ using a piecewise-constant parameterization, denoting the parameters as $\theta$. 
We then solve the wave equation~\eqref{eq:wave} using the finite difference method. The method is thus naturally dependent on the discretization of $a(z)$ and the PDE solver used to evaluate the source-to-receiver operator. We use the same experimental setup as provided in~\cite{deng2021openfwi}. A similar squared $L^2$ loss to~\eqref{eq:L2 misfit} is used as the objective function to measure the mismatch between the observed data and the simulated receiver data based on the current prediction of $a(z)$. We also solve a corresponding adjoint wave equation whose solution is used to compute the gradient of the objective function $J(\theta)$ with respect to $\theta$. We use the standard gradient descent method to perform the optimization. Each gradient evaluation requires two time-dependent PDE solves, which is the most expensive part of this PDE-constrained optimization approach.

\begin{figure*}[ht!]
    \begin{subfigure}{1\textwidth}
        \centering
        \includegraphics[width=1\linewidth]{{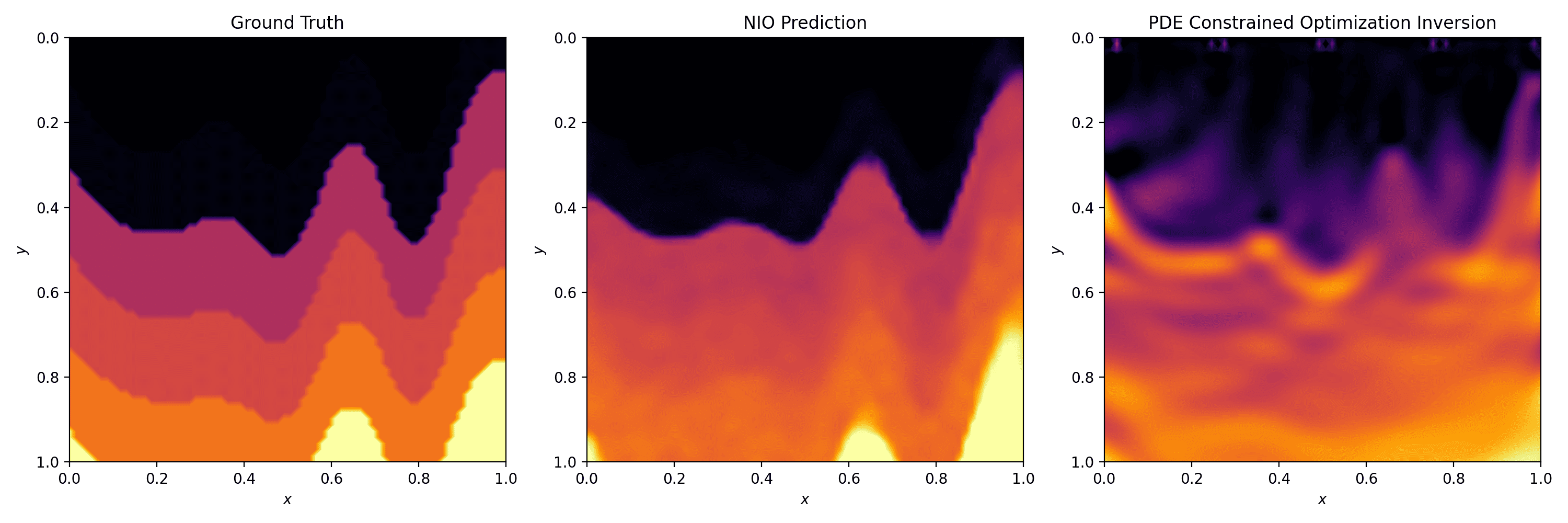}}
        \caption{Test Sample 1}
        \label{fig:wave:pde_cons:0}
    \end{subfigure}
    \begin{subfigure}{1\textwidth}
        \centering
        \includegraphics[width=1\linewidth]{{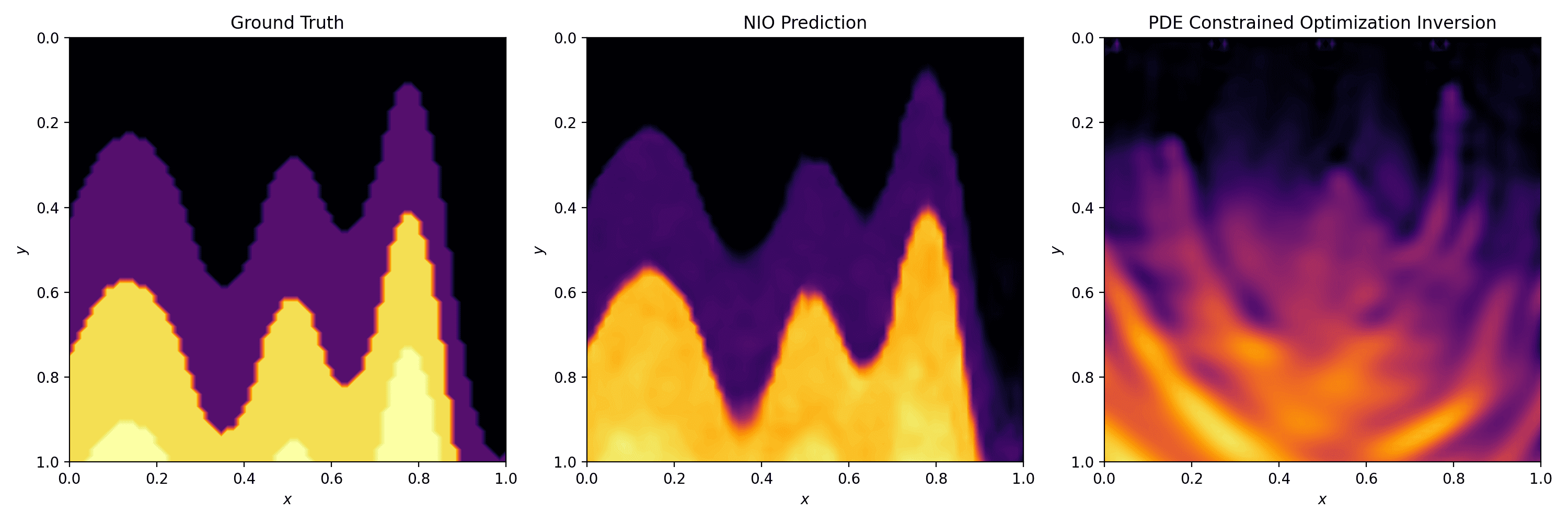}}
        \caption{Test Sample 2}
        \label{fig:wave:pde_cons:2}
    \end{subfigure}
    \caption{Exact and predicted coefficients for two different test samples (Rows) for the Curve Vel family. Left Column: Ground Truth. Middle Column: NIO reconstruction. Right Column: Reconstruction with the PDE-constrained optimization method.}
\label{fig:wave:pde_cons}
\end{figure*}

The method leads to errors (in $L^1$) $2.5$-$4$ times larger than CNIO for the same test sample while taking approximately $30$ minutes of run-time on an 8-core M1-chip CPU. Figure~\ref{fig:wave:pde_cons} depicts the results for two test samples. The $L^1$-error achieved with NIO is  $2.03\%$ and $4.84\%$ for samples 1 and 2, respectively, whereas the ones obtained with PDE-constrained optimization are $5.05\%$ and $16.9\%$.

\end{document}